\documentclass{egpubl}
\usepackage{pg2020}
\usepackage{subcaption}
\usepackage{multirow}
\usepackage{mathtools}

\SpecialIssuePaper         %
\usepackage[T1]{fontenc}
\usepackage{dfadobe}

\biberVersion
\BibtexOrBiblatex
\usepackage[backend=biber,bibstyle=EG,citestyle=alphabetic,backref=true]{biblatex} 
\electronicVersion
\PrintedOrElectronic

\ifpdf \usepackage[pdftex]{graphicx} \pdfcompresslevel=9
\else \usepackage[dvips]{graphicx} \fi
\usepackage{pgf-pie}
\usepackage{egweblnk}

\usepackage[most]{tcolorbox}
\usepackage{enumitem}

\usepackage{colortbl}
\definecolor{amethyst}{rgb}{0.6, 0.4, 0.8}
\usepackage{wrapfig}
\usepackage{caption}

\usepackage[mathscr]{euscript}
\usepackage{dsfont}
\usepackage{amsmath}
\usepackage{amsfonts}
\usepackage{bm}  %

\title[Image-Driven Furniture Style for Interactive 3D Scene Modeling]%
      {Image-Driven Furniture Style for Interactive 3D Scene Modeling}

\author[Weiss et al.]
{\parbox{\textwidth}{\centering
Tomer Weiss$^{1}$\orcid{0000-0001-6649-6916}\quad%
 Ilkay Yildiz$^{2}$\orcid{0000-0002-2827-7672}\quad
 Nitin Agarwal$^{3}$\orcid{0000-0002-8303-2918}\quad
 Esra Ataer-Cansizoglu$^{4,5}$\orcid{0000-0002-1941-4241}\quad
 Jae-Woo Choi$^{5}$\orcid{0000-0003-2578-1213}
        }
        \\
{\parbox{\textwidth}{\centering
$^1$New Jersey Institute of Technology \quad
$^2$Northeastern University \quad
$^3$University of California, Irvine \quad
$^4$Facebook Inc. \quad
$^5$Wayfair Inc.}
}
}

\newcommand{\noofimg}{N}
\newcommand{\noofcomp}{N_c}
\newcommand{\noofexperts}{M}
\newcommand{\nooffeat}{d}
\newcommand{\noofstyles}{L}
\newcommand{\imgindex}{i}
\newcommand{\imgindexone}{i}
\newcommand{\imgindextwo}{j}
\newcommand{\expertindex}{e}
\newcommand{\expertindexone}{e_1}
\newcommand{\expertindextwo}{e_2}
\newcommand{\styleindex}{l}
\newcommand{\imgset}{\mathscr{N}}
\newcommand{\expertset}{\mathscr{M}}
\newcommand{\styleset}{\mathscr{L}}
\newcommand{\feature}{\bm{x}}
\newcommand{\stylelabel}{y}
\newcommand{\score}{s}
\newcommand{\minvote}{l_{\text{min}}}
\newcommand{\compthr}{t}
\newcommand{\recallpos}{k}
\newcommand{\allstylelabels}{\mathcal{D}_s}
\newcommand{\allcleanstylelabels}{\mathcal{D}_{s,\text{clean}}}
\newcommand{\allcomplabels}{\mathcal{D}_c}
\newcommand{\realnumbers}{\mathbb{R}}
\newcommand{\indicator}{\mathds{1}}
\newcommand{\basenet}{f}
\newcommand{\weights}{\bm{W}}
\newcommand{\loss}{\mathcal{L}}

\newcommand{\shiftleft}[2]{\makebox[0pt][r]{\makebox[#1][l]{#2}}}

\definecolor{myGreen}{rgb}{0.839,0.922,0.769}
\definecolor{myRed}{rgb}{0.553,0,0.027}

\begin{document}
\captionsetup{labelfont=bf,textfont=it}

\teaser{
 \centering
  \includegraphics[width=0.99\textwidth]{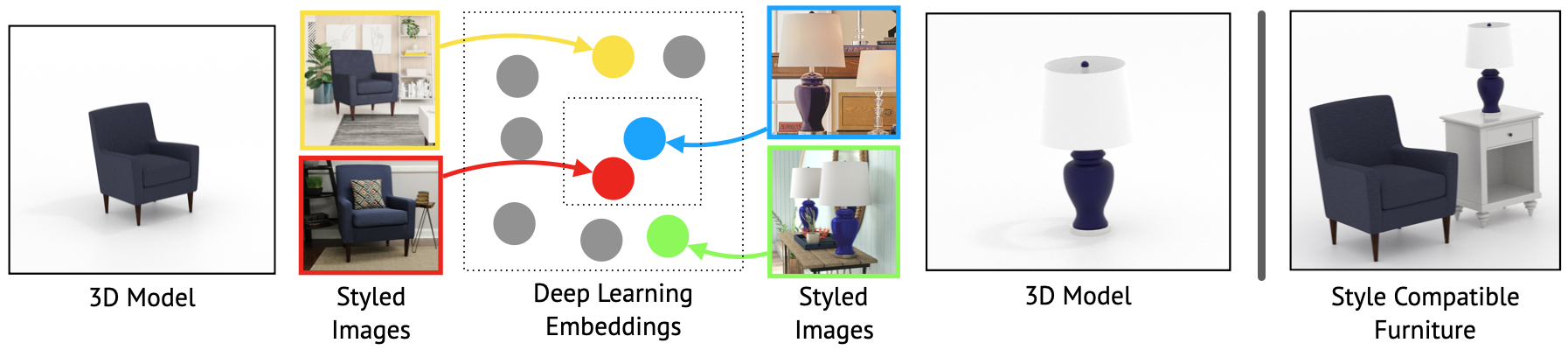}
  \caption{We learn 3D furniture model style using a deep neural network that
  was trained on a data set of annotated curated styled images.
  Using the embeddings from the network, we estimate a 3D models's style compatibility with other models. We demonstrate our approach with an interactive system to create styled interior scenes.}
\label{fig:teaser}
}

\maketitle
\begin{abstract}
Creating realistic styled spaces is a complex task, which involves design know-how for what furniture pieces go well together.
Interior style follows abstract rules involving color, geometry and other visual elements.
Following such rules, users manually select similar-style items from large repositories of 3D furniture models, a process which is both laborious and time-consuming.
We propose a method for fast-tracking style-similarity tasks,
by learning a furniture's style-compatibility from interior scene images.
Such images contain more style information than images depicting single furniture.
To understand style, we train a deep learning network on a classification task.
Based on image embeddings extracted from our network,
we measure stylistic compatibility of furniture.
We demonstrate our method with several 3D model style-compatibility results, and with an interactive system for modeling style-consistent scenes.

%

\printccsdesc
\end{abstract}

\section{Introduction}
With the increasing demand for realistic virtual worlds,
methods for effectively creating virtual scenes is greater than ever.
Such demand is driven by multiple industries: high-fidelity gaming, visual effects and even retail, to name a few. For example,
a realistic game, or scene in a movie, might need to convey
a certain style, appropriate for the story, or time-period.
Similarly, home goods and retail companies rely on virtual scenes for demonstrating their products.
When creating such scene, a user selects objects from a catalog of 3D models that fits their theme, and scene style.
Selecting 3D models that fit well together is necessary for creating appealing and consistent visual quality.

In this work, we propose a method for inferring stylistic compatibly of 3D furniture models, using style context images.
We focus on a set of furniture classes that are prevalent in interior spaces.
Contrary to previous approaches (c.f.~Section~\ref{sec:related}) that mostly focus on learning style
from geometry, or rendered shots of 3D models, we decouple a 3D model's geometry by learning directly from images of styled scenes,
since multiple elements influence style, such as color, texture, material, and the use of the space.
Styled interior images, judged by professionals, contain such elements.
Hence, we learn a 3D model's style based on images of furniture in a curated settings.

\begin{figure}[b!]
\captionsetup[subfigure]{labelformat=empty,textfont=normalfont}
\centering
\subcaptionbox{Modern}{\includegraphics[height=2.8cm,keepaspectratio]{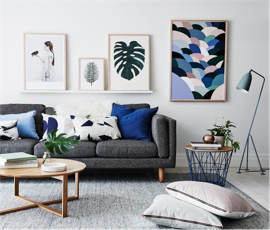}}
\subcaptionbox{Traditional}{\includegraphics[trim=5 0 32 0, clip, height=2.8cm,keepaspectratio]{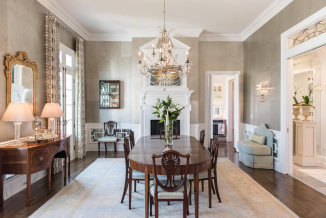}}
\\
\subcaptionbox{Cottage}{\includegraphics[trim=11 0 25 0, clip, height=2.8cm,keepaspectratio]{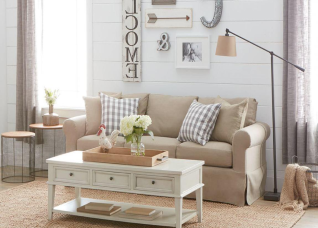}}
\subcaptionbox{Coastal}{\includegraphics[height=2.8cm,keepaspectratio]{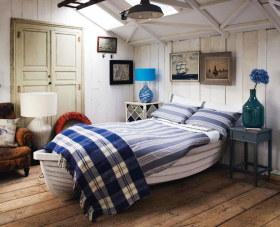}}
\caption{Major interior styles. Each image in our data set is labeled with such styles, to be used by our deep learning style estimation network.}
\label{fig:style_def}
\end{figure}

To learn style, we first need to define what "style" is.
Interior designers define multiple style guidelines.
Typically, each style is described with certain criteria about fabric, color scheme, material, use of space, light, texture, and flooring~\cite{kilmer2014designing}.
In this work, we focus on 4 major highly popular styles: \emph{modern, traditional, cottage and coastal}, as shown in Figure~\ref{fig:style_def}.
Associating a furniture, or scene with a certain style is a challenging and subjective task, since styles are loosely defined, and smoothly change with trends in the field.
Hence, we employ a data-driven approach to estimate style.
We collect a data set of furniture and scene images that are annotated with styles by multiple interior style experts.
Given such an annotated data set, we train a deep siamese network~\cite{bromley1994signature} based on the VGG16 architecture~\cite{simonyan2014very}, to classify an image's style.
Since images in our data set depict furniture, by estimating an image's
style, we estimate a furniture's style.
We evaluate our network's performance on several scenarios, showing that it successfully learns the style spectrum of each image, and improves the classification accuracy compared to other baselines.

We utilze embeddings of the neural network to infer style-compatibility. Once our network is trained, we extract the embeddings for each image. Embeddings encode the low-level style features. Computing the euclidean distance between the embeddings of two images gives us a measure of their stylistic compatibility, e.g., the smaller such distance is, the more style-compatible images are.
Since the images are also associated with 3D furniture models,
we use image embeddings to predict the stylistic-compatibility between two
furniture models.

We perform extensive evaluation experiments on our style compatibility
framework, both qualitative and quantitative, on furniture images and 3D models.
To demonstrate our approach,
we implemented a tool for interactive, style-compatible scene building.
In summary, our contributions are:
\begin{enumerate}[leftmargin=*]
  \item Learning a furniture's style-compatibility from styled images. Styled images contain more information than renderings of pure, isolated 3D models.
  \item Our style estimation embodies texture, color and materials, which are not considered in previous geometry-focused work.
  \item A novel comparison-based approach for learning style from images, which improves the classification accuracy over discrete style labels.
  \item An interactive system for style-compatible modeling of virtual scenes.
\end{enumerate}

\section{Related Work}
\label{sec:related}
Real, and virtual scenes are often designed in a certain style.
Understanding such style is difficult, as evident by numerous interior design professionals offering their services. Organizations, such as Ikea, Pinterest, and Wayfair are actively working toward understanding their customers style-needs~\cite{sachidanandan2019designer, jing2015visual, ataer2019room}.
Researchers have proposed numerous methods for understanding the style of 3D objects and learning their visual compatibility.
Below we classify previous work based on their approach.

\subsection{Geometry-based Style Estimation}
Most earlier methods focused on the geometry of 3D models to infer style.
Xu et al.~\cite{Xu2010anisotropic} defined style as anisotropic part scaling among models.
They relied on consistent segmentation of parts and define the style distance between shapes based on the differences in scales and orientations of part bounding boxes.
Yumer et al.~\cite{yumer2015semantic} proposed to interactively deform models according to precomputed style handles, which are limited to predefined perceptual parameters.
Both Liu et al.\ and Lun et al.~\cite{liu2015style,lun2015elements} learned 3D model style based on human perception. They employ crowd-sourcing to quantify the different components of style.
In a later work, Lun et al.~\cite{lun2016functionality} transferred style from an example to target 3D model.
Hu et al.~\cite{hu2017style} present a method for discovering elements that characterize a given style, where the elements are co-located across different 3D model associated with such style.

While geometry-based methods have made impressive progress in understanding 3D model style compatibility, style estimation is still a challenging problem in the interior furniture style domain, since style is subjective and is also influenced by multiple non-geometric criteria.
Moreover, scaling geometric-based approaches to large data sets is difficult, as they often require manual processing of individual 3D models.
For example, for most of the above methods, a user needs to provide hand-crafted geometric features specific for their application.
Additionally, since such methods rely on part-aware geometric features, they require consistent part segmentation of 3D models within an object class. Acquiring such segmentation, especially for models from new object classes is often a manual and challenging step.
Even though recent deep learning methods have recently shown considerable promise in this area~\cite{Mo:2019ab}, automatically segmenting 3D models is still an active area of research.

\begin{figure}[t]
  \centering
  \includegraphics[width=0.49\textwidth]{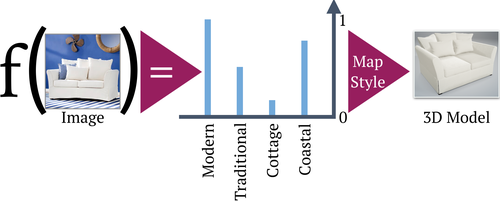}
  \caption{Our deep learning model learns furniture style from images. Given a context image with a target furniture piece (left), we predict its style embeddings (middle). We then map such embedding values to its corresponding, visually similar 3D model (right).}
\label{fig:embeddings}
\end{figure}

We propose to use deep neural networks to learn furniture style directly from annotated images reviewed by interior style experts.
Images include more style-relevant features, such as material, textures, and color~\cite{kilmer2014designing}, which are not embodied in current geometric approaches.
In this work, we demonstrate that such features are important for style compatibility.
Additionally, we show that our workflow for learning style from images is
fast and more flexible, since estimating the style of a new 3D
models translates into a single pass of its image through the neural
network (Figure~\ref{fig:embeddings}) as opposed to the geometric approaches where the processing itself could be time consuming.

\subsection{Image-based Style Estimation}
With the rise of deep neural networks, there have been numerous work on methods which rely on synthetic and real images to learn style~\cite{nikolenko2019synthetic}.
Such approaches usually formulate style compatibility as a metric learning problem~\cite{bellet2013survey} where they train a variation of a siamese network~\cite{bromley1994signature} to place stylistically compatible models close to each other in a embedding space. Using such a framework, Bell et al.~\cite{bell2015learning} detect visual similarity between product images. Li et al.~\cite{li2015joint} proposed to learn a joint embedding of images and 3D models for the task of image and shape based retrieval.
Lim et al.~\cite{lim2016identifying} employ deep metric learning to identify style using rendered views of 3D models. Specifically they use grayscale renderings of untextured objects and use triplet loss to cluster stylistically similar 3D models in a 512 dimensional embedding space.
Similarly, Liu et al.~\cite{liu2019learning} also embeds renderings of 3D models in high dimensional embedding spaces.
However, unlike~\cite{lim2016identifying}, they use textured renderings of isolated 3D models from the Unity Asset Store. They scrape approximately 4000 3D models from 121 model packages comprising of both indoor and outdoor 3D models and build their training data by assigning the same style label to all the models within the same package.

\begin{figure}[tb]
\centering
\includegraphics[height=1.48cm,keepaspectratio]{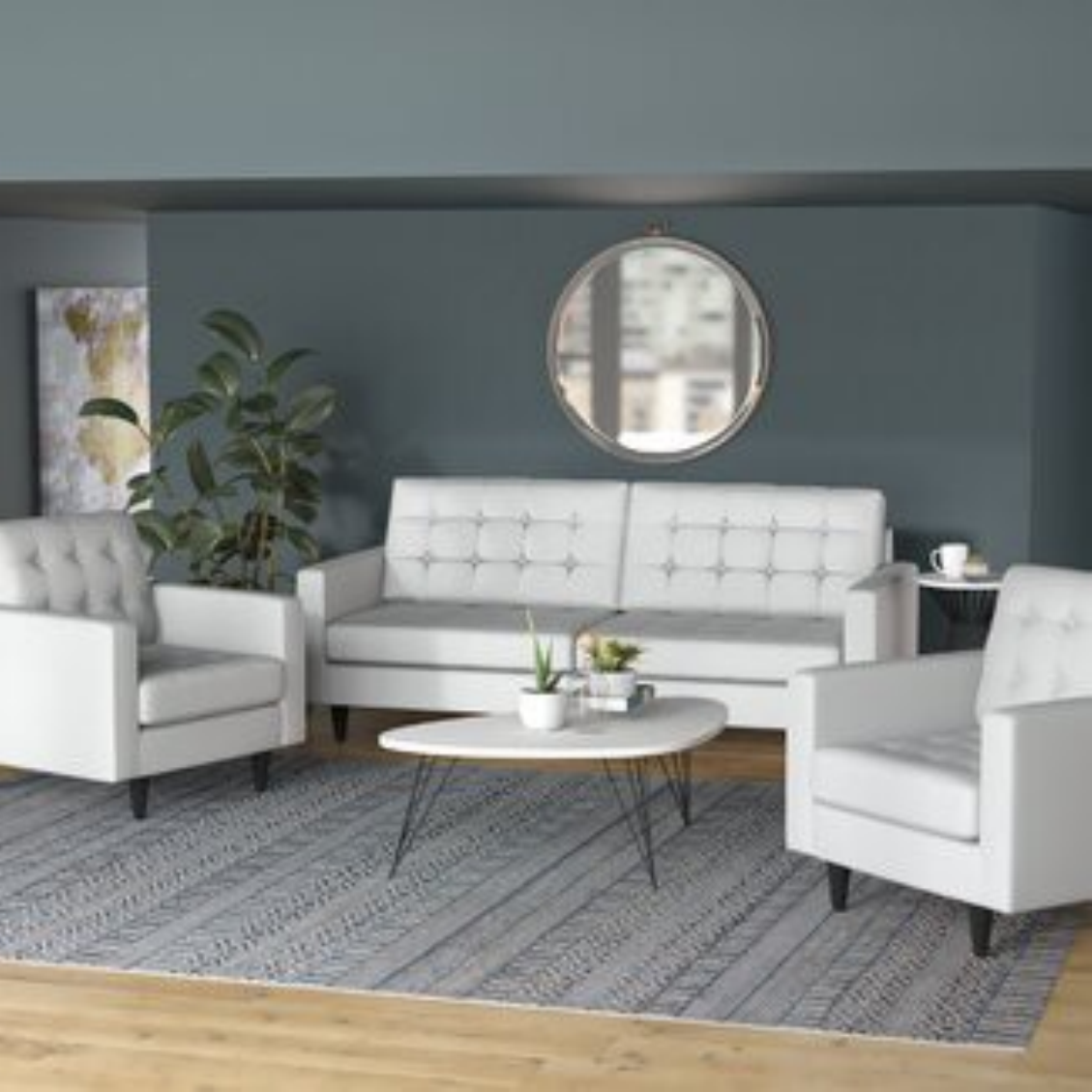}
\includegraphics[height=1.48cm,keepaspectratio]{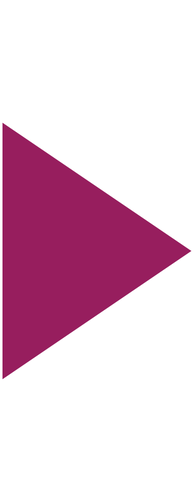}
\includegraphics[height=1.48cm,keepaspectratio]{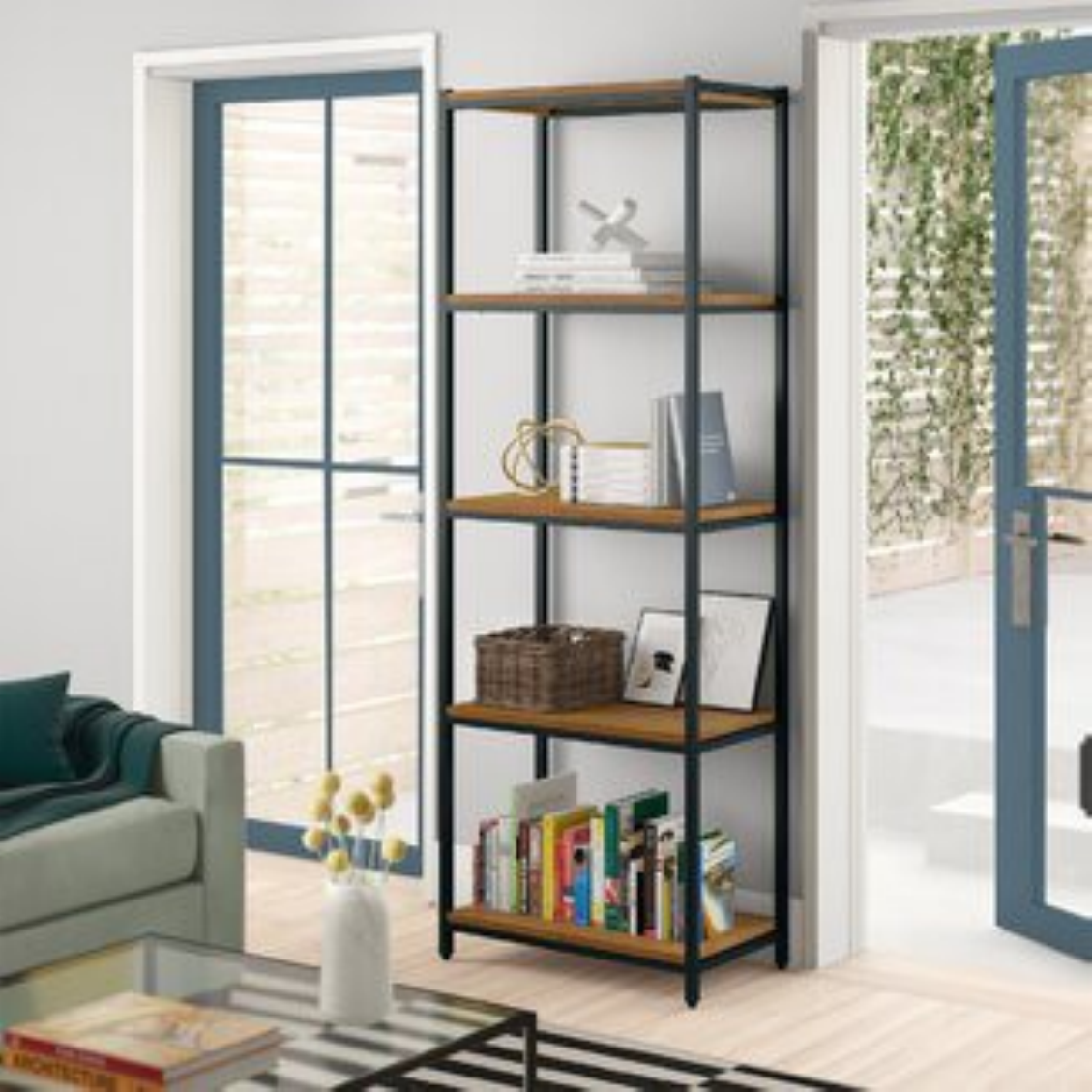}
\includegraphics[height=1.48cm,keepaspectratio]{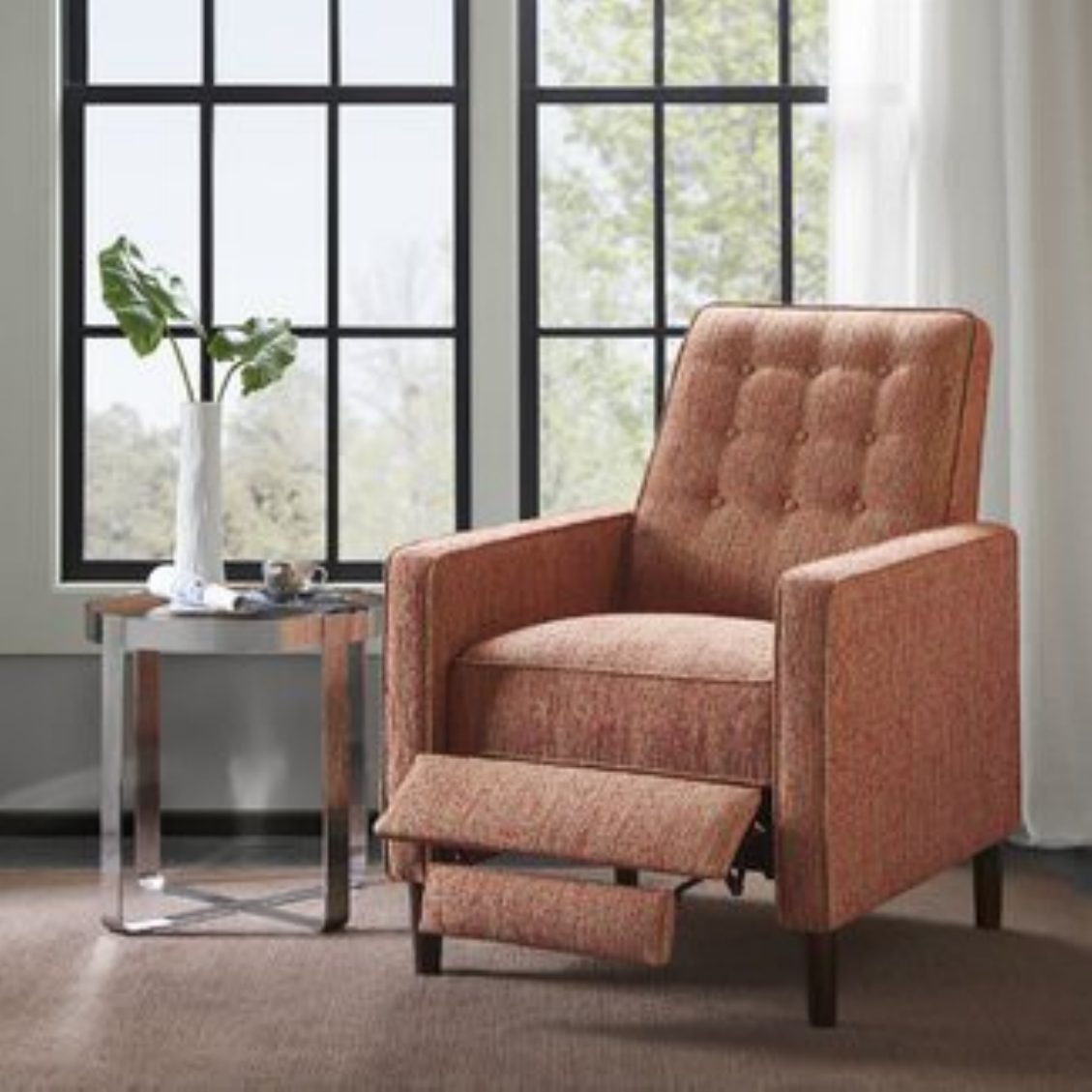}
\includegraphics[height=1.48cm,keepaspectratio]{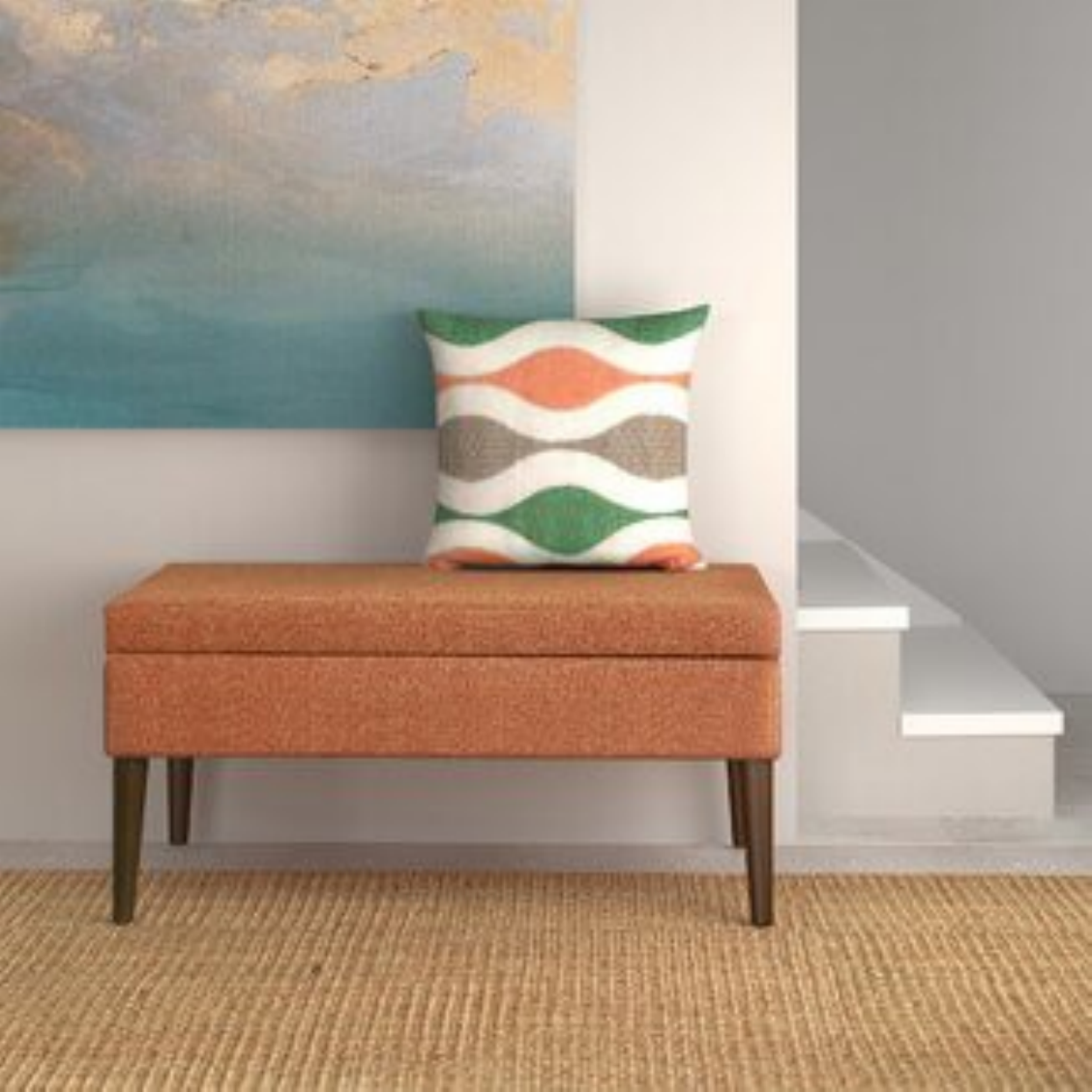}
\includegraphics[height=1.48cm,keepaspectratio]{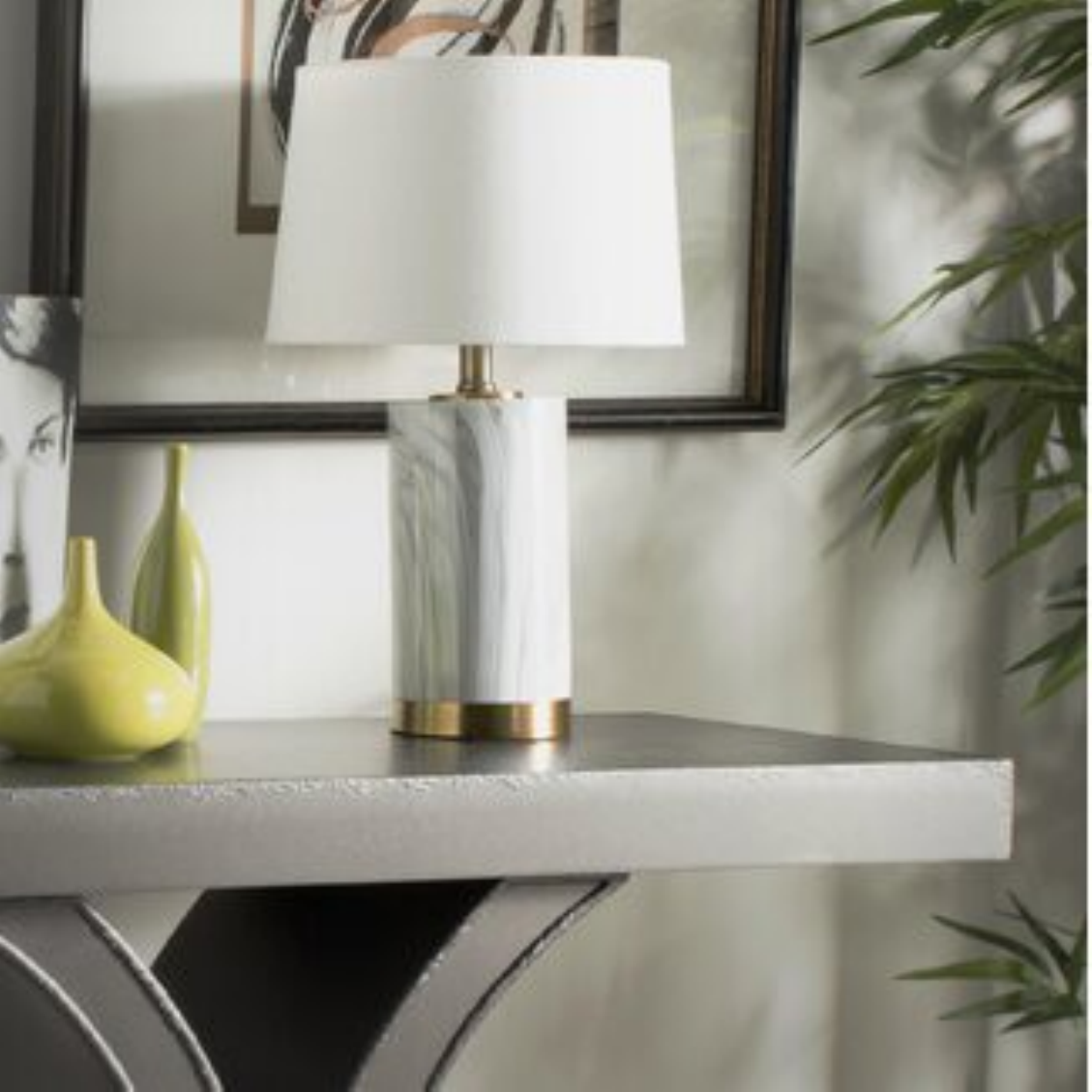}
\caption{Query furniture image and similar style furniture suggestions. Given the query furniture image on the left, our model suggests similar style furniture on the right.}
\label{fig:product_examples}
\end{figure}

Our work specifically focuses on indoor furniture style compatibility where we propose to use \emph{styled images} of furniture which are annotated by several design experts.
Such styled images not only describe the texture, color, and material of the 3D object, but also portray how it is used in the space (Figure~\ref{fig:product_examples}).
All elements combined play a vital role in determining the style~\cite{kilmer2014designing}.
Further, we propose to use comparison labels (Section~\ref{sec:comparison}) as opposed to discrete style labels to asses style compatibility.
Since design experts collectively often do not agree on a single style for a 3D object, we show that training with such comparison labels allows us to better learn the style spectrum of 3D models.
Lastly, unlike previous works, we use high quality real world images and 3D models for our style compatibility framework.

\section{Method}\label{sec:imagerec}
We use a deep neural network to estimate the style of images containing furniture. The input to our network are furniture images in a context scene.
Each image is annotated with style labels by multiple interior design experts (Section~\ref{sec:images}). Since style experts might disagree on an image's style,
we add \emph{comparison labels} to improve our network's accuracy (Section~\ref{sec:comparison}).
Image pairs, along with their style and comparison labels are then consumed
by our deep neural network (Section~\ref{sec:architecture}).
For such image pair, our deep learning model is trained to predict whether the first image shows more style-specific characteristics than the second image, e.g., is the first image more modern than the second one (Section~\ref{sec:loss}).
Based on our experiments (Section~\ref{sec:imageeval}),
we found that our training procedure (Section~\ref{sec:train}) enables us to learn a more accurate style spectrum compared to a discrete classification framework.
Using the trained style neural network, we extract image embeddings to calculate stylistic compatibility of furniture.

\begin{table}[b]
\vspace{-2mm}
\caption{Do experts agree on style?
To answer this question, we asked multiple participants to label each image.
The measure of agreement between participants is represented by the green
shaded areas.
Each cell's value is computed with:
$
1 - \tfrac{ || \{ x| x \in I \cap J \} || }{ || \{ x| x \in I \cup J \} || }
$,
where $I$ and $J$ are the set of images with style labels $i$ for rows and $j$ for columns.
Note that there is a considerable percentage of images that interior style experts have labeled as both Traditional-Modern, and Cottage-Traditional. Based on such subjectivity, we introduce comparison labels for improved style estimation (Section~\ref{sec:comparison}).} \label{fig:confusion}
\small

\begin{center}

\begin{tabular}{ c c c l }
Traditional  & Cottage  & Coastal  &  \shiftleft{8pt}{ \emph{Style} } \\

 \begin{tikzpicture}[yscale=0.16,xscale=0.36]
 \pie[hide number,square, rotate=90, text=inside, color={myRed,myGreen} ]
 {46/ ,  54/ }
 \draw (1.5, 0) node {54\%};
 \end{tikzpicture}
&

\begin{tikzpicture}[yscale=0.16,xscale=0.36]
\pie[hide number,square, rotate=90, text=inside, color={myRed,myGreen} ]
{27/ ,  73/ }
\draw (1.5, 0) node {73\%};
\end{tikzpicture}
&

\begin{tikzpicture}[yscale=0.16,xscale=0.36]
\pie[hide number,square, rotate=90, text=inside, color={myRed,myGreen} ]
{17/ ,  83/ }
\draw (1.5, 0) node {83\%};
\end{tikzpicture}

&
\shiftleft{10pt}{ \raisebox{13pt}{Modern} }
\\
&

\begin{tikzpicture}[yscale=0.16,xscale=0.36]
\pie[hide number,square, rotate=90, text=inside, color={myRed,myGreen} ]
{35/ ,  65/ }
\draw (1.5, 0) node {65\%};
\end{tikzpicture}

  &
  \begin{tikzpicture}[yscale=0.16,xscale=0.36]
  \pie[hide number,square, rotate=90, text=inside, color={myRed,myGreen} ]
  {13/ ,  87/ }
  \draw (1.5, 0) node {87\%};
  \end{tikzpicture}
  &

\shiftleft{10pt}{ \raisebox{13pt}{Traditional} }
 \\
  &   &
\begin{tikzpicture}[yscale=0.16,xscale=0.36]
\pie[hide number,square, rotate=90, text=inside, color={myRed,myGreen} ]
{18/ ,  82/ }
\draw (1.5, 0) node {82\%};
\end{tikzpicture}

  &
  \shiftleft{10pt}{ \raisebox{13pt}{Cottage} }
  \\

\end{tabular}
\end{center}
\end{table}

\subsection{Style Labels}\label{sec:images}
Each image in our data set is labeled with a style by $\noofexperts=10$ interior design experts. These styles include modern, traditional, cottage and coastal (Figure~\ref{fig:style_def}). Images are indexed by $\imgindex \in \imgset \equiv \{1,\ldots,\noofimg\}$ and experts by $\expertindex \in \expertset \equiv \{1,\ldots,\noofexperts\}$. Thus, given an image $\imgindex$, each expert $\expertindex$ provides a single style  label $\stylelabel_{\imgindex}^{\expertindex} \in \styleset \equiv \{1,\ldots,\noofstyles\}$, where $\styleset$ consists of the
four major style labels ($\noofstyles=4$). We denote the set of all such labels by $\allstylelabels \equiv \{(\imgindex, \expertindex, \stylelabel_{\imgindex}^{\expertindex})\}$.

\subsection{Comparison Labels}\label{sec:comparison}
Style assessment is subjective, even among interior style experts as shown in Table~\ref{fig:confusion}.
This ambiguity introduces noise and inaccuracies for learning style directly from such style labels (Section~\ref{sec:imageeval}).
Hence, we tackle the problem of noisy style labels by introducing \emph{comparisons} into training. A comparison label indicates the \emph{relative order} between a pair of data samples~\cite{yildiz2019classification}.
For each style and each pair of images, we generate a comparison label with respect to the relative order of the number of style labels each image receives. Formally, given a style $\styleindex \in \styleset$, we consider all pairs of images ($\imgindexone$, $\imgindextwo$), where $\imgindexone$, $\imgindextwo \in \imgset$, that receive at least one label for style $\styleindex$, i.e., ($\imgindexone$ $\imgindextwo$) for which there exists $\expertindexone$ such that $\stylelabel_{\imgindexone}^{\expertindexone} = \styleindex$ \emph{and} $\expertindextwo$
such that $\stylelabel_{\imgindextwo}^{\expertindextwo} = \styleindex$.
For each image pair ($\imgindexone$, $\imgindextwo$), the comparison label is:

\[
    \stylelabel_{(\imgindexone, \imgindextwo)}^{\styleindex} =
\begin{dcases}
    +1 , & \text{if } \sum_{\expertindex \in \expertset} \big(\indicator(\stylelabel_{\imgindexone}^{\expertindex} = \styleindex) - \indicator(\stylelabel_{\imgindextwo}^{\expertindex} = \styleindex)\big) > \compthr \\
    -1 , & \text{if } \sum_{\expertindex \in \expertset} \big(\indicator(\stylelabel_{\imgindextwo}^{\expertindex} = \styleindex) - \indicator(\stylelabel_{\imgindexone}^{\expertindex} = \styleindex)\big) > \compthr
\end{dcases}
\]

In other words, $\stylelabel_{(\imgindexone, \imgindextwo)}^{\styleindex} = +1$ indicates that image $\imgindexone$ is
ranked higher than image $\imgindextwo$ with respect to style $\styleindex$, and $\stylelabel_{(\imgindexone, \imgindextwo)}^{\styleindex} = -1$, otherwise. We discard pairs that do not satisfy either one of these conditions. The label threshold $\compthr$ controls the trade-off between the noise and additional information introduced by the comparison labels.
We repeat this process for all styles $\styleindex \in \styleset$ and obtain the comparison label set $\allcomplabels \equiv \{(\imgindexone, \imgindextwo, \styleindex, \stylelabel_{(\imgindexone, \imgindextwo)}^{\styleindex})\}$, where $\imgindexone$ and $\imgindextwo \in \imgset$ and $\styleindex \in \styleset$.

\begin{table}[t!]
    \caption{
    Style predictions from our method.
    Given images on the left, our model estimates their style.
    P denotes the Prediction of our network, and GT denotes Ground Truth labels given by $10$ style experts.
    Note that in case of interior experts disagreeing on style,
    which manifests with mixed GT style labels, our model estimates that images belong to the same style subsets (Section~\ref{sec:imageeval}).
    }\label{tbl:predictions}
    \vspace{-1 em}
 \begin{center}
 \begin{tabular}{p{1.29in} !{\color{amethyst}\vrule} r !{\color{amethyst}\vrule} r !{\color{amethyst}\vrule} l !{\color{amethyst}\vrule} }
\arrayrulecolor{amethyst}\cline{2-4}
   & \emph{P} & \emph{GT} &  \emph{Style} \\
 \hline\hline
\multirow{4}{*}{
\includegraphics[trim=0 40 0 40, clip, width=1.3in]{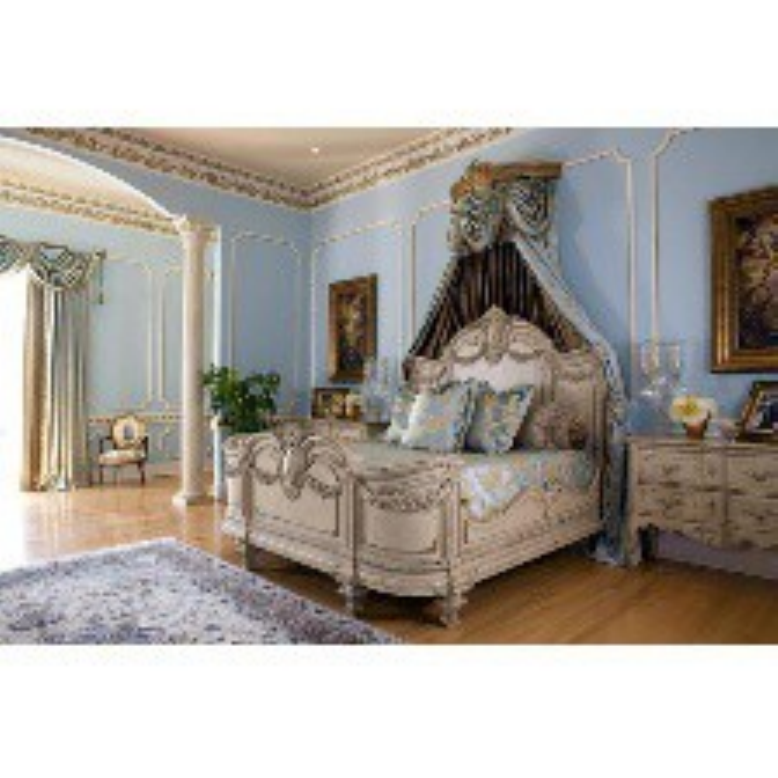}
} & 0 & 0 & Modern  \\[1.05ex]
\cline{2-4}
& 1.0 & 10 & Traditional \\[1.05ex]
\cline{2-4}
& 0 & 0 & Cottage \\[1.05ex]
 \cline{2-4}
& 0 & 0 & Coastal \\[1.05ex]
 \hline  \hline
 \multirow{4}{*}{
 \includegraphics[trim=0 40 0 40, clip, width=1.3in]{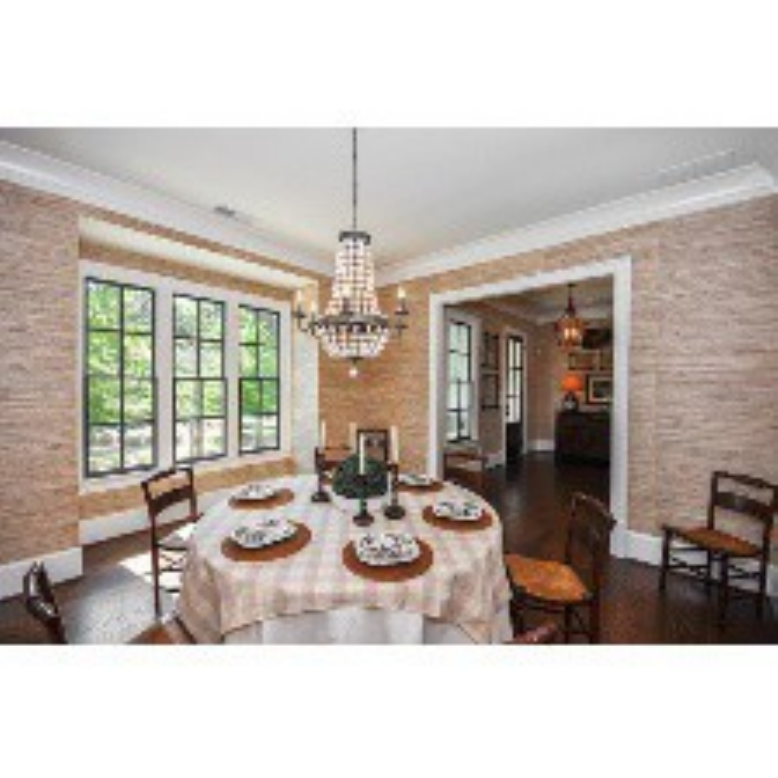}
 } & 0 & 0 & Modern  \\[1.05ex]
 \cline{2-4}
 & 0.34 & 6 & Traditional \\[1.05ex]
 \cline{2-4}
 & 0.63 & 4 & Cottage \\[1.05ex]
  \cline{2-4}
 & 0.03 & 0 & Coastal \\[1.05ex]
  \hline  \hline
 \multirow{4}{*}{
 \includegraphics[trim=0 40 0 40, clip, width=1.3in]{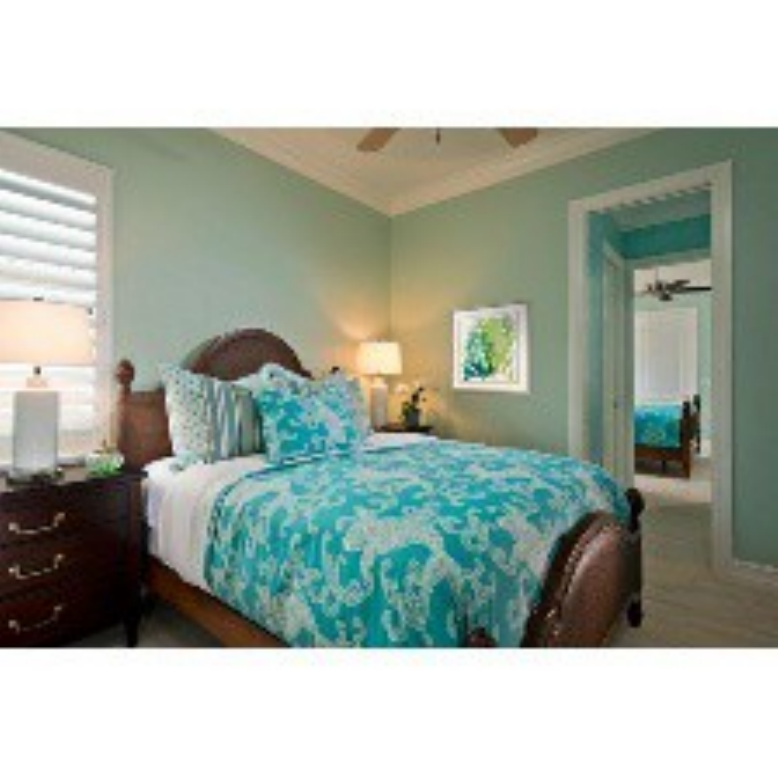}
 } & 0 & 0  & Modern \\[1.05ex]
 \cline{2-4}
 & 0 & 2 & Traditional \\[1.05ex]
 \cline{2-4}
 & 0 & 0 & Cottage \\[1.05ex]
 \cline{2-4}
 & 1.0 & 8 & Coastal \\[1.05ex]
 \hline
\end{tabular}
\end{center}
\end{table}

\subsection{Network Architecture}\label{sec:architecture}
Our network architecture is inspired by the siamese network~\cite{bromley1994signature}. A classical siamese network comprises of two base networks which share their weights, followed by a layer regressing the similarity between the network inputs.
The parameters between the twin networks are tied.
Such relationship gurantees that for an input of two similar images, the networks would map the images close to each other in feature space, since each network computes the same function. Hence, siamese networks are useful for ranking similarity between images~\cite{koch2015siamese}.

We use VGG16~\cite{simonyan2014very} as our base network $\basenet$ and replace its last layer with two fully connected layers: a $16$-dimensional layer with ReLU activation and an $\noofstyles=4$-dimensional layer with softmax activation. This allows us to associate each input image $\imgindex$ to the base network $\basenet$ with a $\nooffeat=16$-dimensional feature vector, $ \feature_{\imgindex} \in \realnumbers^{\nooffeat}$.

\subsection{Loss Function}\label{sec:loss}
We extend the generic application of siamese networks to learn from comparison labels.
Comparisons are regressed via the siamese architecture, consisting of two identical base networks $\basenet(\cdot;  \weights): \realnumbers^{\nooffeat} \rightarrow \realnumbers^{\noofstyles}$, parameterized by the weight matrix $ \weights$.
$\basenet(\cdot;  \weights)$ receives a image with feature vector $\feature_{\imgindex}  \in \realnumbers^{\nooffeat}$ and regresses the style label predictions $\basenet( \feature_{\imgindexone};  \weights)^{\styleindex}$, $\styleindex \in \styleset$. The siamese architecture receives a pair of images with feature vectors $\feature_{\imgindexone} \in \realnumbers^{\nooffeat}$
and $ \feature_{\imgindextwo} \in \realnumbers^{\nooffeat}$ and predicts the comparison labels $\hat \stylelabel_{(\imgindexone, \imgindextwo)}^{\styleindex}$, $\styleindex \in \styleset$, governed by the relative order between style predictions.
For all $\{(\imgindexone, \imgindextwo, \styleindex, \stylelabel_{(\imgindexone, \imgindextwo)}^{\styleindex})\} \in \allcomplabels$, the comparison label prediction is given by:
\begin{align}
    \hat \stylelabel_{(\imgindexone, \imgindextwo)}^{\styleindex} = \basenet( \feature_{\imgindexone};  \weights)^{\styleindex} - \basenet( \feature_{\imgindextwo};  \weights)^{\styleindex}
\label{equ:comp_pred}
\end{align}
where superscript $\styleindex \in \styleset$ indicates the $\styleindex$-th element in their softmax prediction. Given these predictions, we learn $\basenet(\cdot;  \weights)$ via the following optimization problem:
\begin{equation}
\begin{aligned}
\label{equ:masterequation}
& \underset{ \weights}{\text{min}}
\!\!\!\!\!\! \sum_{\substack{\{(\imgindexone, \imgindextwo, \styleindex, \stylelabel_{(\imgindexone, \imgindextwo)}^{\styleindex})\} \in \allcomplabels}}\!\!\!\!\!\!\!\! \loss(\stylelabel_{(\imgindexone, \imgindextwo)}^{\styleindex}, \hat \stylelabel_{(\imgindexone, \imgindextwo)}^{\styleindex})
\end{aligned}
\end{equation}

For comparison loss $\loss$,
we extend and reparametrize the Bradley-Terry model~\cite{bradley1952rank}, which considers relative order between input pairs and is used for modeling comparisons.
The Bradley-Terry model assumes that for each image $\imgindex \in \imgset$ and style $\styleindex \in \styleset$, there exists a latent score $\score_{\imgindex}^{\styleindex} \in [0,1]$ governing all comparison events in $\allcomplabels$ involving style $\styleindex$.
Comparisons in $\allcomplabels$ are independent, with marginal probabilities given by:
\begin{align}
\label{equ:comp}
    P(\stylelabel_{(\imgindexone, \imgindextwo)}^{\styleindex} = +1) = \frac{\score_{\imgindexone}^{\styleindex}}{\score_{\imgindexone}^{\styleindex} + \score_{\imgindextwo}^{\styleindex}}, & & \forall (\imgindexone, \imgindextwo, \styleindex, \stylelabel_{(\imgindexone, \imgindextwo)}^{\styleindex}) \in \allcomplabels.
\end{align}
We reparametrize the Bradley-Terry model to regress scores $\score_{\imgindex}^{\styleindex}$, $\styleindex \in \styleset$ as functions of image features $\feature_{\imgindex}$. Hence, $\score_{\imgindex}^{\styleindex}$, $\styleindex \in \styleset$
correspond to the style predictions from $\basenet$, i.e., $\score_{\imgindex}^{\styleindex} = e^{\basenet( \feature_{\imgindex};  \weights)^{\styleindex}}$, $\imgindex \in \imgset$, $\styleindex \in \styleset$. Then, given an image pair $(\imgindexone, \imgindextwo)$, $\imgindexone$, $\imgindextwo \in \imgset$ and the corresponding comparison
labels $\stylelabel_{(\imgindexone, \imgindextwo)}^{\styleindex}$, $\styleindex \in \styleset$, the comparison loss $\loss$ is the negative log-likelihood under the Bradley-Terry model:
\begin{align}
\loss(\stylelabel_{(\imgindexone, \imgindextwo)}^{\styleindex}, \hat \stylelabel_{(\imgindexone, \imgindextwo)}^{\styleindex}) &= \log(1\!+\!e^{-\stylelabel_{(\imgindexone, \imgindextwo)}^{\styleindex} \hat \stylelabel_{(\imgindexone, \imgindextwo)}^{\styleindex}}),
\label{equ:BTLoss}
\end{align}
where $\hat \stylelabel_{(\imgindexone, \imgindextwo)}^{\styleindex}$ is given by Eq.~\eqref{equ:comp_pred}.

\subsection{Training \& Implementation Details}\label{sec:train}

\emph{Image Dataset}.
We use images gathered from multiple sources:
(i) Staging a scene in a physical room, and then capturing it with a camera, (ii) Captured from a virtual scene, curated by a 3D artist, and
(iii) Scraped from web and 3rd parties.
Due to noisy image metadata, an image's origin might not always be known. Our data set contains about $\noofimg = 672K$ images,
depicting $20$ different furniture categories, such as sofas, coffee tables, and dining tables. Images depict furniture in a styled room, (Figure~\ref{fig:product_examples}), and most show a wide view, providing ample style context (Table~\ref{tbl:predictions}).
We reserve 80\% of $\noofimg$ images for training, 10\% for validation, and 10\% for testing. Images in the training set are not paired with images in the validation or test sets for generating comparison labels.
We also resize and apply white padding to all images, so that all images $\in \mathbb{R}^{224\times224\times3}$.

\emph{Validation Dataset}.
To tune the hyperparameters of our style estimation model, we create a validation data set of clean style labels. For each style $\styleindex \in \styleset$, we only consider images that receive at least $\minvote$ labels, i.e., image $\imgindex$ is in style category $\styleindex$ if $\sum_{\expertindex \in \expertset} \indicator(\stylelabel_{\imgindex}^{\expertindex} = \styleindex) \geq \minvote$.
We obtain two style validation data sets: (i) set $\allstylelabels$ containing all images with style labels $\minvote = 1$ for all $\styleindex \in \styleset$, and (ii) set $\allcleanstylelabels$ containing only images with high agreement style labels of at least $\minvote=10$ for modern, $\minvote=8$ for traditional, and $\minvote=7$ for coastal and cottage. Such thresholds are selected empirically considering number of samples per style category and amount of labeling noise introduced. As a result, each image in set $\allcleanstylelabels$ is associated with a single ground-truth that has the maximum number of style labels, whereas in set $\allstylelabels$, the same image can appear with different style labels due to expert disagreement.

\emph{Training.}
We train our siamese architecture on the comparison label set $\allcomplabels$.
We employ transfer learning and initialize VGG16 with weights pre-trained on the Places365 data set~\cite{zhou2017places} for place classification.
After initialization, we freeze the weights of the first $21$ layers and fine tune the weights of the last $3$ fully connected layers, comprising $16.8\text{M}$ parameters.
We use RMSProp optimizer~\cite{tieleman2012lecture} with learning rate $0.0001$, to optimize $\loss$ in Eq.~\eqref{equ:masterequation}. We add L2 regularizers with a regularization parameter $\lambda$ to all layers.
We choose the regularization parameter $\lambda$, comparison label threshold $\compthr$, and number of training comparisons $\noofcomp$ w.r.t. the classification performance on the validation set, via a grid search.
For each combination of $\lambda \in [0.002,0.0002,0.00002]$, $\compthr \in [1,2,3]$, and $\noofcomp \in [500\text{K}, 1\text{M}, 2\text{M}, 3\text{M}, 4\text{M}, 5\text{M}, 10\text{M}, 15\text{M}]$,
we train the style estimation model on $\noofcomp$ comparisons selected uniformly at random over the training set of $\allcomplabels$.
Finally, we evaluate the resulting models to determine optimal parameters for predicting style, in which we found $\compthr=3$ to be experimentally optimal.

Our machine learning model was implemented using Tensorflow.
Training of our style-learning model took about 5 hours on a NVIDIA Tesla V100 GPU'\@.' Below we evaluate our network both qualitatively and quantitatively, w.r.t.~several metrics.

\begin{figure}[t!]
\centering
\raisebox{10mm}{\parbox[b]{2mm}{\rotatebox[origin=b]{90}{\emph{Images}}}}
\begin{subfigure}[b]{0.24\columnwidth}
\tcbox[size=fbox,colframe=white!60!black,
            colback=white!60]{
            \includegraphics[height=1.84cm,keepaspectratio]{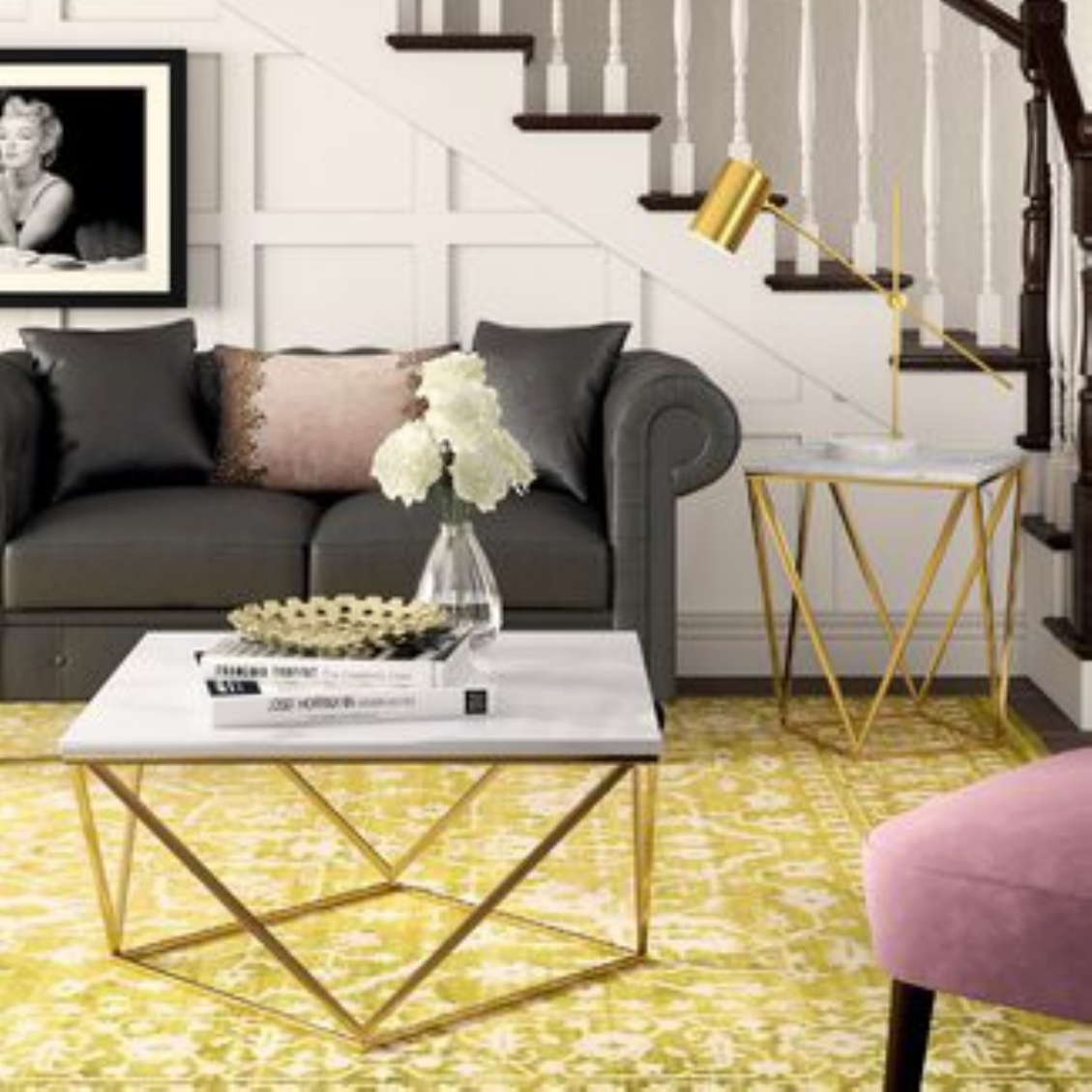}

}
\end{subfigure}
\begin{subfigure}[b]{0.47\columnwidth}
\tcbox[size=fbox,colframe=amethyst!60!black,
          colback=amethyst!60]{
          \includegraphics[height=1.84cm,keepaspectratio]{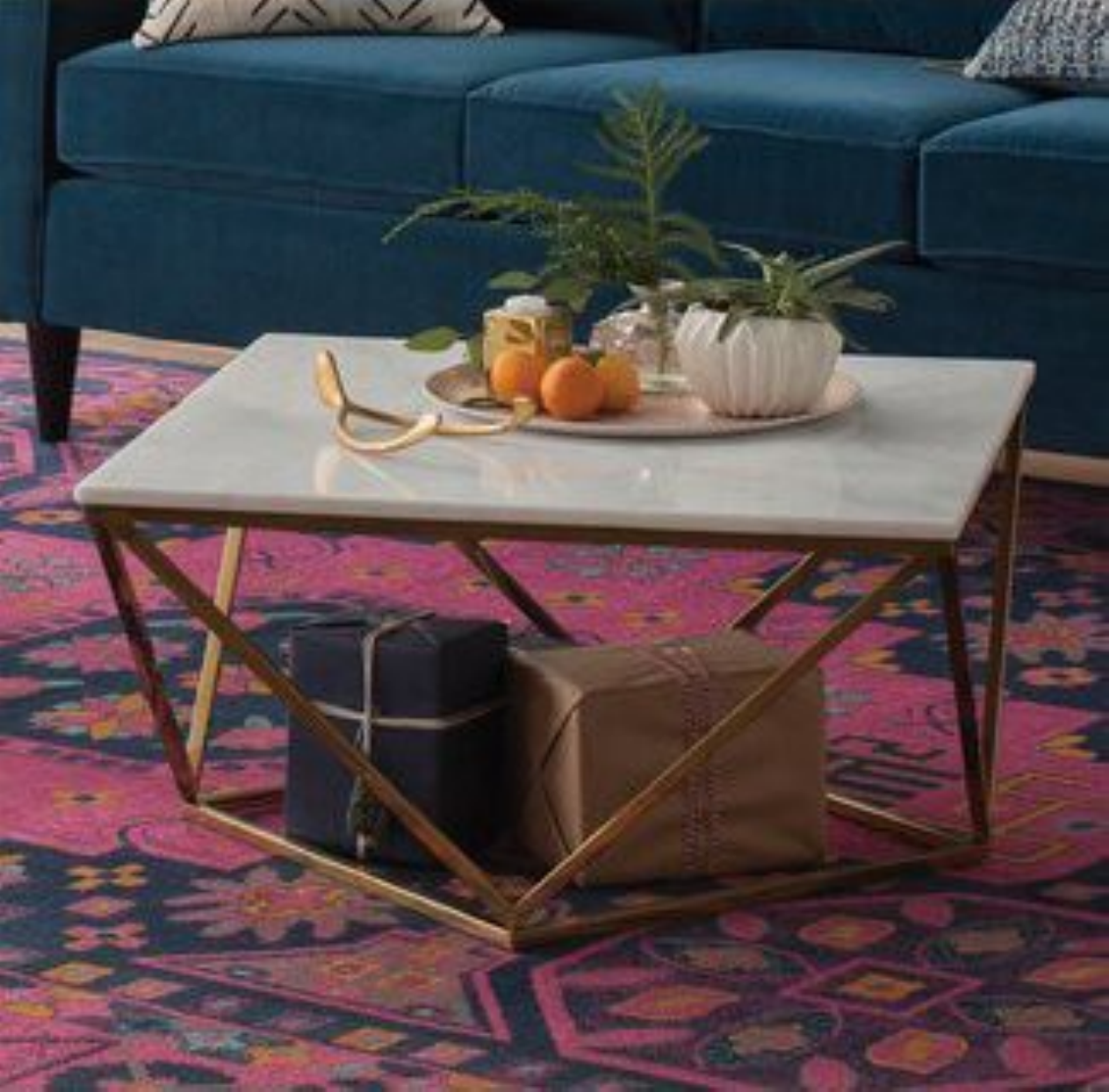}
          \includegraphics[height=1.84cm,keepaspectratio]{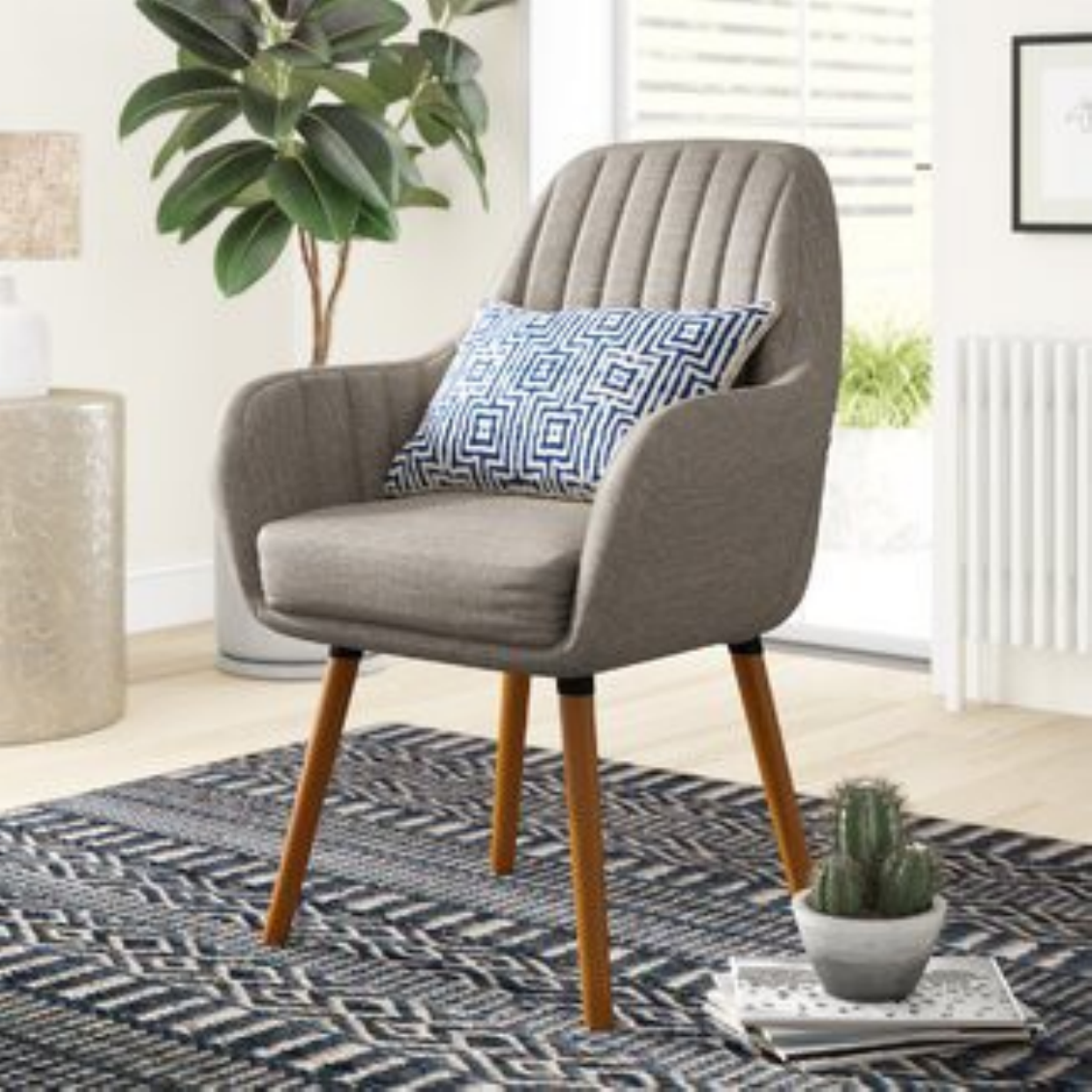}
}
\end{subfigure}
\begin{subfigure}[b]{0.24\columnwidth}
\tcbox[size=fbox,colframe=white!60!black,
              colback=white!60]{
              \includegraphics[height=1.84cm,keepaspectratio]{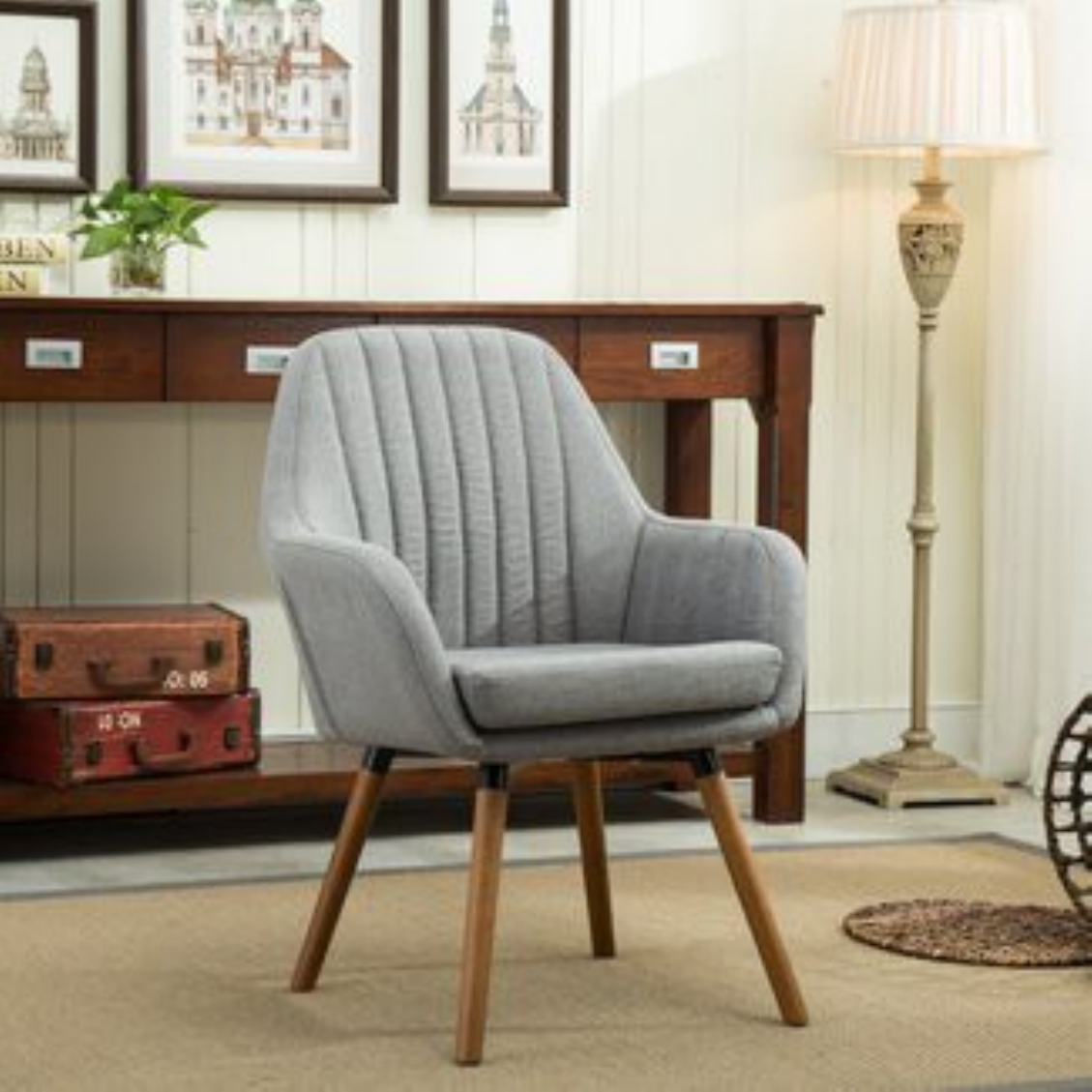}
}
\end{subfigure}
\\
\raisebox{8mm}{\parbox[b]{2mm}{\rotatebox[origin=b]{90}{\emph{3D Models}}}}
\includegraphics[height=1.58cm,keepaspectratio]{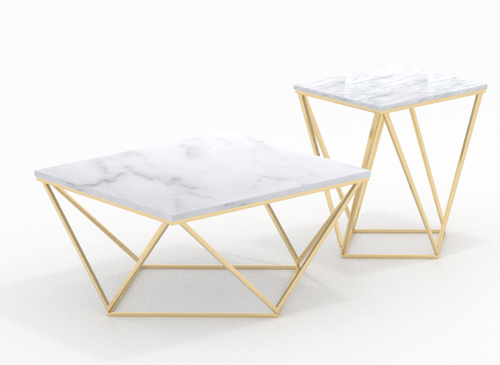}%
\hfill%
\includegraphics[height=1.58cm,keepaspectratio]{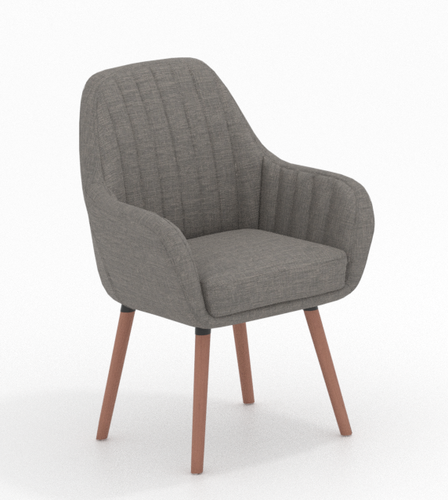}%
\caption{We use images of curated interiors to estimate a 3D models style.
In this example, we are interested to estimate the style compatibility between the coffee table set on the bottom left, to accent chair on the bottom right.
We use images on the top to estimate such compatibility.
Images framed in purple are the closest style-wise, according to the embeddings distance (Eq.~\ref{eq:furnituredistance}).}
\label{fig:3dmodeldistance}
\end{figure}

\begin{figure}[bt]
\centering
\includegraphics[width=0.31\columnwidth,keepaspectratio]{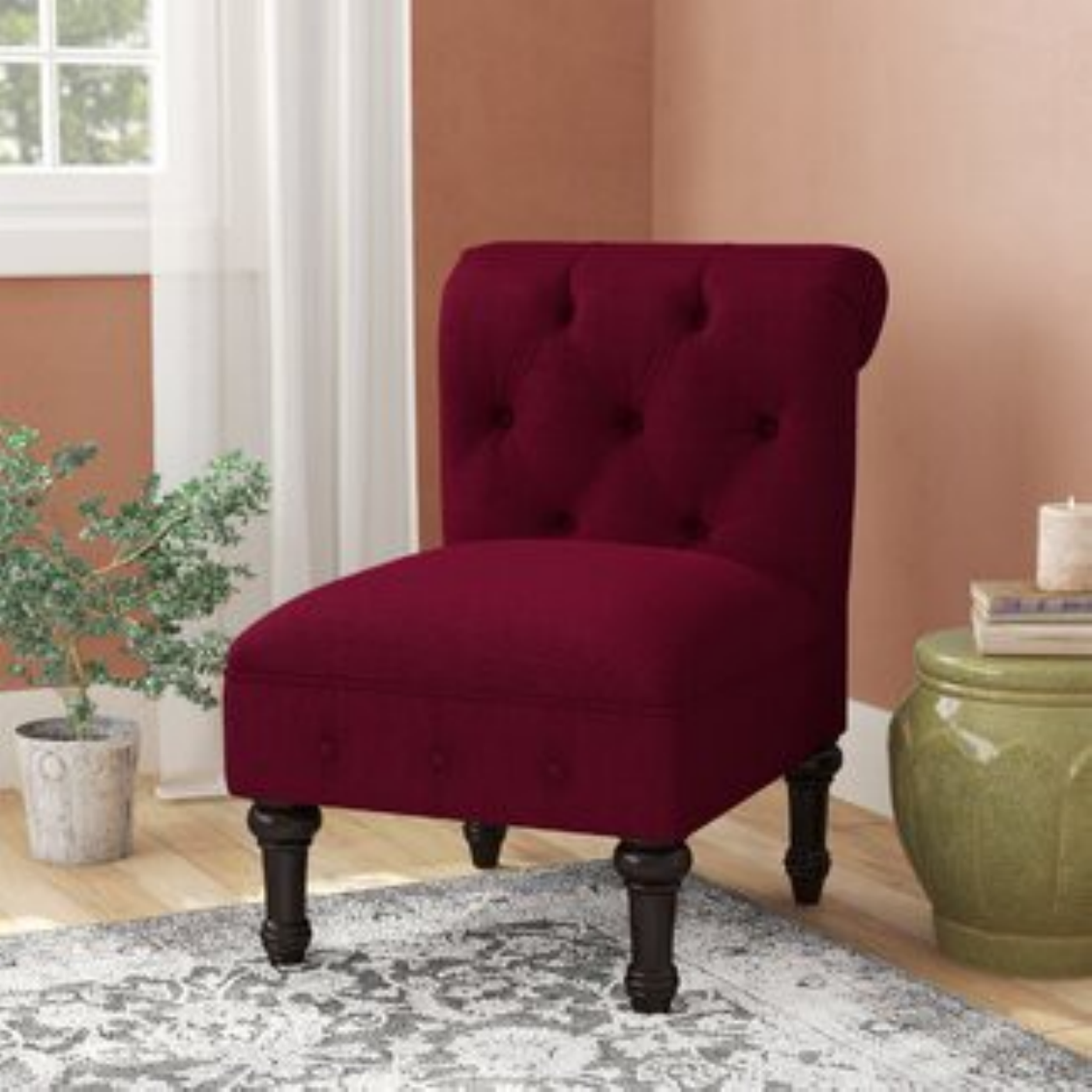}
\includegraphics[width=0.31\columnwidth,keepaspectratio]{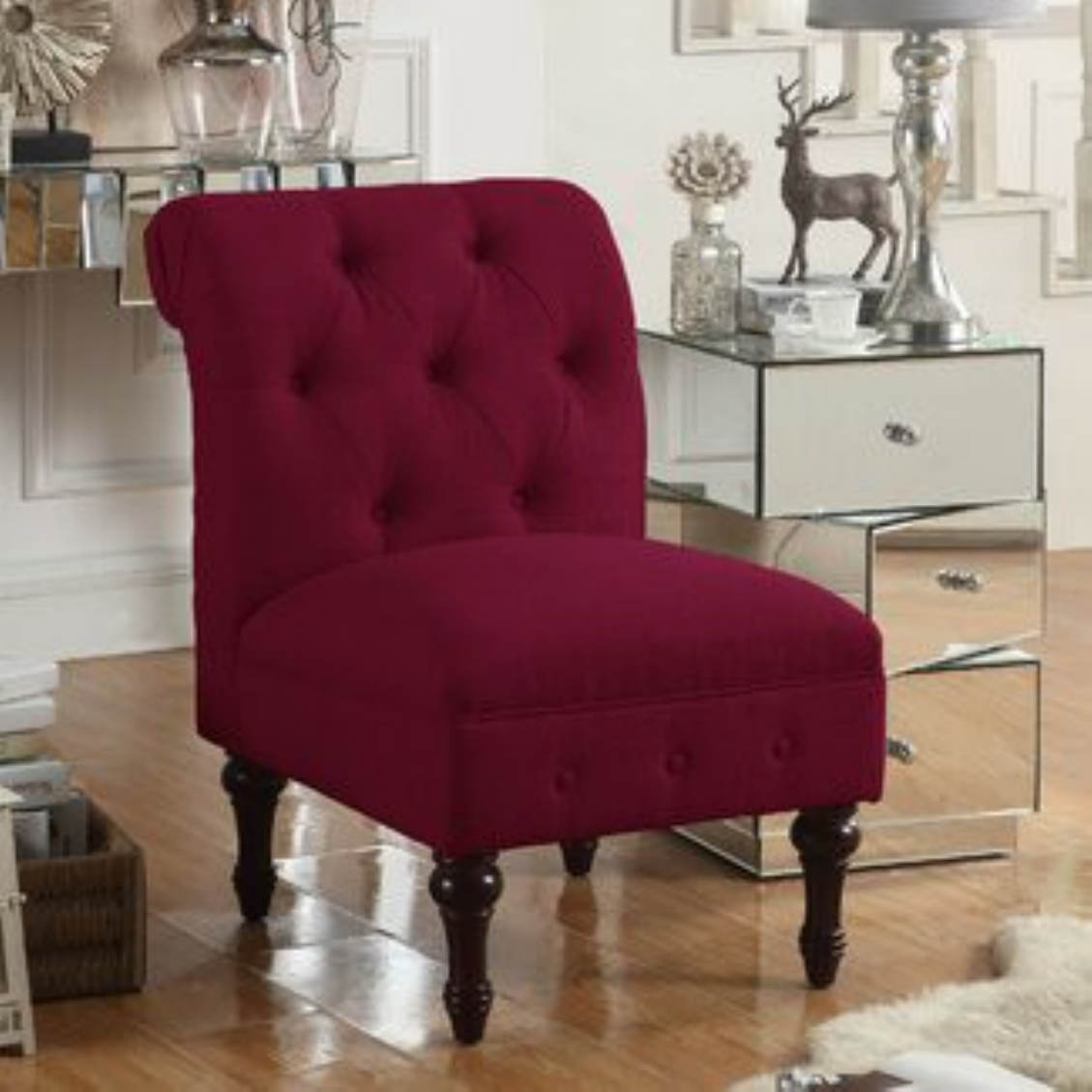}
\includegraphics[width=0.31\columnwidth,keepaspectratio]{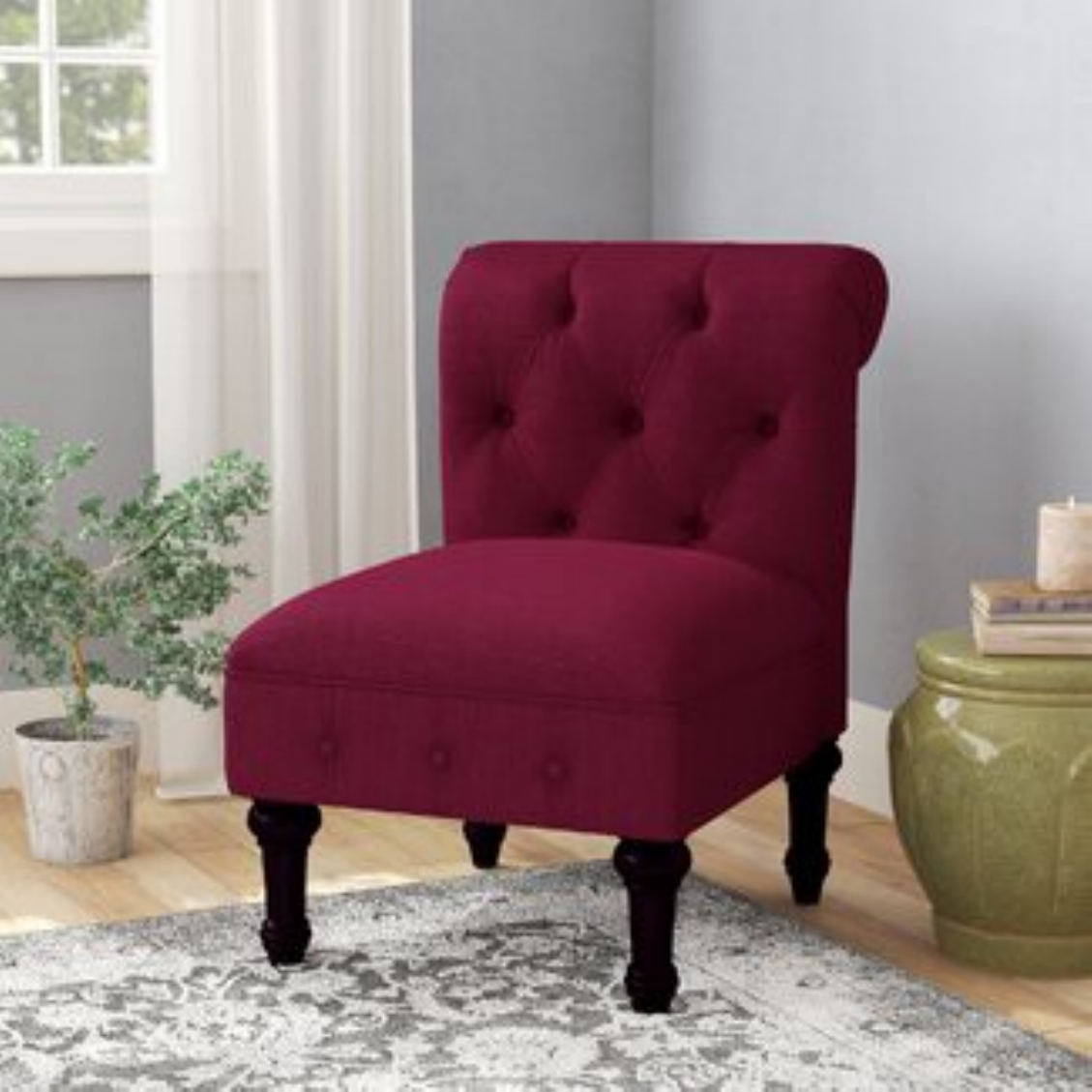}
\caption{A furniture piece, such as the accent chair, can appear in multiple styled scenes.
We consider all associated images when searching for other style-complementary furniture.}
\label{fig:furnitureimages}
\end{figure}

\begin{figure*}[t!]
\caption{Learning from style comparison labels improve accuracy
over non-comparison deep learning networks.
We compare our model trained on comparison labels with two baseline models, one trained on all images ($\allstylelabels$) and second trained on clean images ($\allcleanstylelabels$).
Classification accuracy increased by $24\%$ on cottage, $45\%$ on coastal, and $5\%$ on average over styles.
Such results validate that comparisons labels reveal more information than style labels, even with the challenges of class imbalance and label noise.
Note that the poor performance of Cottage and Coastal is due to class imbalance.
In our data set, the number of images with at least 1 label for each style is about 500k for modern, 400k for traditional, 300k for cottage and 150k for coastal.
}
\centering
\begin{subfigure}{0.33\textwidth}
    \centering
    \includegraphics[width=1.0\linewidth]{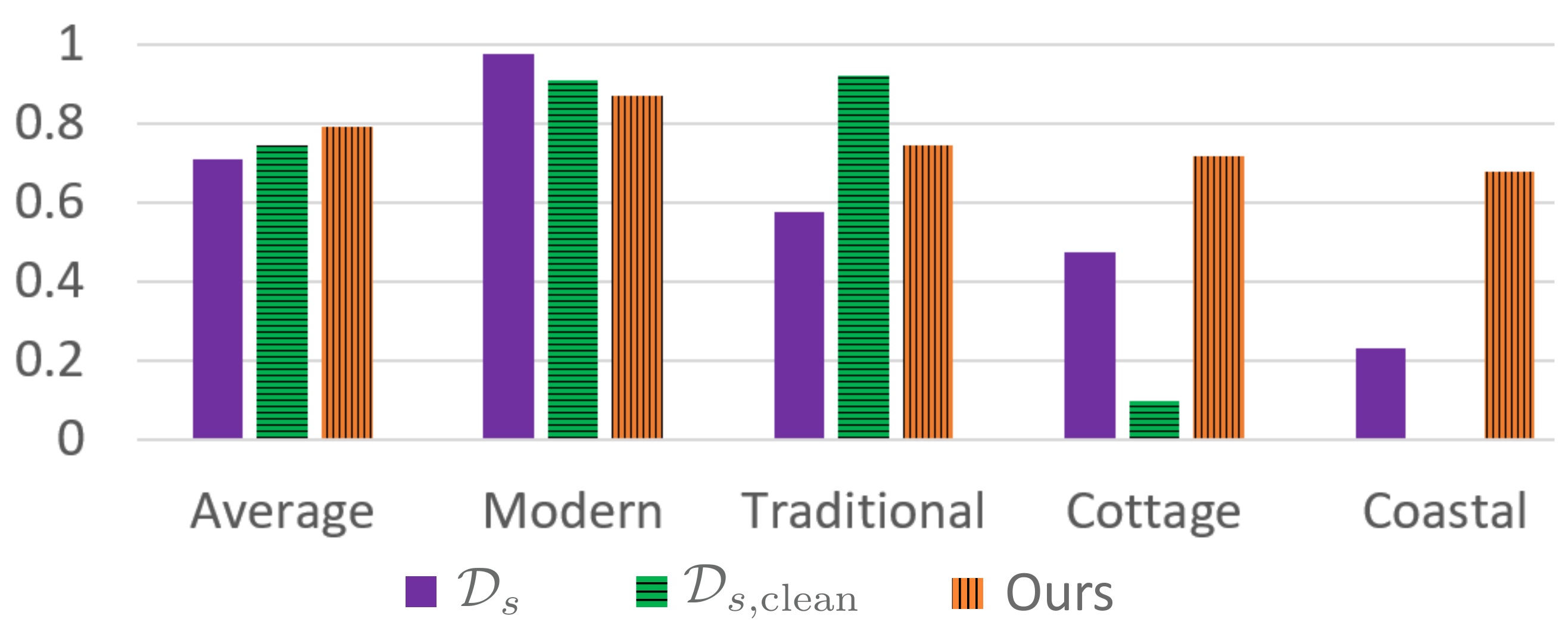}
    \caption{Classification Accuracy}
    \label{fig:classification_v1_vs_v2_4class}
\end{subfigure}%
\begin{subfigure}{0.33\textwidth}
    \centering
    \includegraphics[width=1.0\linewidth]{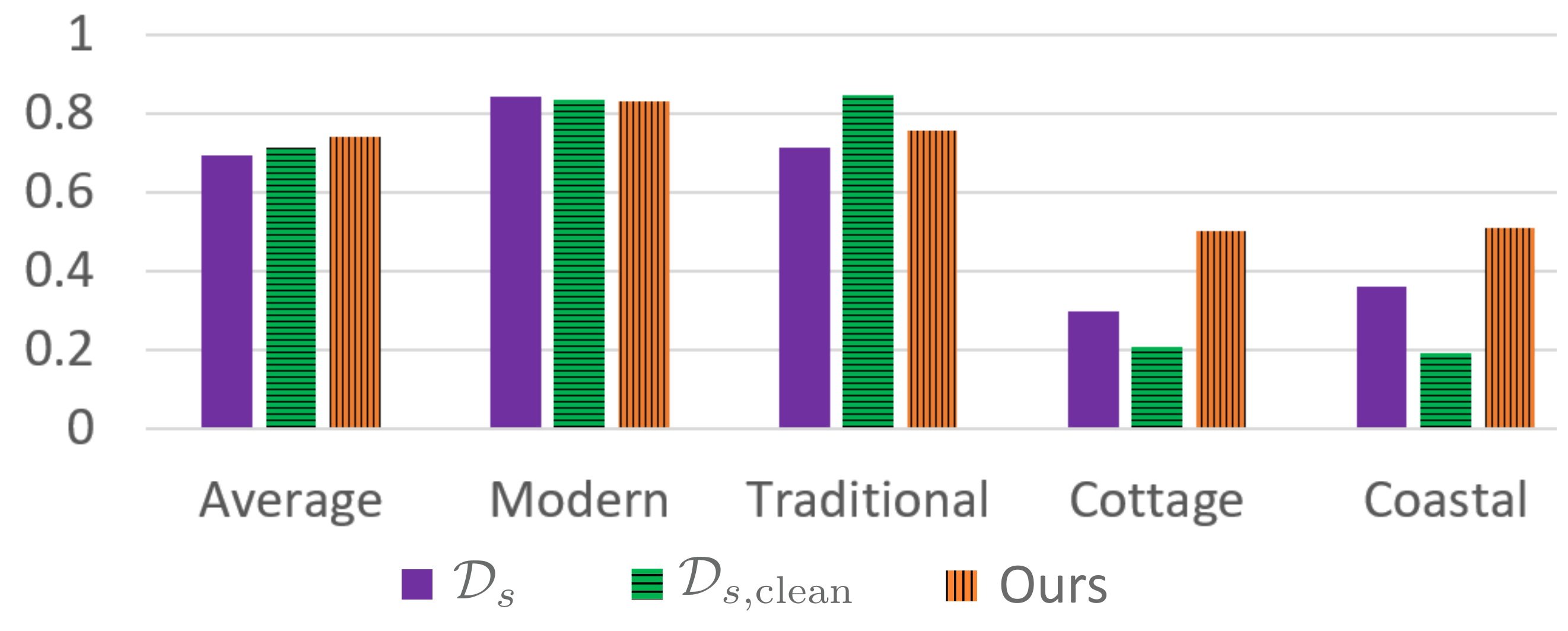}
    \caption{Recall rate at $k$=1}
    \label{fig:retrieval1_v1_vs_v2_4class}
\end{subfigure}%
\begin{subfigure}{0.33\textwidth}
    \centering
    \includegraphics[width=1.0\linewidth]{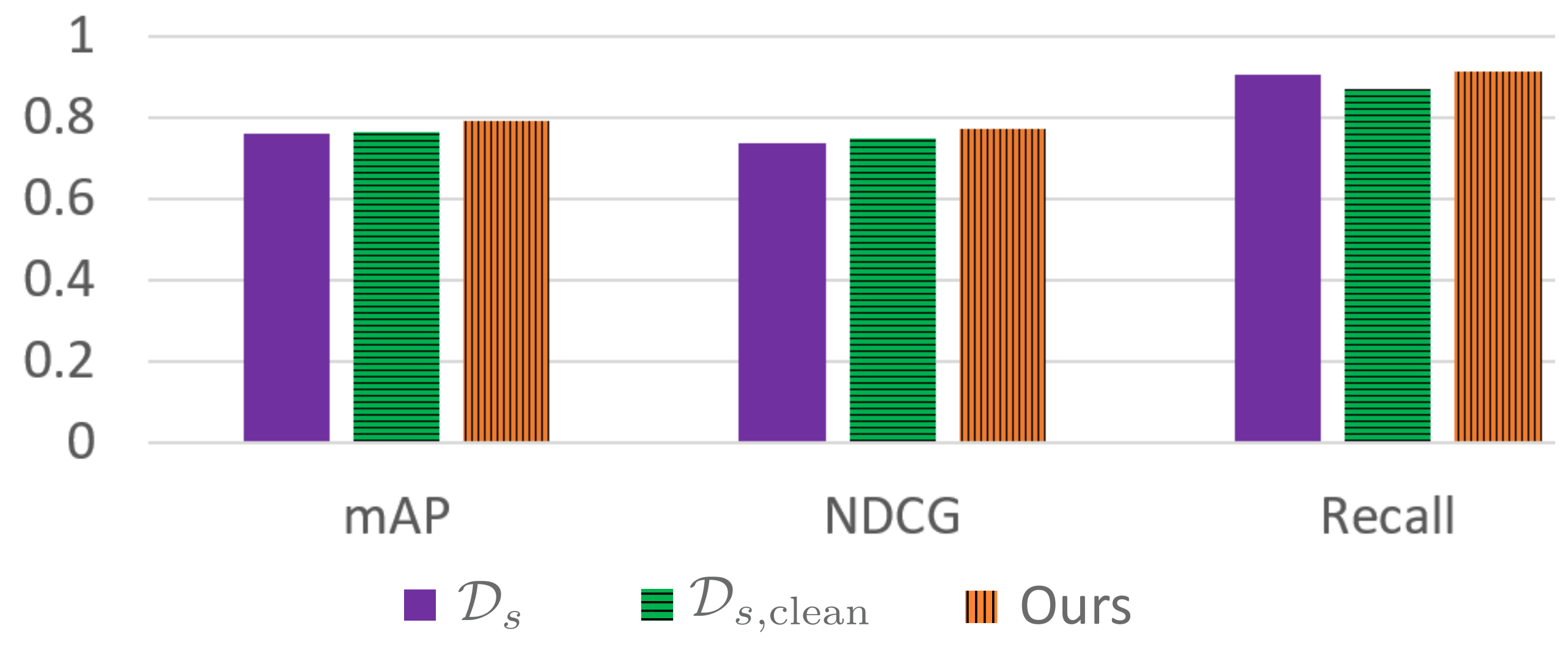}
    \caption{Retrieval at $k$=5}
    \label{fig:retrieval5_v1_vs_v2_4class}
\end{subfigure}%
\label{fig:v1_vs_v2_4class}
\end{figure*}

\begin{figure}[bt]
\captionsetup[subfigure]{labelformat=empty,textfont=normalfont}
\centering
\subcaptionbox{Bed}{\includegraphics[width=0.323\columnwidth,keepaspectratio]{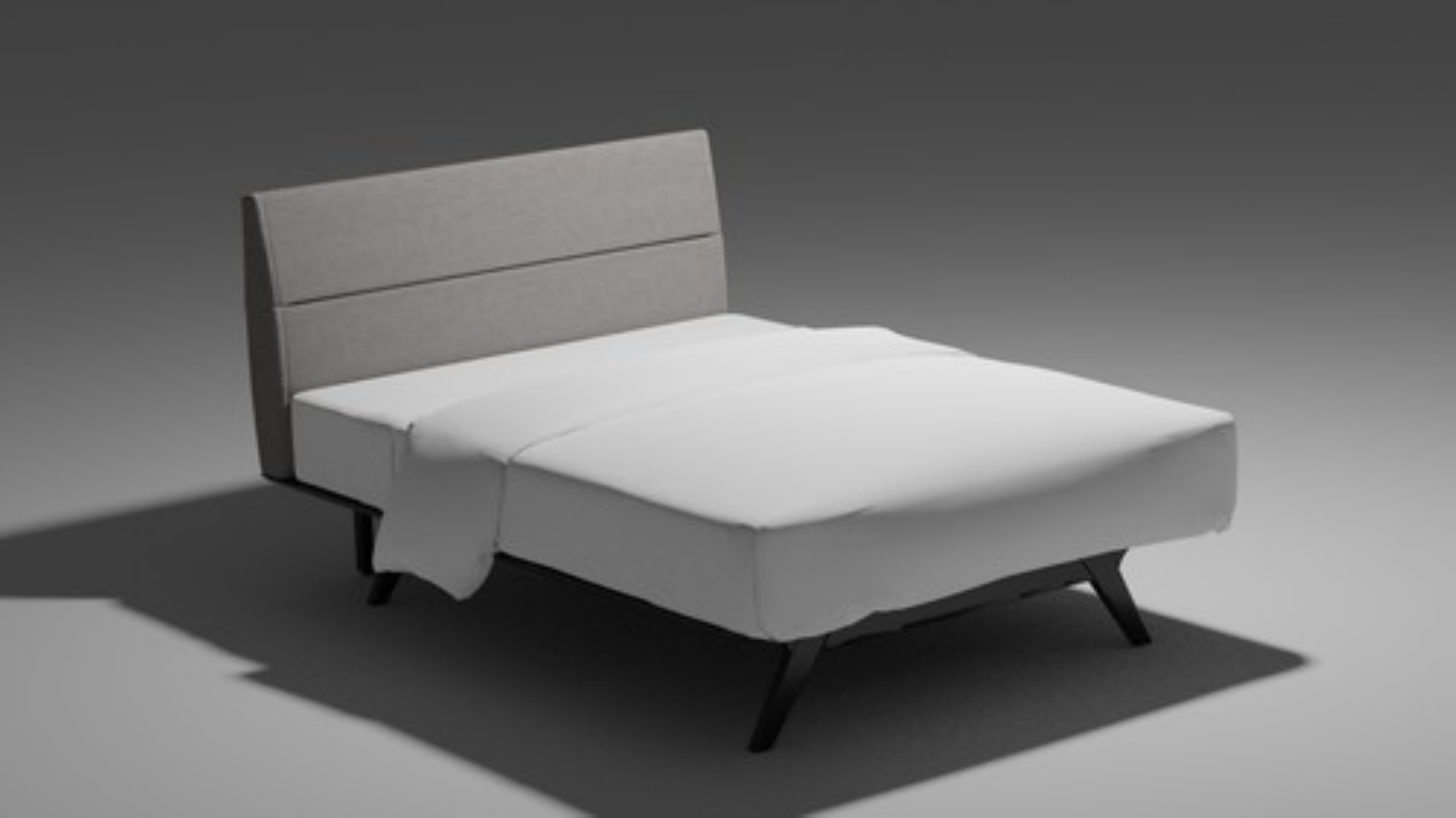}}
\subcaptionbox{Ottoman}{\includegraphics[width=0.323\columnwidth,keepaspectratio]{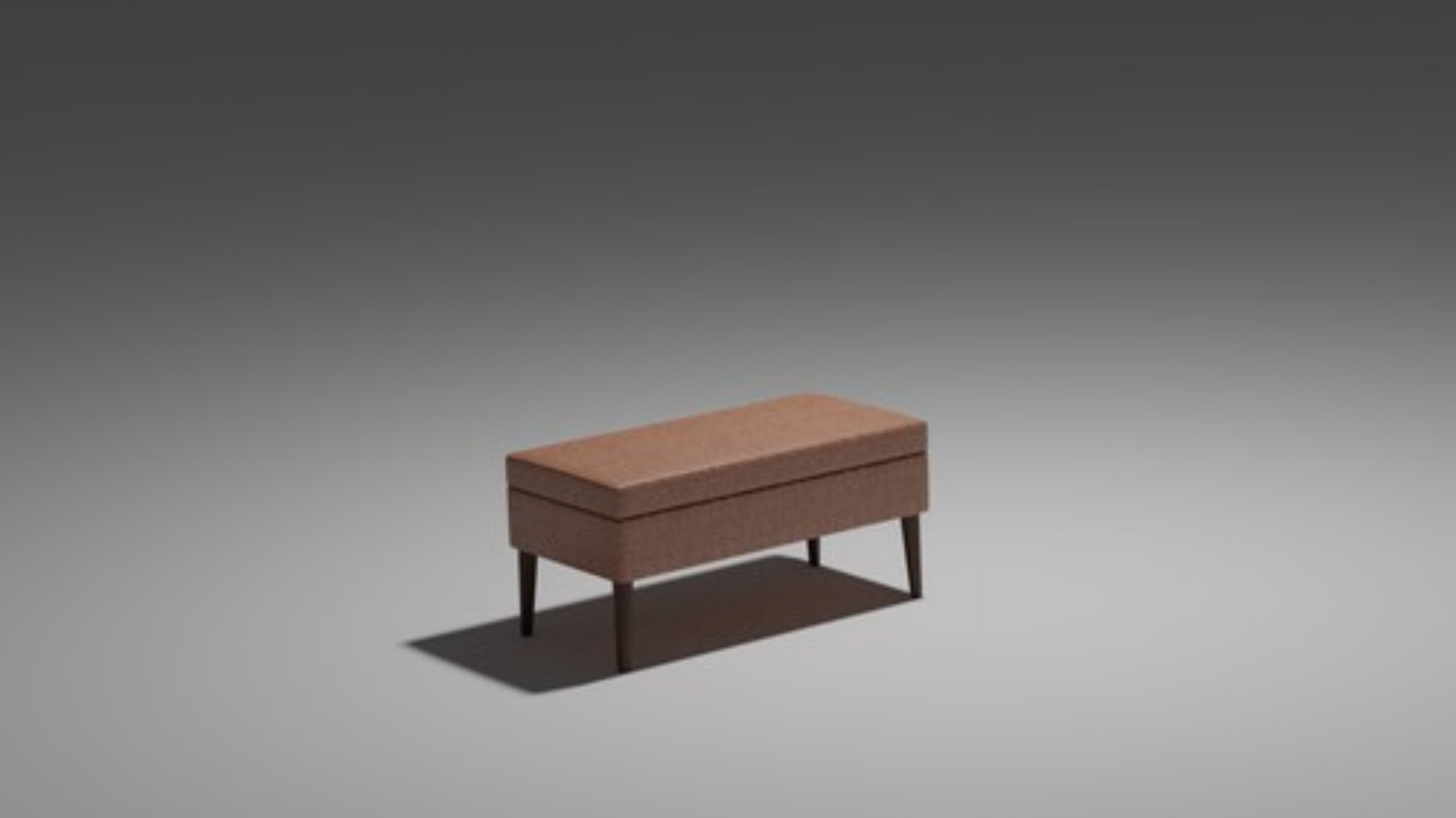}}
\subcaptionbox{TV Stand}{\includegraphics[width=0.323\columnwidth,keepaspectratio]{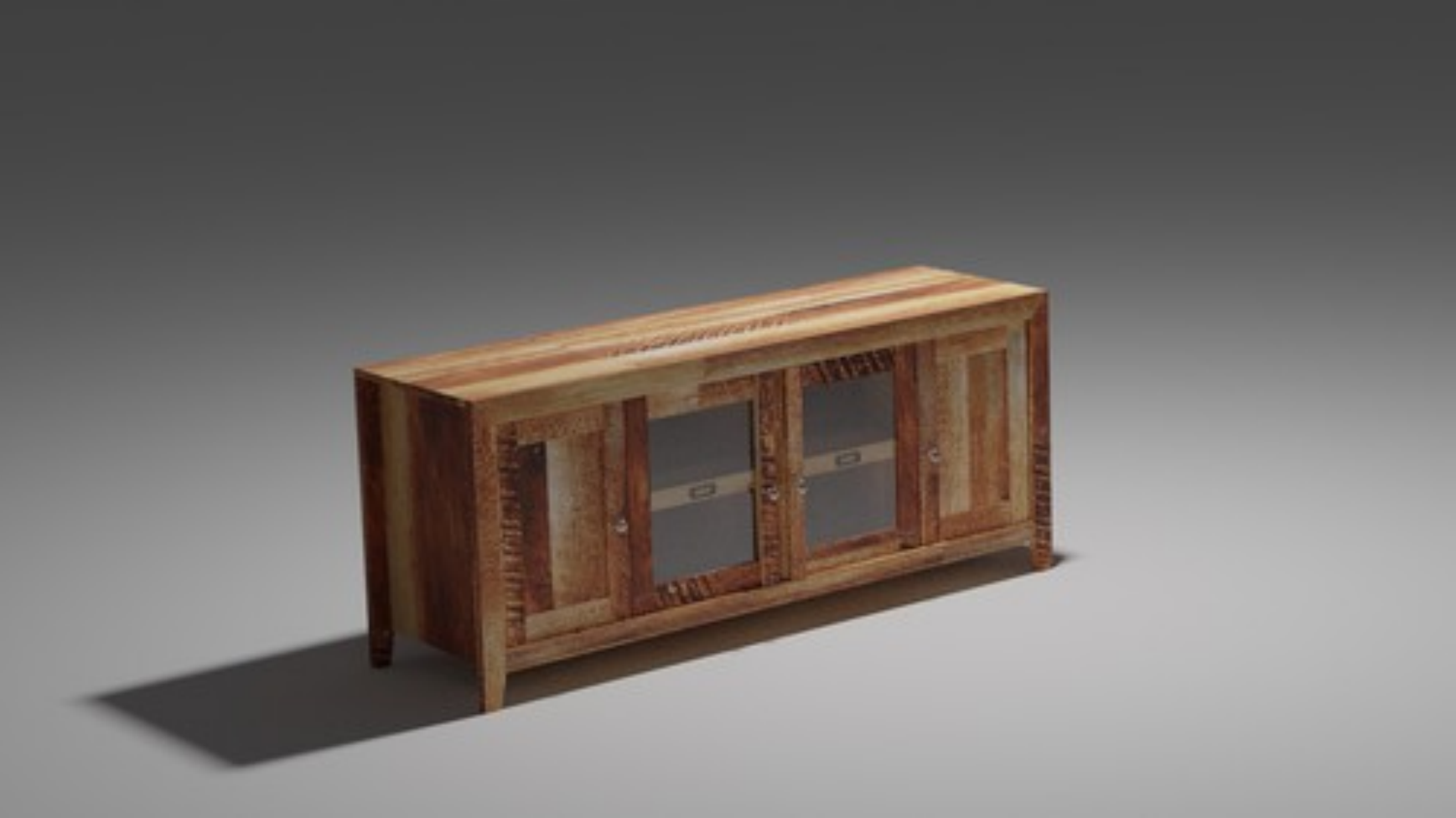}}
\subcaptionbox{Dining Chair}{\includegraphics[width=0.323\columnwidth,keepaspectratio]{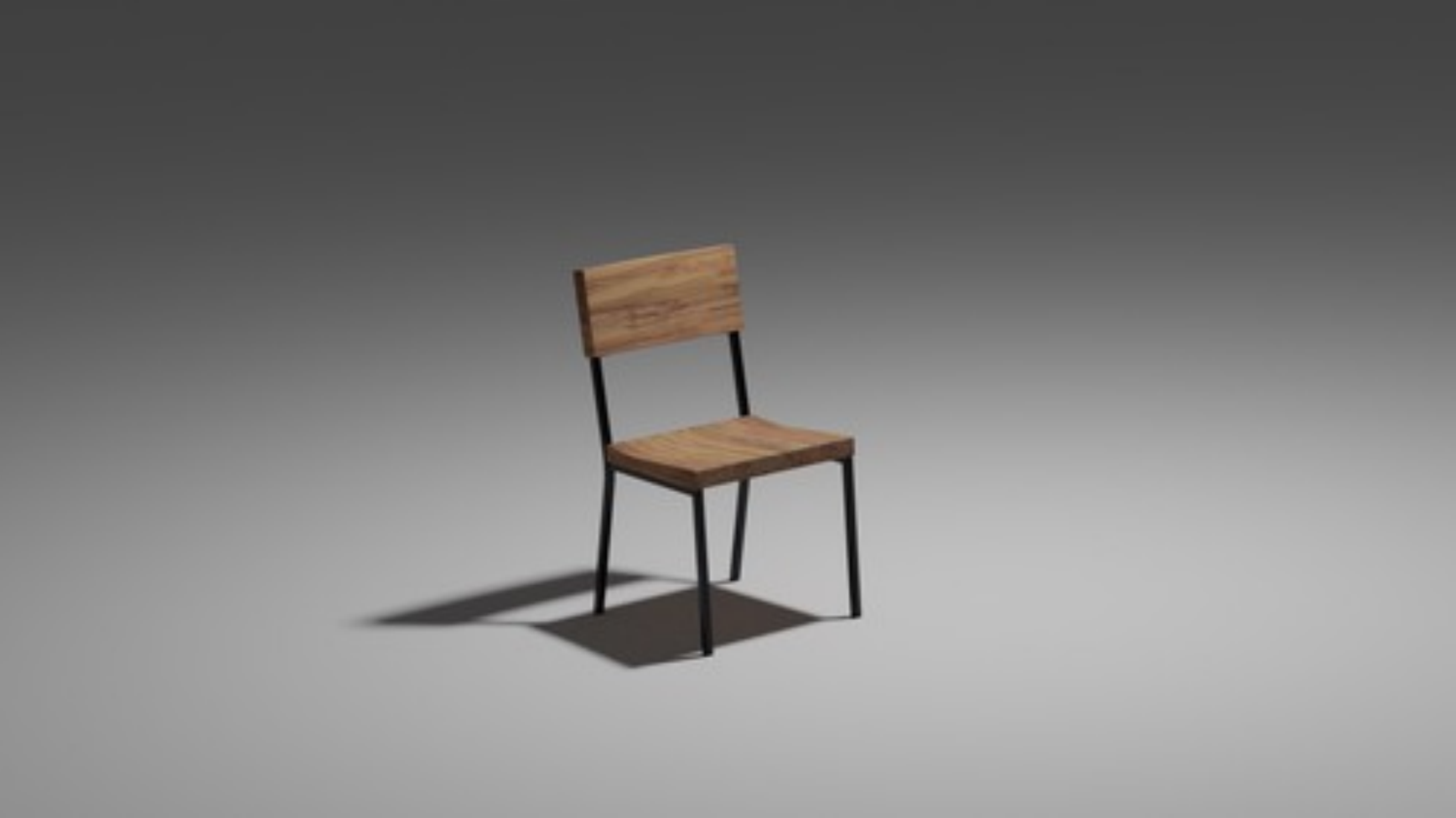}}
\subcaptionbox{Dining Table}{\includegraphics[width=0.323\columnwidth,keepaspectratio]{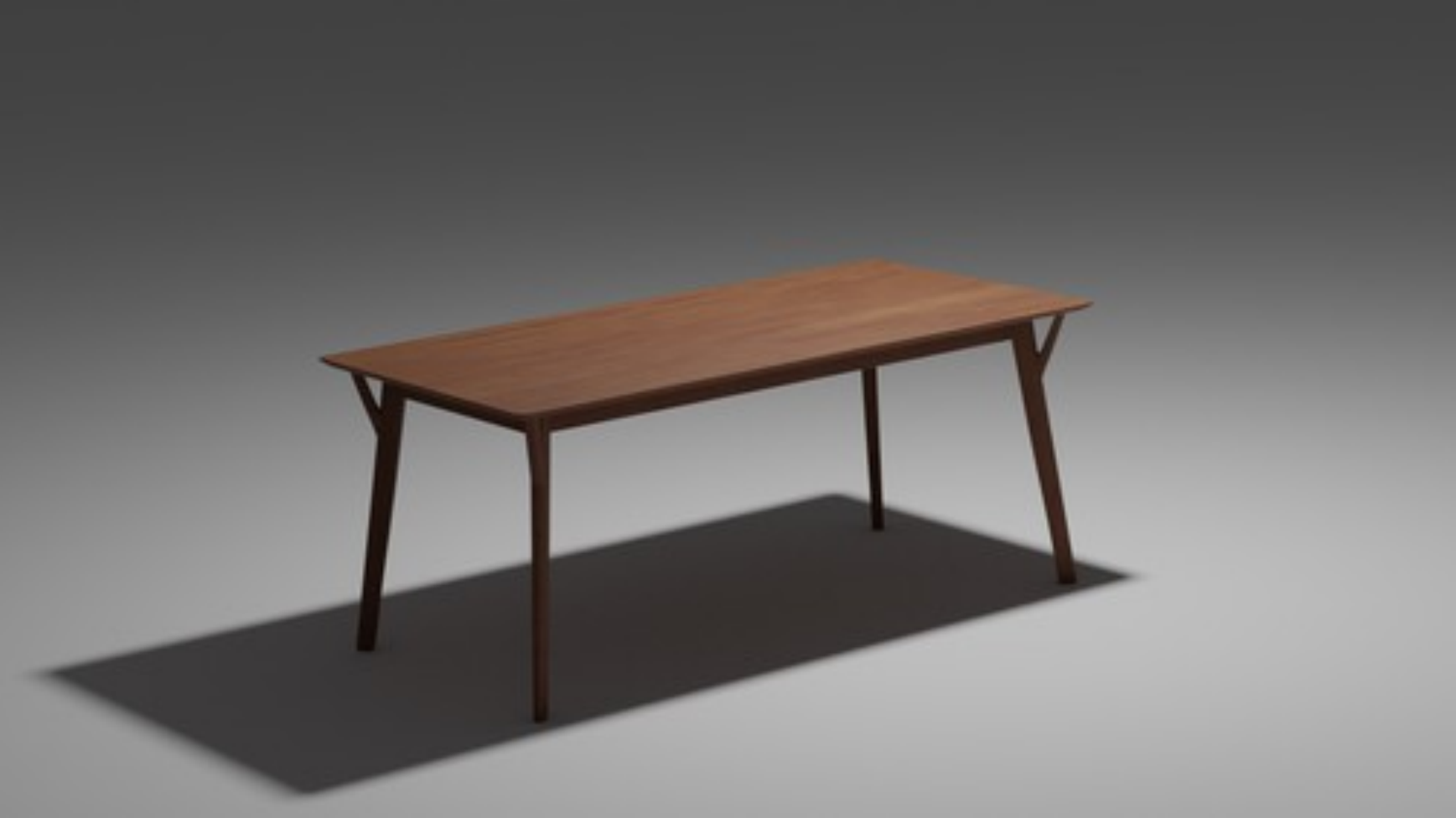}}
\subcaptionbox{Sofa}{\includegraphics[width=0.323\columnwidth,keepaspectratio]{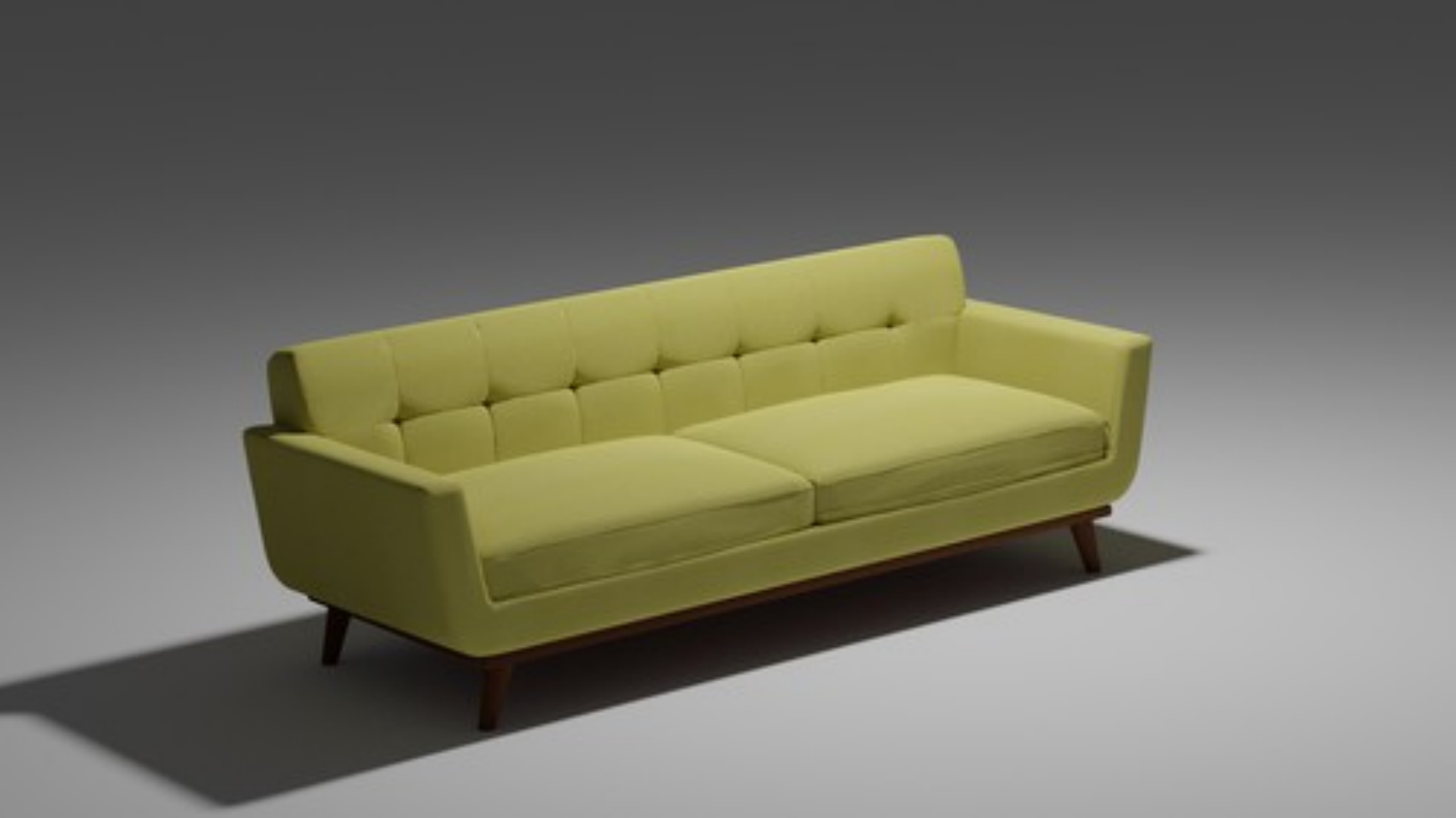}}
\subcaptionbox{Dresser}{\includegraphics[width=0.323\columnwidth,keepaspectratio]{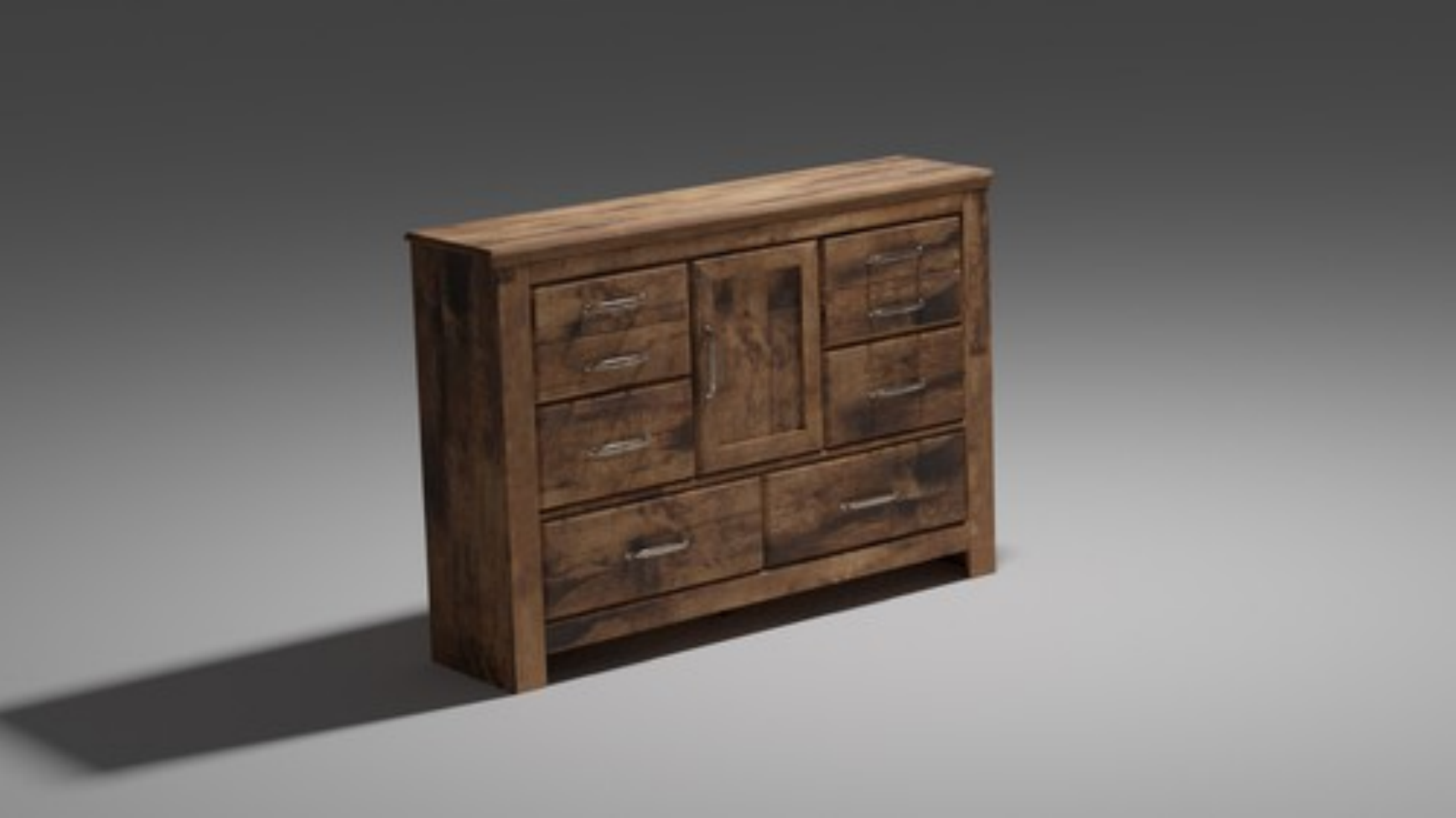}}
\subcaptionbox{Chest}{\includegraphics[width=0.323\columnwidth,keepaspectratio]{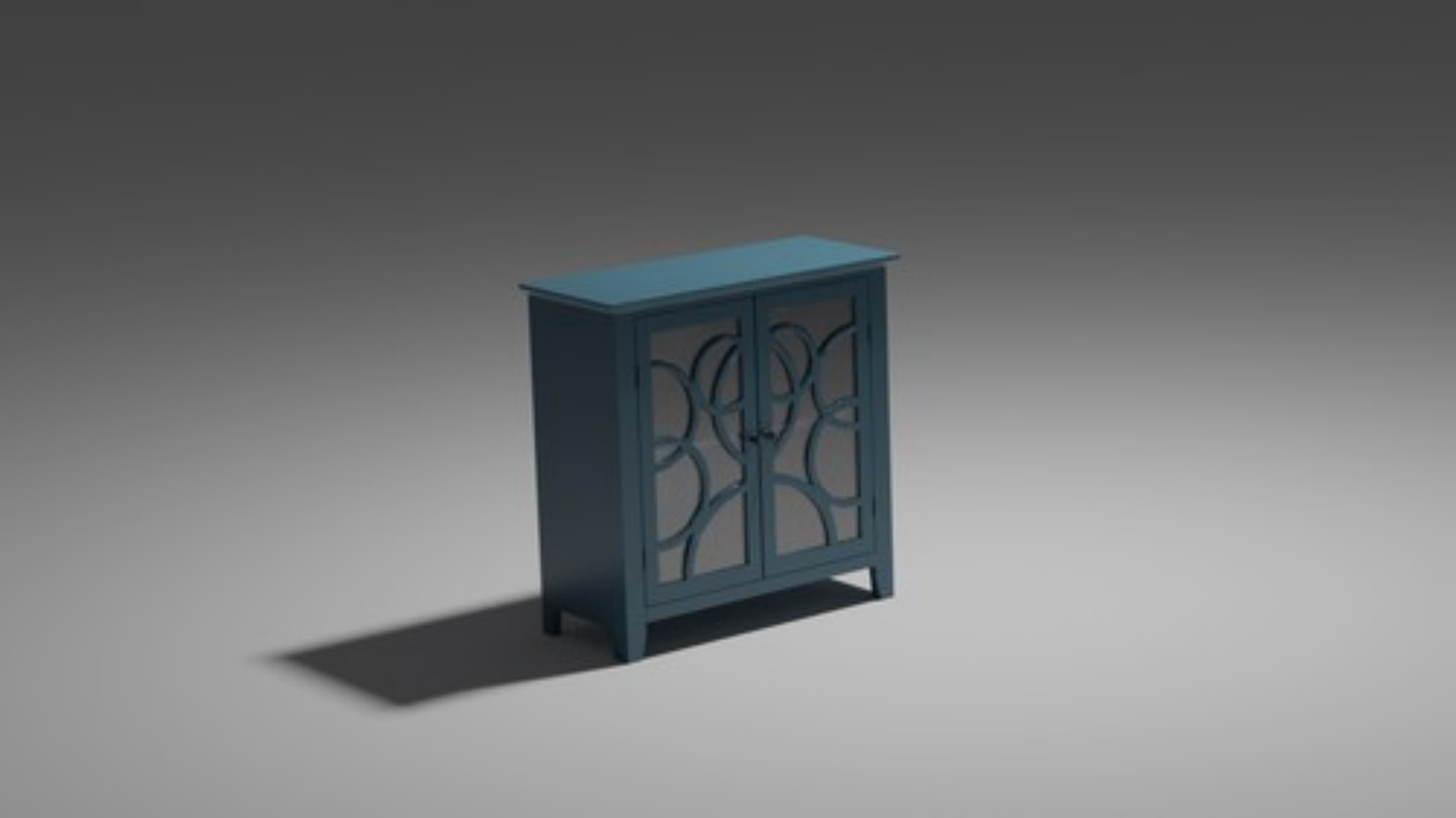}}
\subcaptionbox{Lamp}{\includegraphics[width=0.323\columnwidth,keepaspectratio]{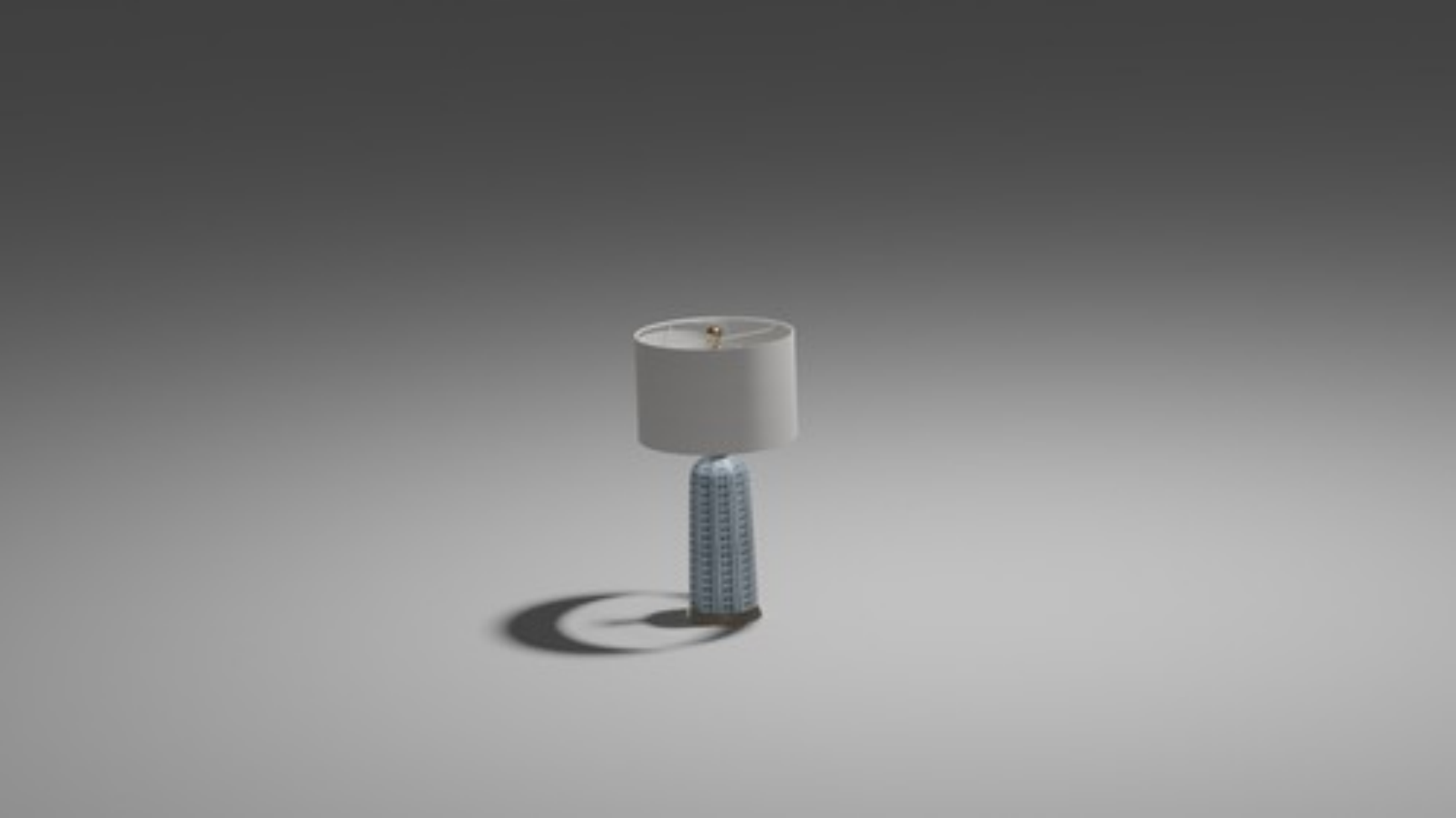}}
\caption{Examples of 3D model classes in our data set.}
\label{fig:exampleclasses}
\end{figure}

\section{Evaluating Image Style Compatibility} \label{sec:imageeval}

\emph{Qualitative Evaluation.}
In Table~\ref{tbl:predictions} we show example images comparing the predictions from our model with the ground truth labels provided by interior design experts. For images with distinct style labels, our model generates similar equivalent scores. In case of mixed style labels, our model recommends images with labels disagreeing on the same subset of styles as experts. For each mixed classification, we observe that the image's style is more of a spectrum of different visual features, and it can indeed belong to either of the estimated styles. Hence, in addition to estimating style, our framework has learned
distinctive visual features reflecting style. Our results validate that training with comparison labels can successfully capture noise in the style labels and accurately model the style spectrum of each image.

\emph{Quantitative Evaluation.}
We evaluate our model with images in our test set, which comprise of
$10\%$ of our total image data set.
Additionally, we also compare our model's performance against
two baseline models which estimate style without comparison labels.
Such baseline models were trained on:
(i) set $\allcleanstylelabels$ representing high agreement images where most experts agreed on the same style label. (ii) set $\allstylelabels$ containing all images.
High agreement images were given style labels of $\minvote=10$ for modern, $\minvote=8$ for traditional, and $\minvote=7$ for coastal and cottage (Figure~\ref{tbl:predictions}).
Such thresholds were selected empirically for best classification accuracy.
Below we summarize our quantitative evaluation outcomes.

\emph{Style Classification Accuracy.}
For each image $\imgindex$ in the test set, we predict its style from its softmax prediction, i.e $\text{argmax}_{\styleindex \in \styleset} \basenet( \feature_{\imgindex};  \weights)^{\styleindex}$ and compute its accuracy against the ground truth. In Figure~\ref{fig:classification_v1_vs_v2_4class} we show the classification performance where we perform better than the baselines for styles such as Cottage and Coastal but worse for Modern and Traditional. However, our network has better classification accuracy on average over the baseline models with $79\%$ accuracy and $4.7-8.1\%$ improvement.

\emph{Style Retrieval Performance.}
To evaluate the retrieval performance, we use the $16$-dimensional output of the second last layer of $\basenet$ as the embedding of each image in the test set.
Then, for each image, we retrieve the nearest image(s) in the remaining test set w.r.t. the Euclidean distance of extracted embeddings.
A retrieved image is considered relevant if its style matches with the style of the query image.
We compute mean Average Precision (mAP), recall rate at $\recallpos$, and Normalized Discounted Cumulative Gain (NDCG) on the retrieved images~\cite{liu2011learning}.
Figure~\ref{fig:retrieval1_v1_vs_v2_4class} and \ref{fig:retrieval5_v1_vs_v2_4class} shows the retrieval performances of our network and the compared baselines.
We observed approximately equally successful retrievals on all $4$ style categories for recall rate at 5,
with our model performing better in terms of mAP with $2.9-3.2\%$ improvement and NDCG with $2.4-3.5\%$ improvement.
In recall rate at 1, our model improved per class retrieval performance by $15-21\%$.

\emph{Comparison Labels and Global Compatibility.}
Comparison labels imply of a hidden global style score of each image, within each of our discrete styles.
For example, given images with a style score of $a$, $b$ and $c$, where $a > b$ and $b > c$, then $a > c$ should hold with high probability, since the Bradley-Terry model assumes stochastic transitivity.
We did not eliminate training comparisons that did not satisfy this condition. However, our network was able to rectify such conditions based on its performance.
Overall, comparison labels exhibit less noise compared to discrete class labels (Figure~\ref{fig:v1_vs_v2_4class}). Elimination or correction of such noise is a subject of future work, and an active area of research~\cite{saha2018ranking}.

\section{Furniture Style Compatibility}\label{sec:embeddings}

\subsection{3D Model Similarity}\label{sec:ret3d}
\begin{figure}[bt]
\centering
\subcaptionbox{Similar}{\includegraphics[width=0.46\columnwidth,keepaspectratio]{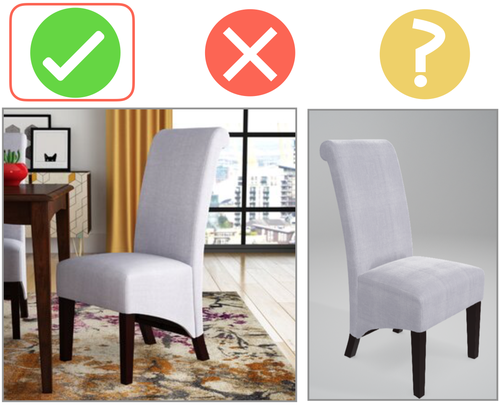}}
\hfill
\subcaptionbox{Dissimilar}{\includegraphics[width=0.46\columnwidth,keepaspectratio]{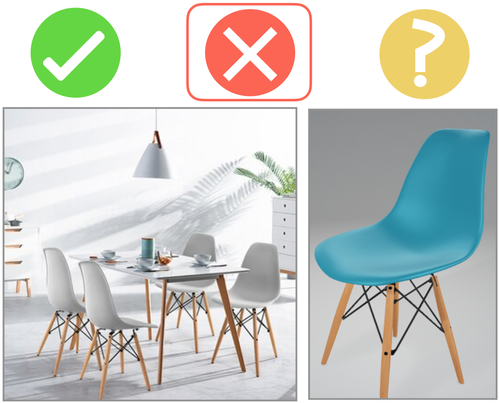}}
\caption{Validating visual similarity of furniture.
Furniture might appear visually different when comparing rendered views of 3D models to curated scene images.
We manually verify the furniture's similarity with a tool we created, which includes the following options:
(a) Furniture in image, and its 3D model exhibit visual
similarity.
(b) Furniture's model has a visually different color, and hence disallows us from considering it part of
our recommendation model.
(c) A question mark button, in case the user is not sufficiently confident.}
\label{fig:image_model_match}
\end{figure}

We infer the style compatibility of 3D furniture models using images,
similar to recent work that infers style of 3D furniture models from rendered images~\cite{liu2019learning,lim2016identifying}.
However, the \emph{key difference} of our method is that we utilize the entire styled scene, to estimate a 3D furniture's style compatibility. The machinery we use to that end is a furniture's image embeddings.
A 3D furniture model might appear in multiple images (Figure~\ref{fig:furnitureimages}).
Hence, we first validate that 3D models have a consistent visual appearance in all images they appear in (Figure~\ref{fig:image_model_match}).
We then measure the stylistic compatibility between two furniture pieces
by calculating the minimal embedding distance between all images the furniture items are associated with
\begin{equation} \label{eq:furnituredistance}
d(m_i,m_j) = \min_{\forall i \in I,\forall j \in J} ||e(i),e(j)||,
\end{equation}
where $m_i$ and $m_j$ are 3D furniture models,
$I$ is the set of images associated with the furniture piece $m_i$,
and $J$ is the set of images associated with furniture $m_j$,
and $e$ is a function that returns an image's embedding.
Based on the above, we create a system that given a seed
3D furniture model(s), retrieves similar style 3D furniture's (Section~\ref{sec:mapping3dmodels}).

\begin{figure*}[t]
\captionsetup[subfigure]{position=top, labelformat=empty,textfont=normalfont}
\begin{center}
\subcaptionbox{\vspace{-2mm}Query 3D Model}{
\tcbox[top=0pt,left=0pt,right=0pt,bottom=0pt,colframe=orange!60!black,
           colback=orange!60]{
\includegraphics[trim=430 20 430 20, clip,height=2.0cm,keepaspectratio]{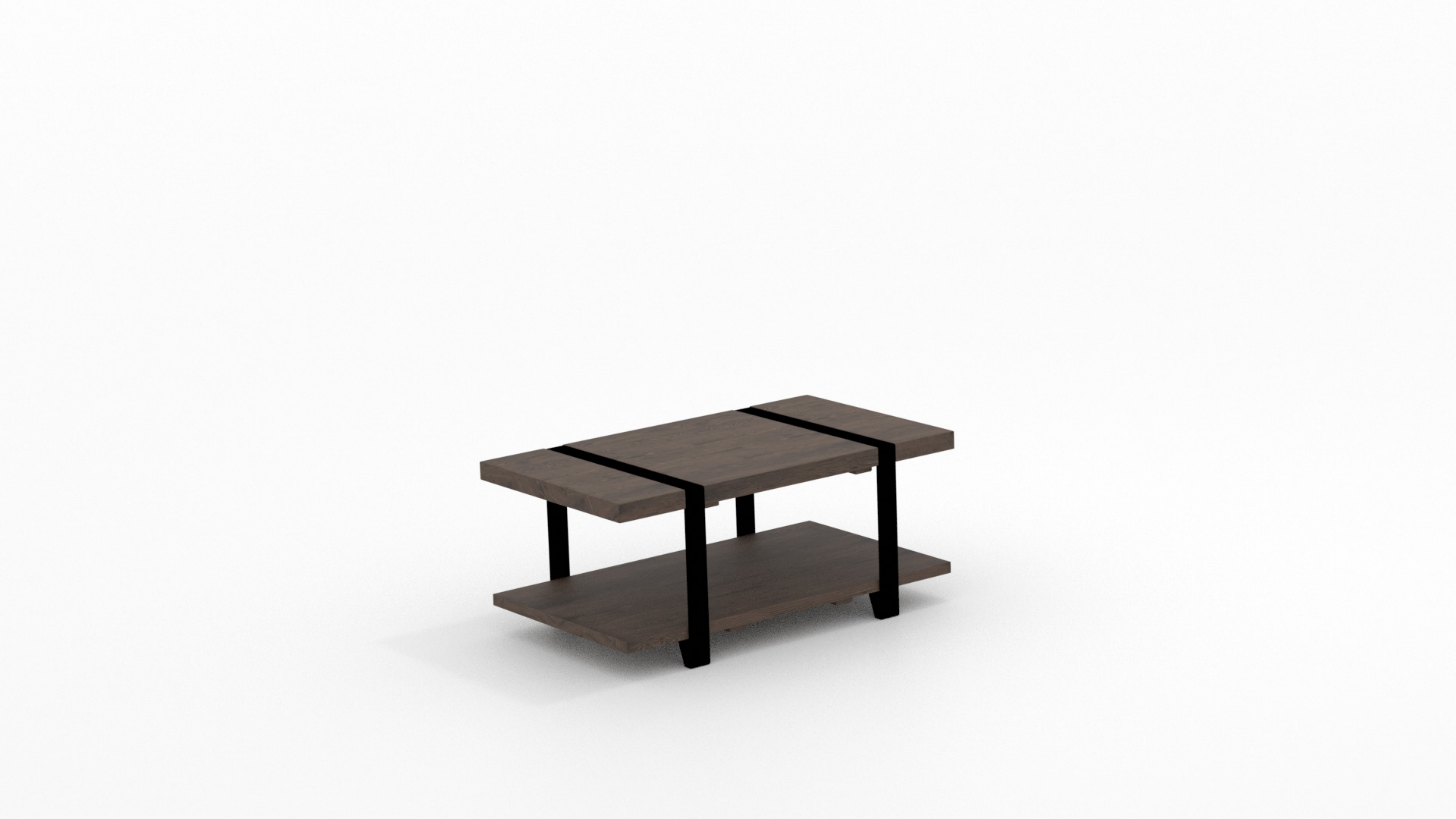}
}}
\subcaptionbox{\vspace{-2mm}Ranked Results}{
\tcbox[top=0pt,left=0pt,right=0pt,bottom=0pt,colframe=red!10!black,
           colback=red!10]{
\includegraphics[trim=230 20 230 20, clip,height=2.0cm,keepaspectratio]{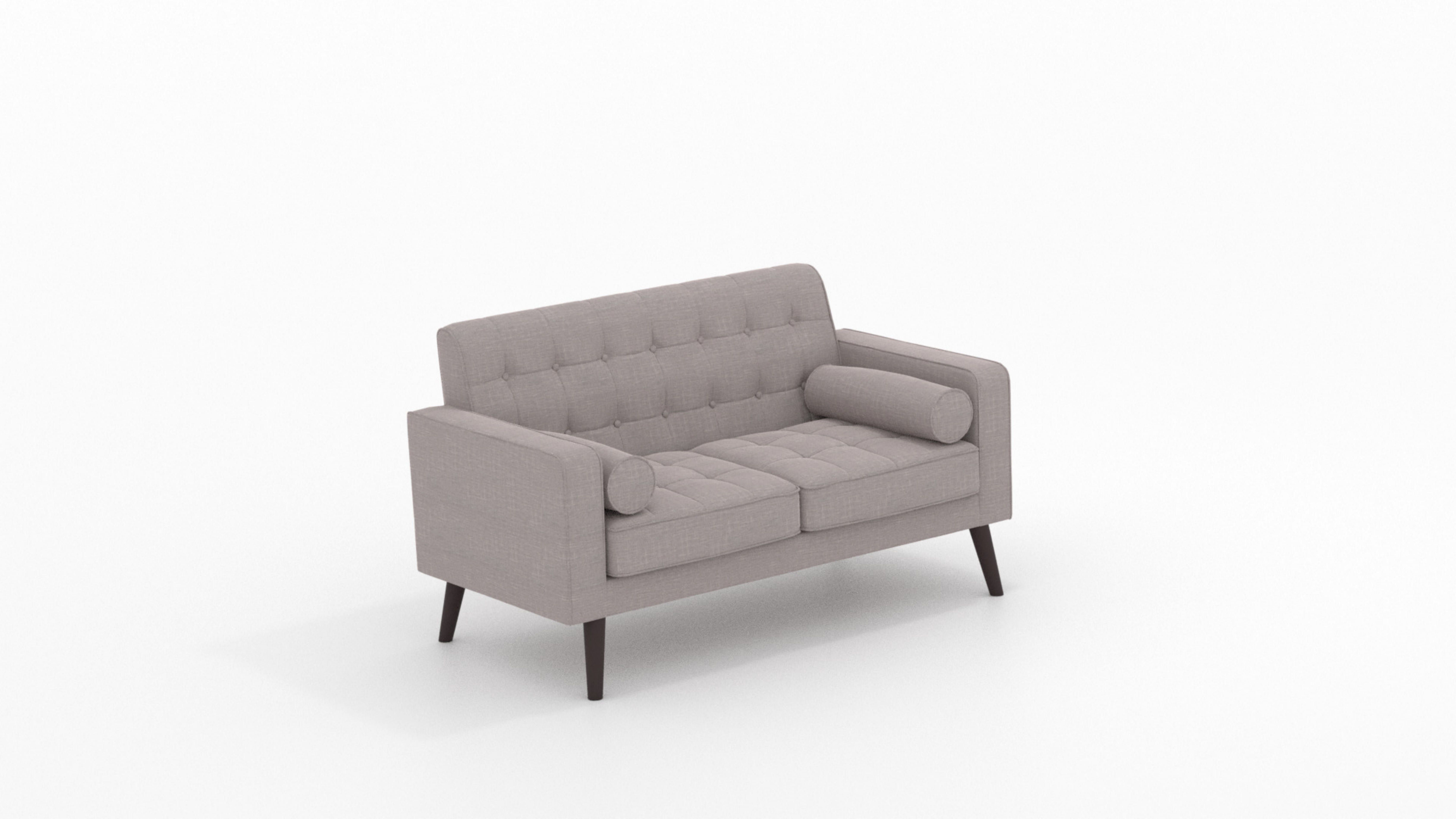}
\includegraphics[trim=230 20 230 20, clip,height=2.0cm,keepaspectratio]{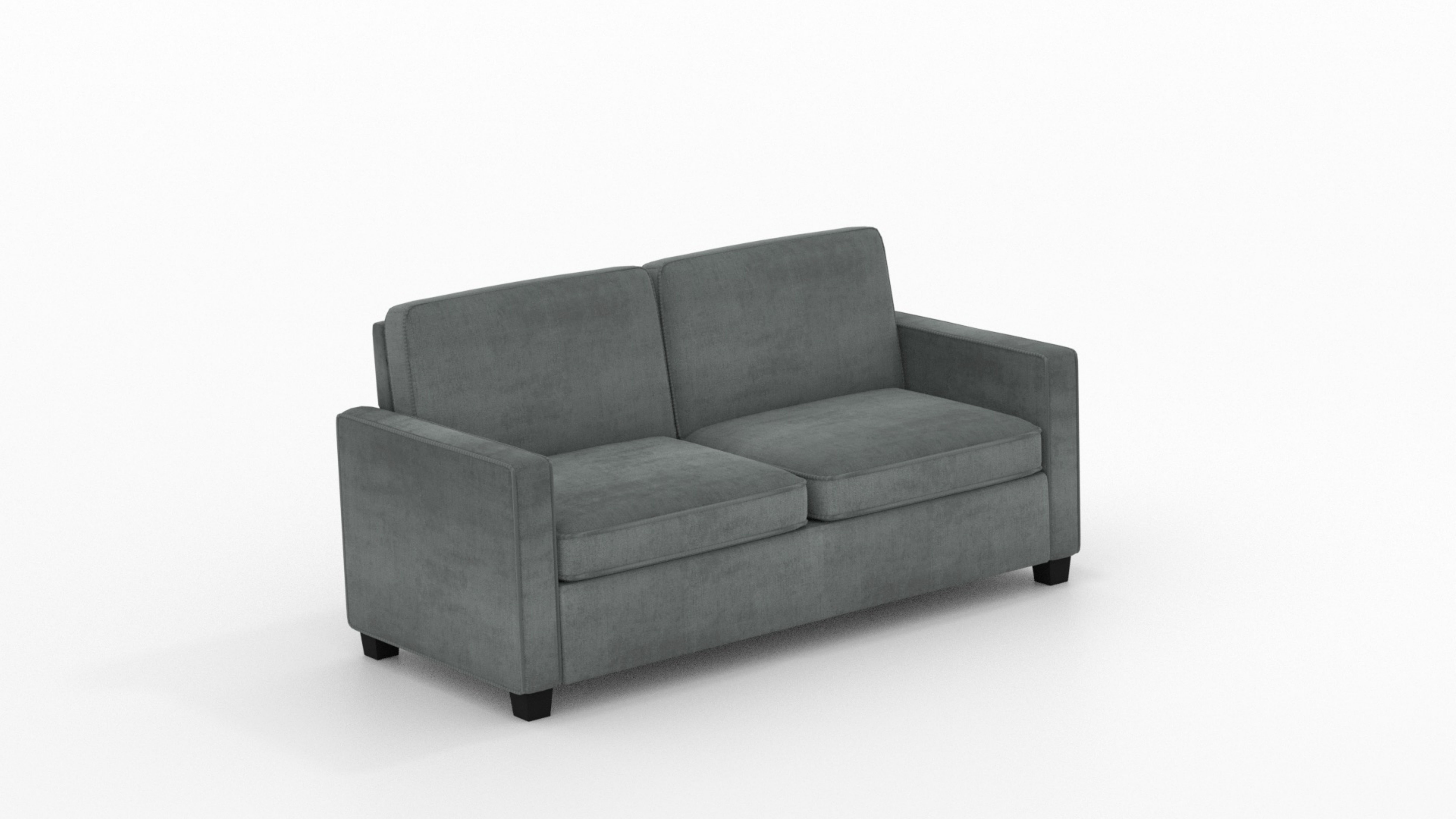}
\includegraphics[trim=230 20 230 20, clip,height=2.0cm,keepaspectratio]{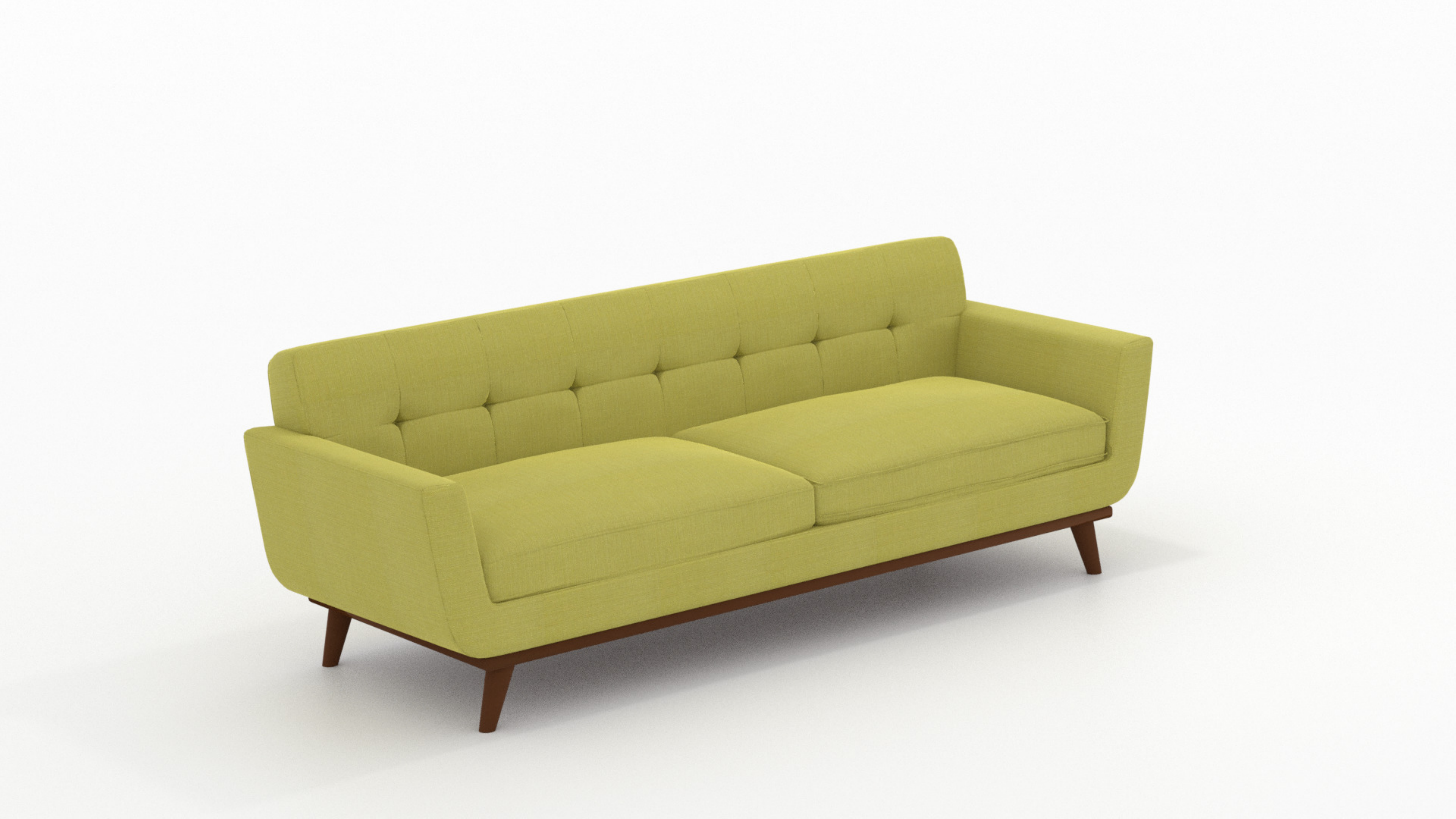}
\includegraphics[trim=230 20 230 20, clip,height=2.0cm,keepaspectratio]{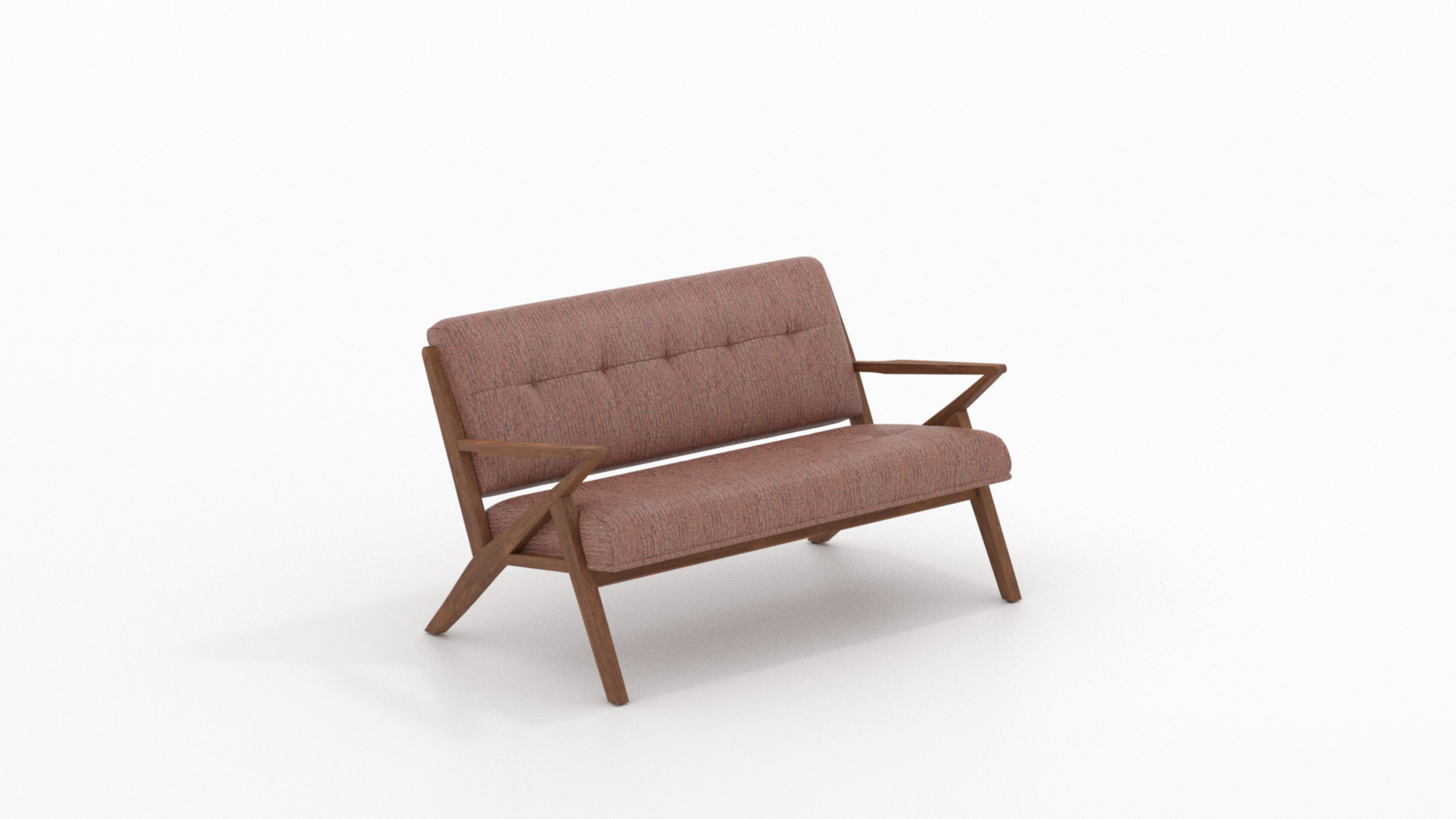}
\includegraphics[trim=230 20 230 20, clip,height=2.0cm,keepaspectratio]{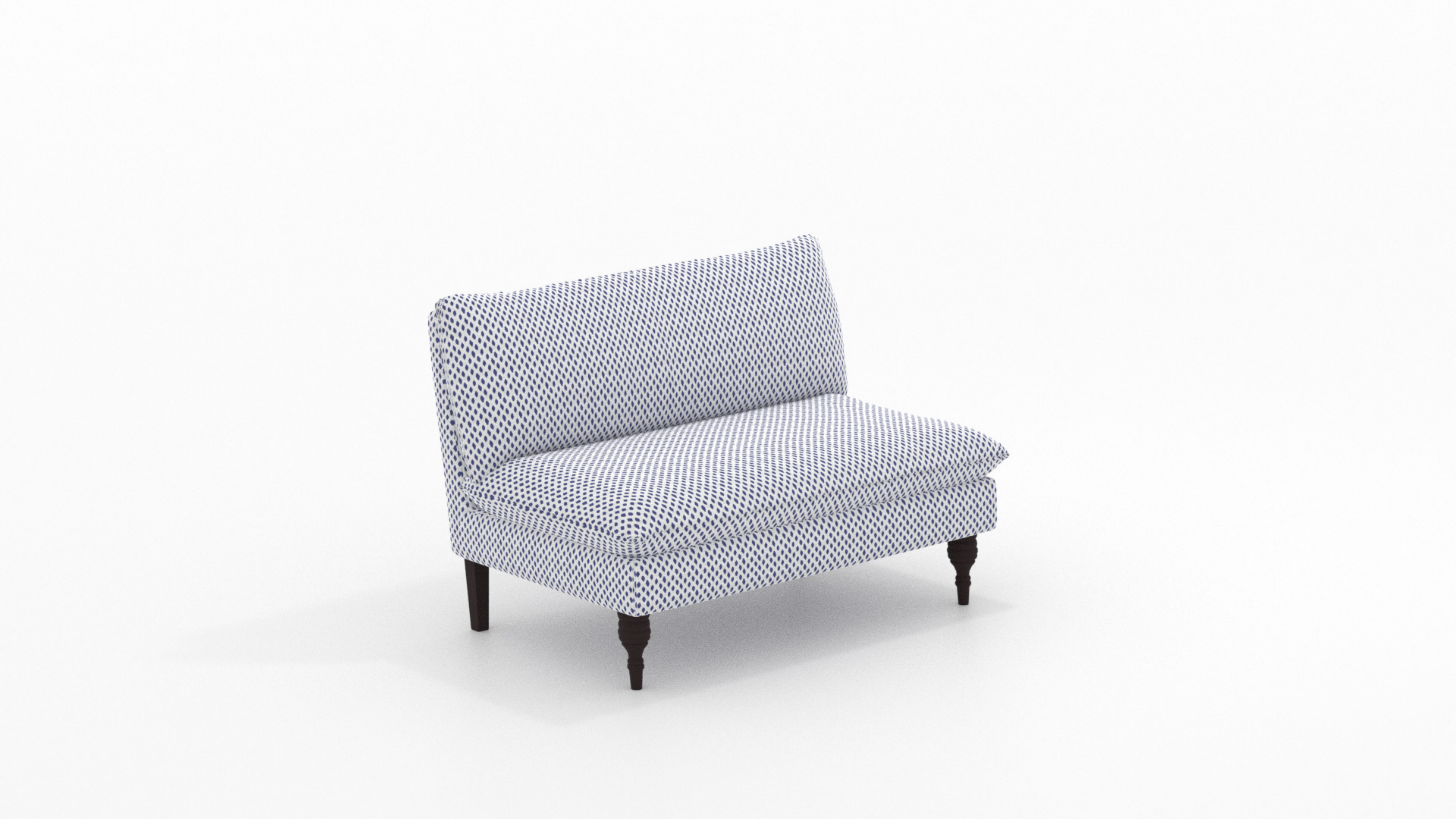}
}}

\end{center}
\caption{Our method provides style-aware 3D model recommendations.
Here, based on the query coffee table on the left, we provide compatible sofas on the right.
\vspace{-5mm}
\label{fig:recs3dmodes}
}
\end{figure*}

\subsection{Style-Compatible Scene Modeling}
\label{sec:curating}
\begin{figure*}[t!]
  \centering
  \subcaptionbox{Junior Bedroom}{\includegraphics[height=3.23cm,keepaspectratio]{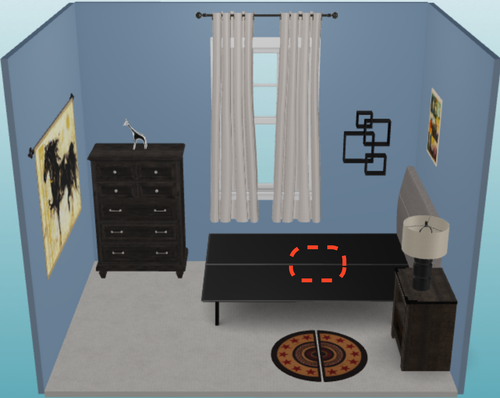}}
  \subcaptionbox{Dining Room}{\includegraphics[height=3.23cm,keepaspectratio]{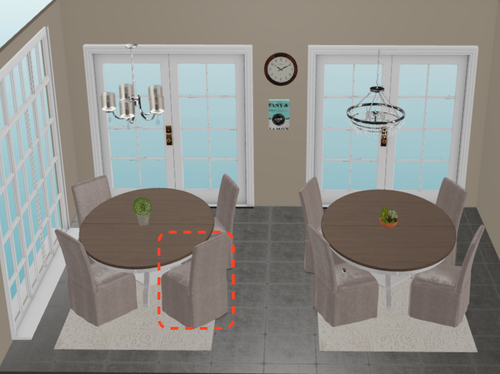}}
  \subcaptionbox{Meeting Room}{\includegraphics[height=3.23cm,keepaspectratio]{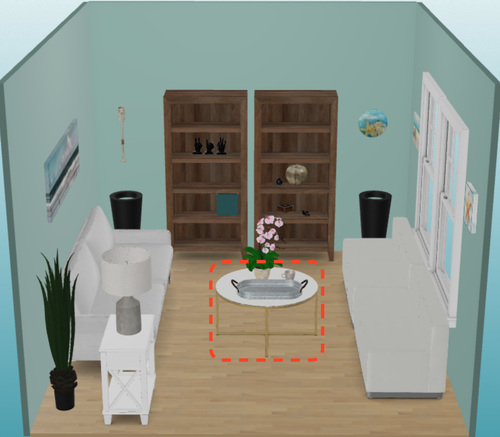}}
  \subcaptionbox{Family Room}{\includegraphics[height=3.23cm,keepaspectratio]{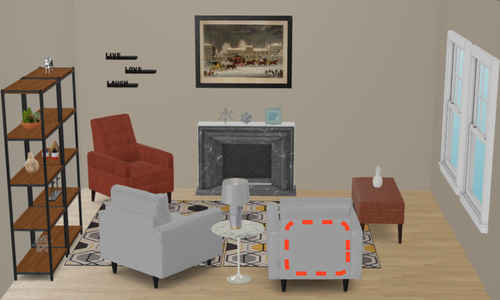}}
  \subcaptionbox{Living Room}{\includegraphics[height=3.23cm,keepaspectratio]{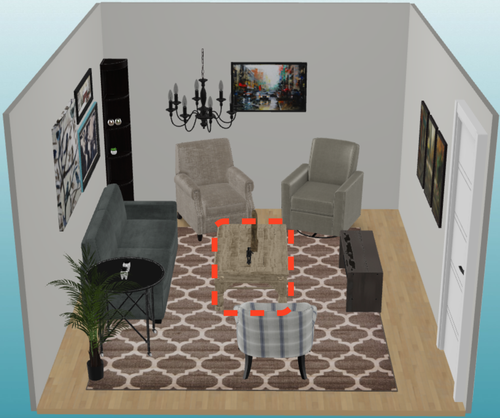}}
  \subcaptionbox{Hip Studio}{\includegraphics[height=3.23cm,keepaspectratio]{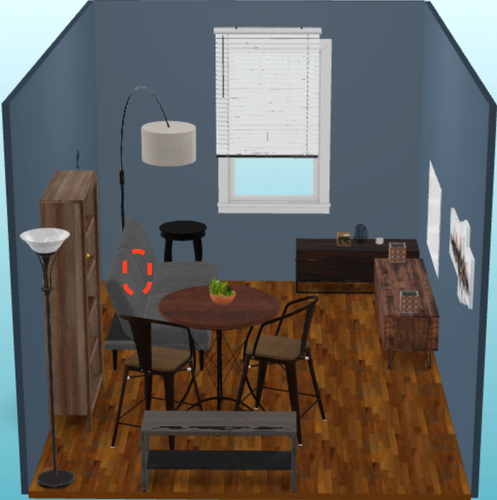}}
  \subcaptionbox{Asian Restaurant}{\includegraphics[height=3.23cm,keepaspectratio]{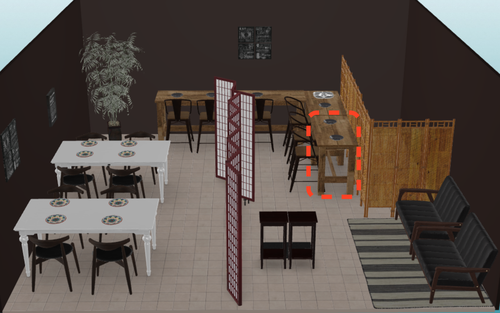}}
  \subcaptionbox{Lounge}{\includegraphics[height=3.23cm,keepaspectratio]{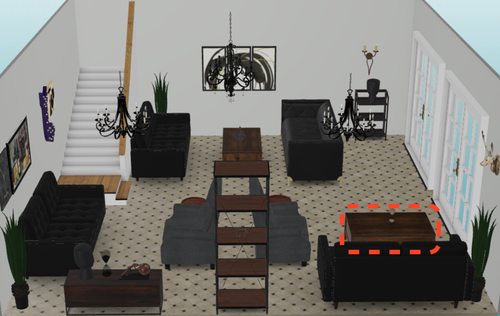}}
  \caption{Given a context seed furniture marked in red, we curate a
  scene by selecting from a set of stylistically-compatible furnitures recommended by our model.
  Our approach is flexible across different room types, from a bedroom to a lounge area.
  }
 \label{fig:roomresults}
\end{figure*}

To qualitatively validate our 3D object style similarity,
we developed a system for style-aware modeling of virtual scenes.
If furniture has similar style, it will be preserved in a 3D setting, allowing to create a more harmonious ambience in the room in which furniture's fit-well together. Another motivation for such a system is that it is difficult
for users to navigate a catalog of 3D furniture models and search for stylistically compatibility furniture fitting a user's scene~\cite{tangelder2004survey}.
To that end, we designed our system to provide style-aware
suggestions based on selected query scene objects. A user may
select one, or multiple 3D furniture's, for which our system
generates a list of style-compatible suggestions.
In response to a user's query, our system displays suggestions of top most compatible furniture, as shown in Figure~\ref{fig:recs3dmodes}.
Suggestions are retrieved in real-time, since image embeddings and embedding distances can be calculated in advance.
We included different strategies to retrieve style-compatible furniture suggestions for virtual scenes:

\emph{Single Seed Style Reference.} %
In this 3D modeling scenario, a user selects a main 3D furniture piece
to anchor a scene's style. All further style-compatible suggestions are based only on the seed 3D model:
\begin{equation}\label{eq:singleseed}
d(m_i, m_{seed}),
\end{equation}
where $d$ measures the stylistic compatibility for a pair of scene objects (Eq.~\ref{eq:furnituredistance}), $m_i$ are scene objects for which we measure the compatibility with the seed furniture $m_{seed}$, which is the scene's main 3D model.
The motivation behind such a suggestion strategy is that 3D artists typically want to focus on a single, main furniture piece for the entire scene~\cite{liu2015composition}. Other furniture objects are placed around the main furniture, to enhance it visually. For example, an artist is creating a scene to market a sofa or an accent chair.
To exemplify such scenario, we created specific room types such
bedroom, studio apartment, and restaurant (Figure~\ref{fig:roomresults}).
Please see section~\ref{sec:single} for more details.

\emph{Multiple Style References.} %
In this scenario, we consider multiple scene objects
for providing style compatible suggestions.
~\cite{liu2015style} propose a compatibility energy function for an entire scene as the sum of the compatibility distances between all scene objects. Similarly, we define the style compatibility for a selected object $m_{select}$ as the
sum of the style distances with all existing scene objects:
\begin{equation}\label{eq:multiseed}
\sum_{m_i \in S } d(m_{select}, m_i),
\end{equation}
where $S$ is the set of scene 3D models, and $d$ is a function that returns the style compatibility distance according to Eq.~\ref{eq:furnituredistance}.
Similarly, to measure the stylistic compatibility of an entire scene:
\begin{equation}\label{eq:entirescene}
\sum\limits_{ \substack{m_i,m_j \in S\\ i \neq j}} d(m_i, m_j).
\end{equation}

To exemplify such a stylistic compatibility scenarios, we investigated
style of furniture arrangements and introduce a tool for interactive
style-aware scene building.
Our tool allows users to create and modify existing furniture arrangements for various room types.
Please see Section~\ref{sec:scenemodeling} for more details.

\section{Evaluating 3D Furniture Style Compatibility} \label{sec:mapping3dmodels}
We implemented a system for interactive scene building of style-compatible furniture arrangements.
The system's input is either single or multiple object
queries and outputs suggestions of style-compatible furniture pieces,
similar to prior work~\cite{liu2015style}.
Suggestions are based on embeddings of furniture images. We precompute image style embeddings and pairwise 3D furniture compatibility distances to achieve interactive run-time for all furniture suggestion queries.
Such quality assists in interactive modeling scenarios with multiple scene objects. Scene modeling, and other interactive components of our system are implemented in Python and JavaScript and tested on a 2.5 GHz Intel Core i7.
We tested our system on the following scenarios:
(i) Creating a scene to fit a single 3D model's style, and
(ii) Modification of scenes for multiple 3D model's styles.
Below we describe our experiment setup, including how we obtained our 3D model data set.

\emph{3D Furniture Models Dataset}.
We collected textured 3D models of furniture from a variety of sources:
manually created, sourced from the web, and provided by the furniture's manufacturer. In our experiments we used about 1148 3D furniture models, which fit residential and public spaces such as restaurants.
Models are divided into 20 object classes, including:
accent chairs, sofas, sectionals, end tables, coffee tables, beds, dressers, bookcases, lamps, dining tables, dining chairs and similar.
We manually normalized the models to a consistent scale and orientation, keeping 3D models that are visually distinct from each other. Figure~\ref{fig:exampleclasses} shows few examples of the furniture classes we used.

\emph{Validating Visual Appearances between 3D Furniture and Images}.
Our data set contains images scrapped form the web, which might contain noise and irregularities either in images or in 3D models.
Irregularities might result from various causes: errors in the 3D model creation pipeline, light conditions, material setup, among other reasons. The style compatibility system we propose is dependent on consistent visual appearance of 3D models and associated images.
Typically, several images are associated with each 3D model, showing the model from different camera angles. To assess such visual similarity,
we manually compare a furniture's 3D model and its images.
Such manual comparison is necessary since it is difficult to computationally judge similarity between 3D models and images.
For annotating such comparisons, we use 3 options:
similar, not similar, and unknown, in case of insufficient confidence in the decision. Experimentally, we found that only $60\%$ (1148 out of 1911) of the 3D models we sampled have a valid visual appearances, suggesting the importance of this step.
Selected 3D models have a matching visual appearance to their respective images (Figure~\ref{fig:image_model_match}).

\begin{figure}[tb]
\centering
\begin{tabular}{c c}
  \raisebox{18mm}{\parbox[b]{2mm}{\hypertarget{fig:morelesscomp:a}{a} }}
  \includegraphics[width=0.45\columnwidth,keepaspectratio]{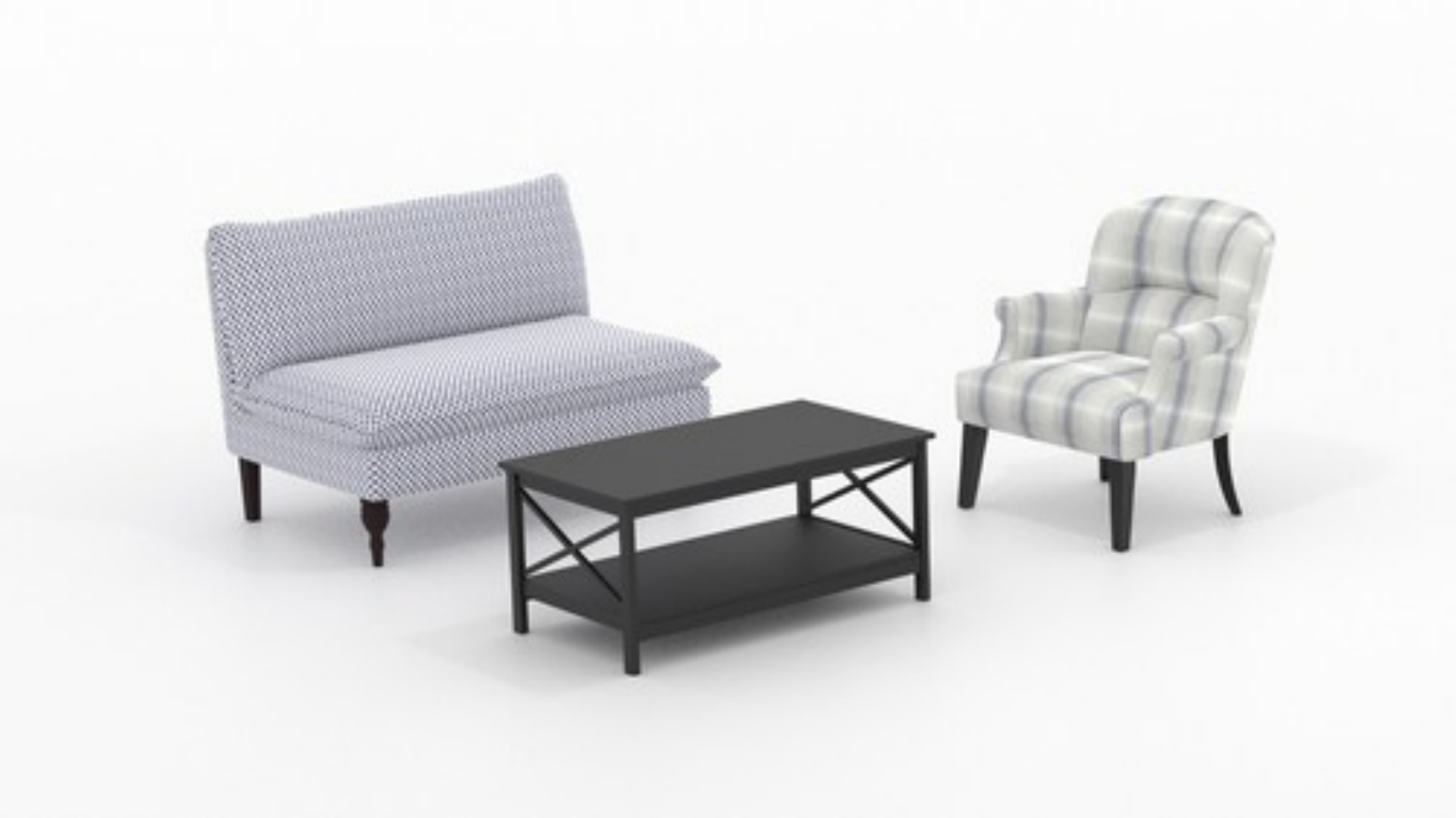} &
  \includegraphics[width=0.45\columnwidth,keepaspectratio]{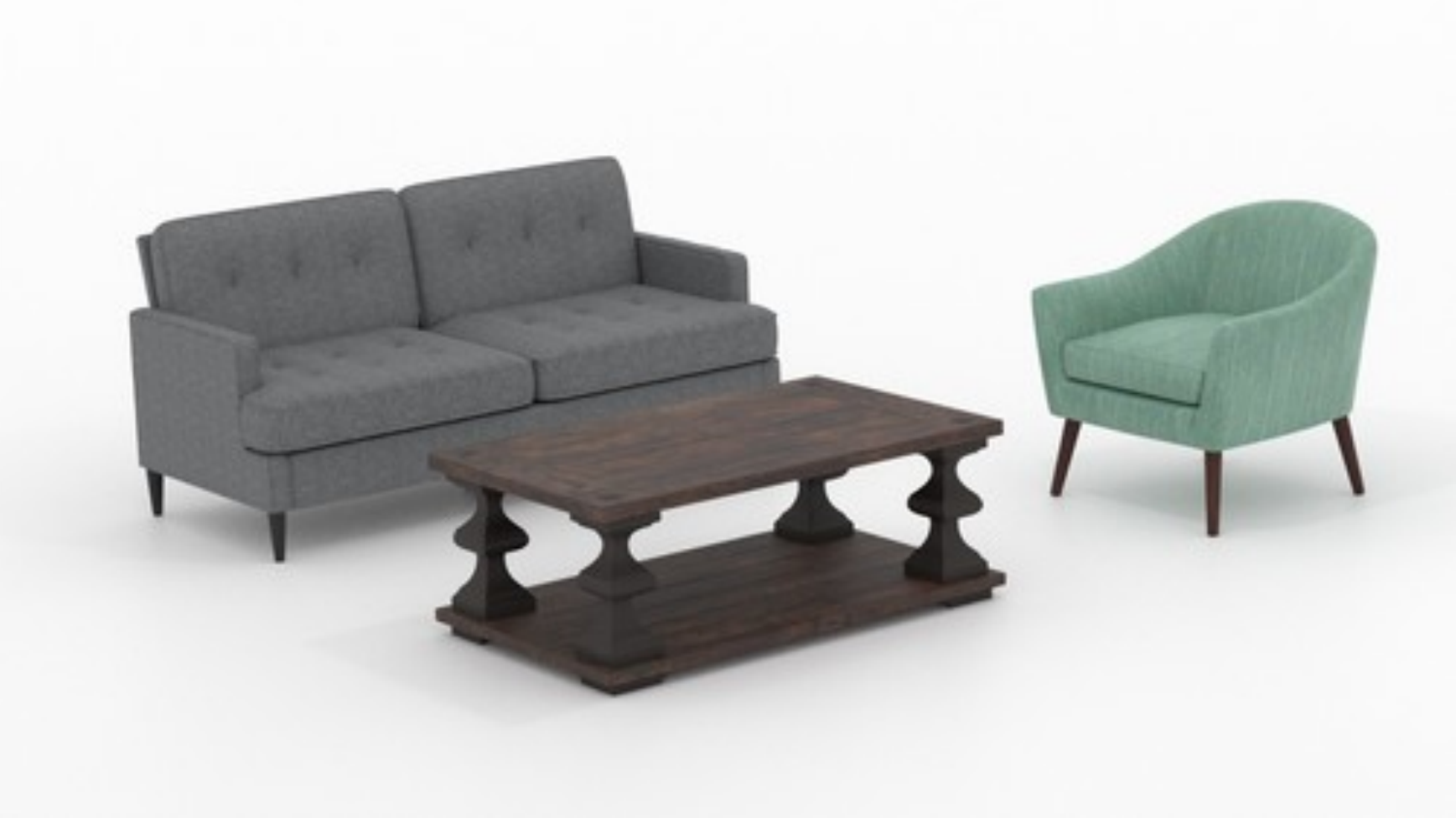} \\
  \raisebox{18mm}{\parbox[b]{2mm}{\hypertarget{fig:morelesscomp:b}{b} }}
  \includegraphics[width=0.45\columnwidth,keepaspectratio]{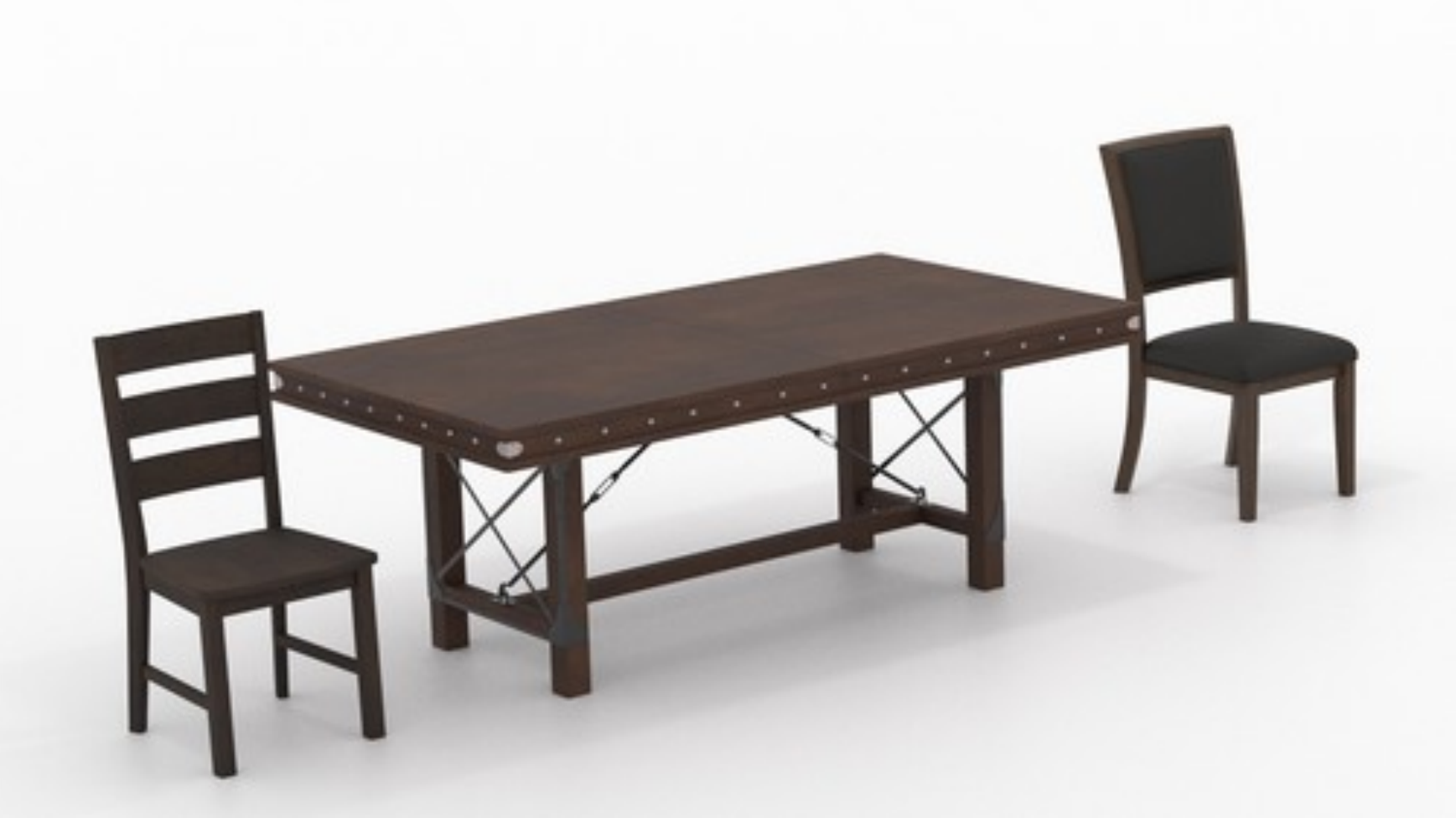} &
  \includegraphics[width=0.45\columnwidth,keepaspectratio]{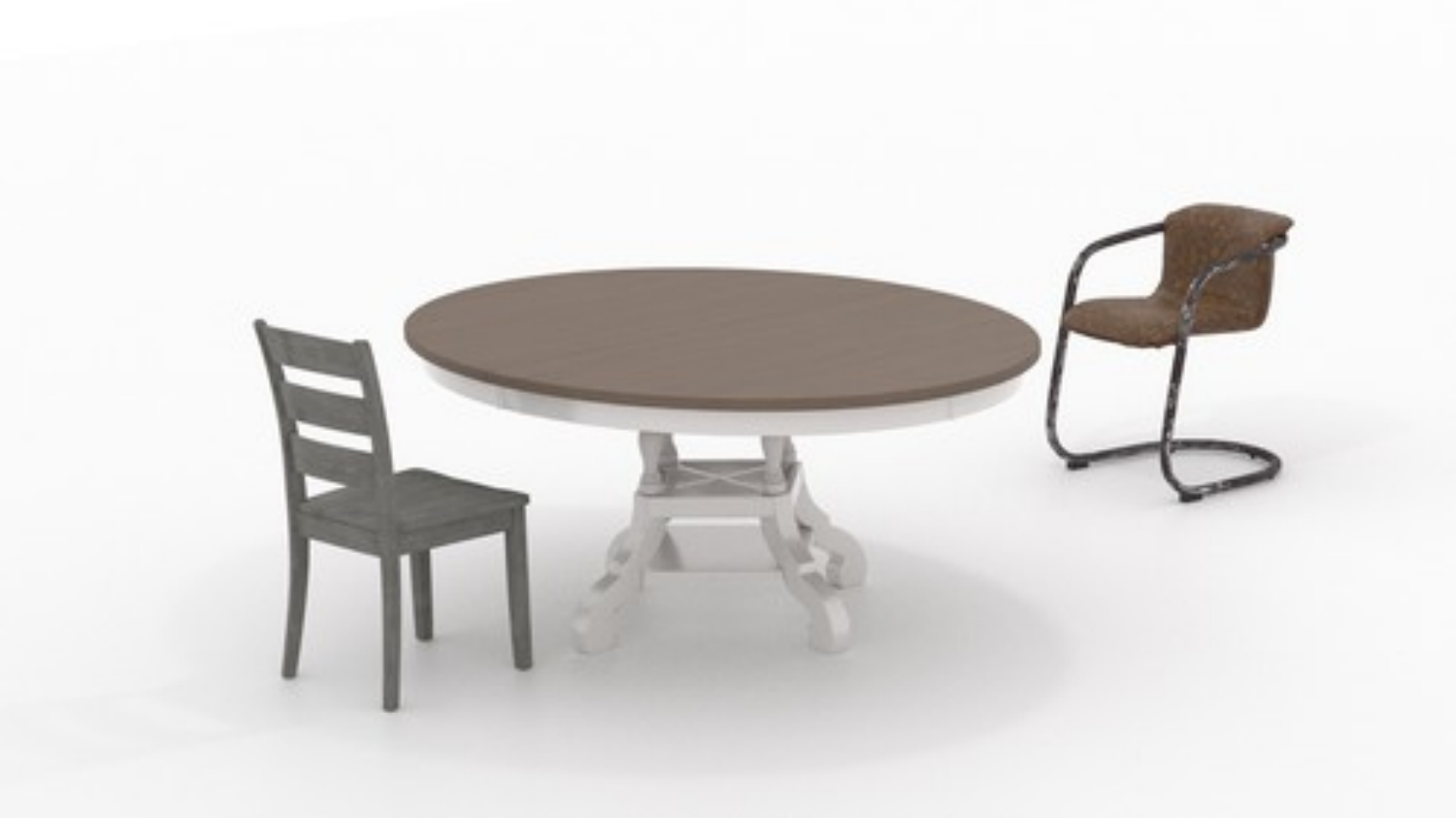} \\
  \raisebox{18mm}{\parbox[b]{2mm}{\hypertarget{fig:morelesscomp:c}{c} }}
  \includegraphics[width=0.45\columnwidth,keepaspectratio]{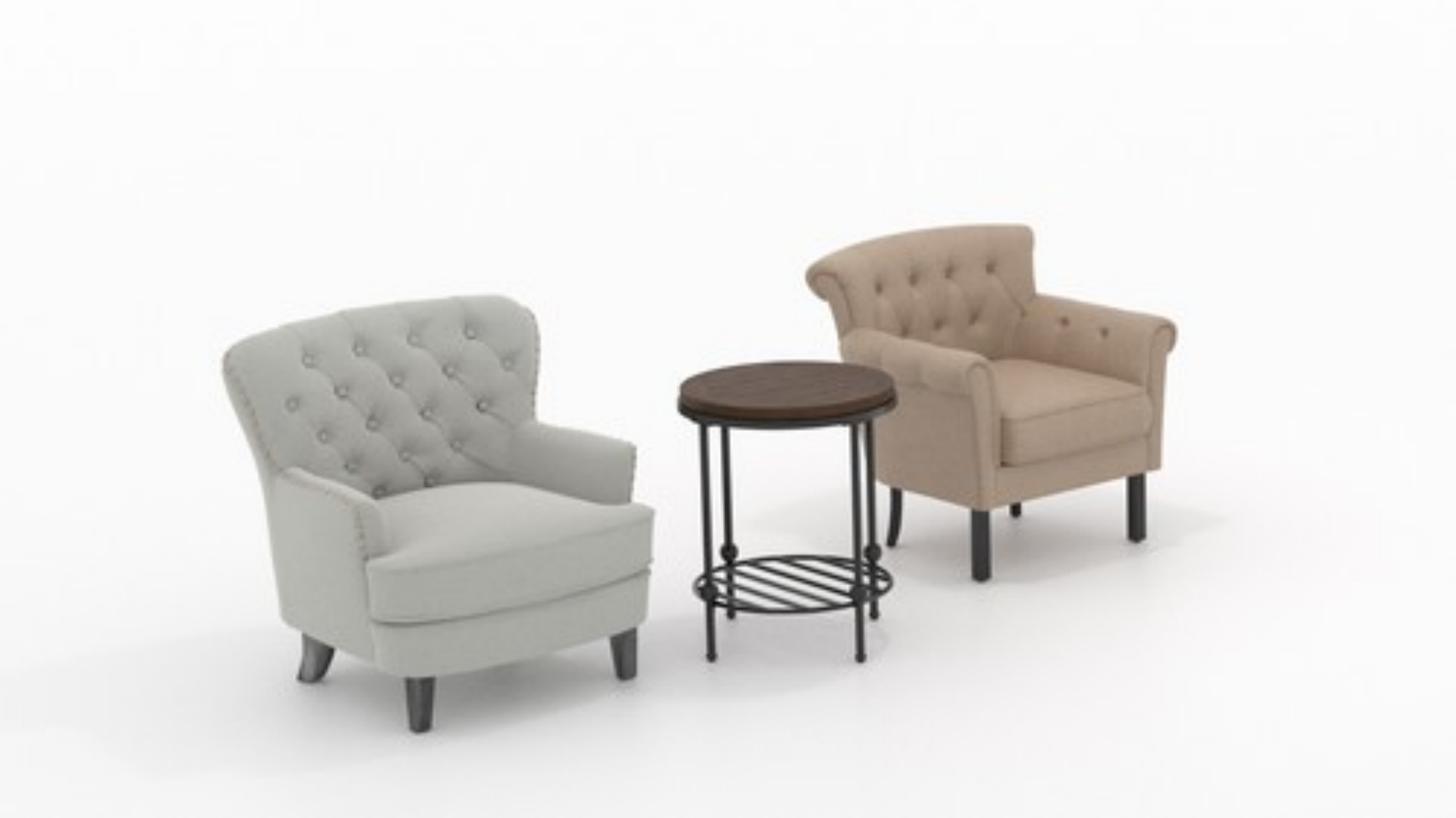} &
  \includegraphics[width=0.45\columnwidth,keepaspectratio]{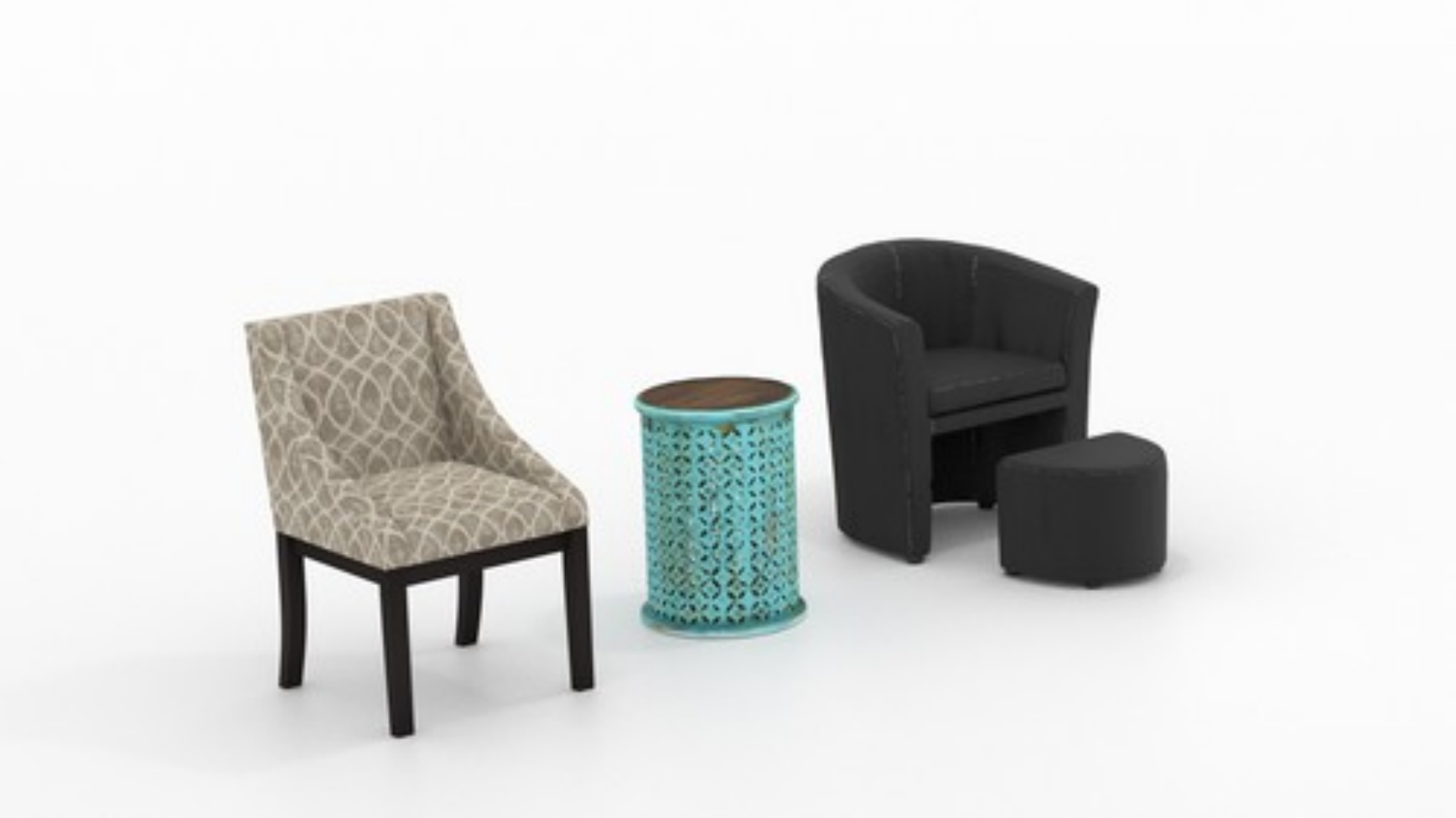} \\
  \raisebox{18mm}{\parbox[b]{2mm}{\hypertarget{fig:morelesscomp:d}{d} }}
  \includegraphics[width=0.45\columnwidth,keepaspectratio]{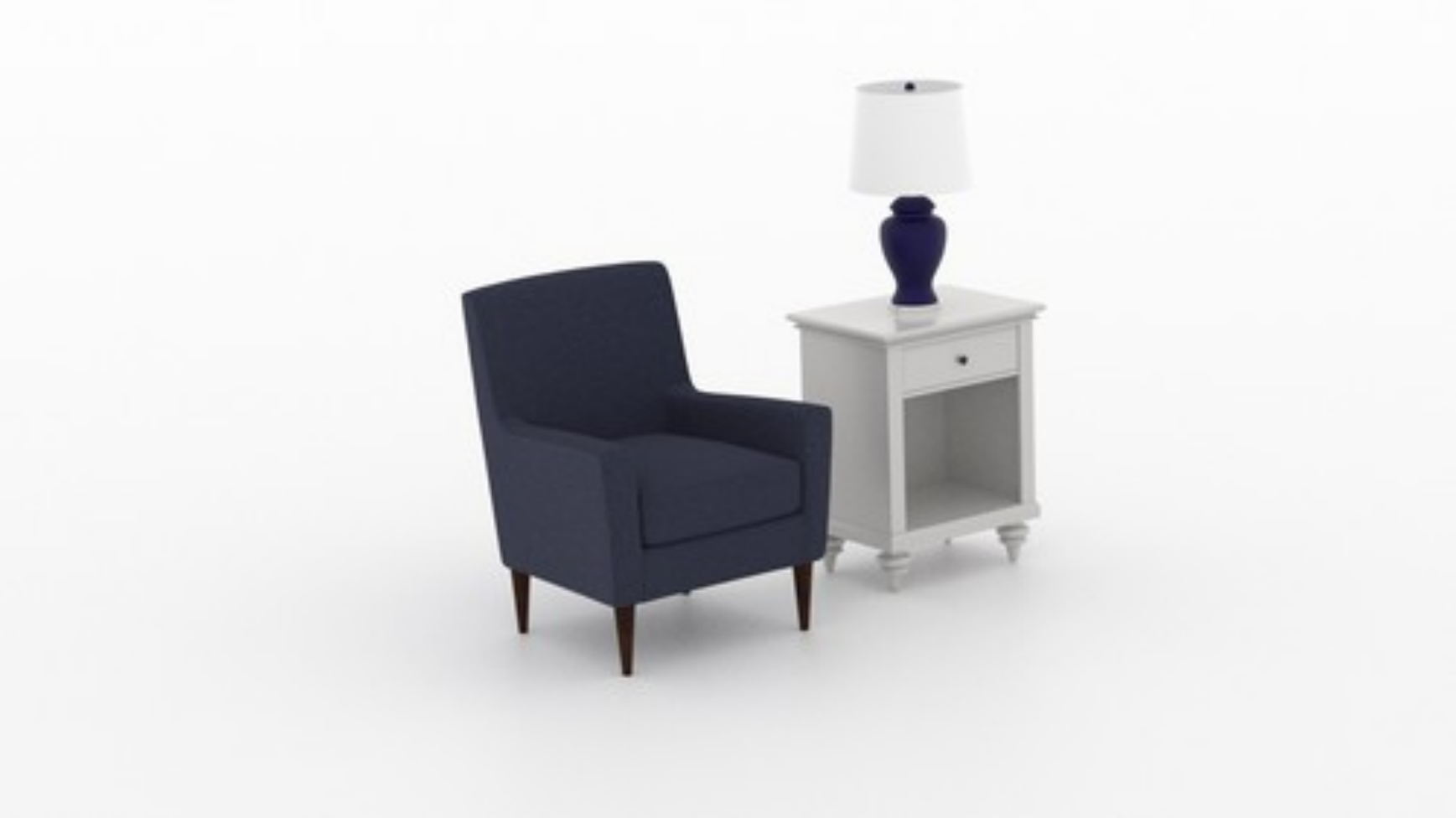} &
  \includegraphics[width=0.45\columnwidth,keepaspectratio]{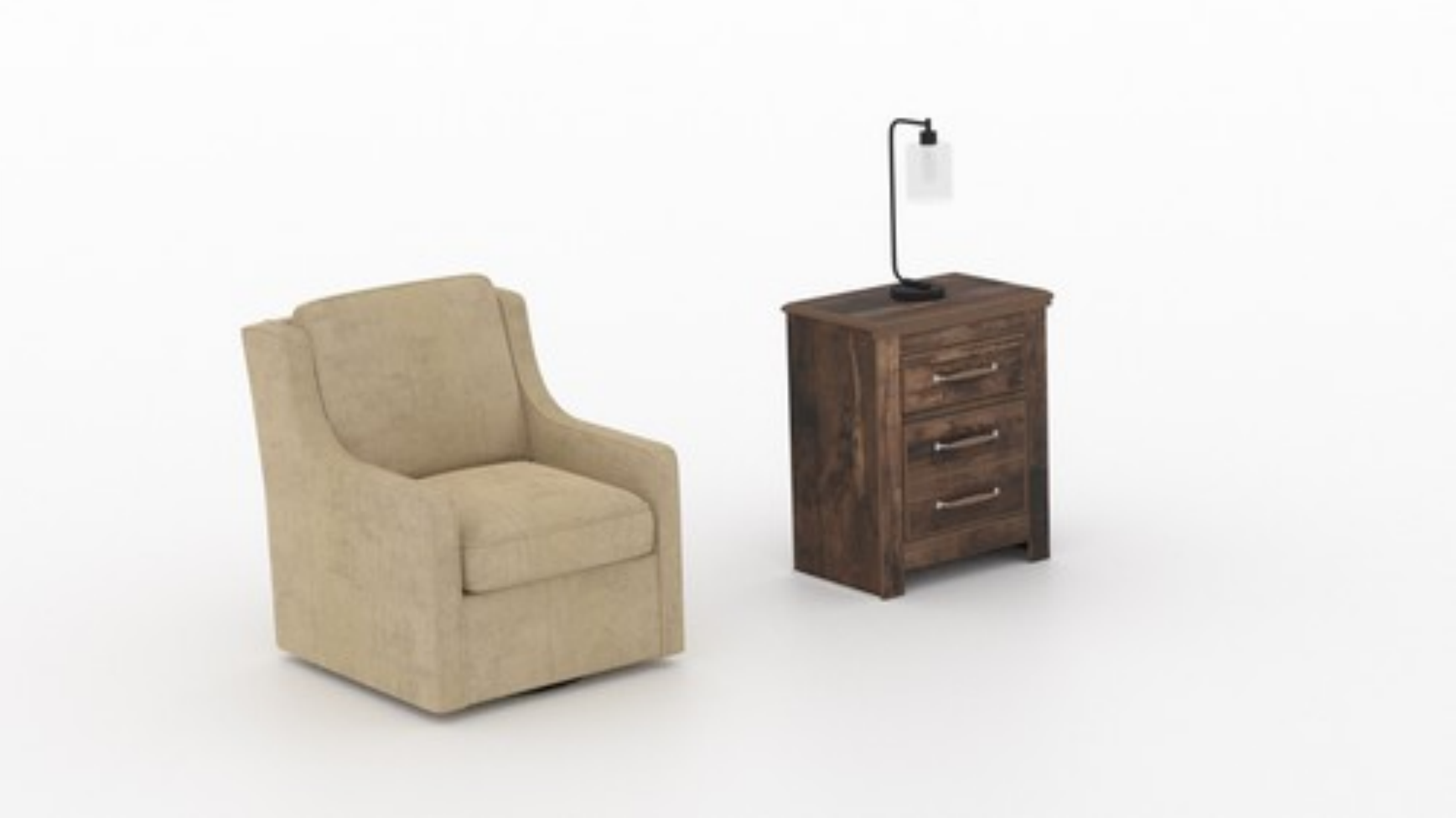} \\
  More Compatible & Less Compatible
\end{tabular}
\caption{Comparing style-compatibility of furniture, where scenes on the left are stylistically more compatible that the one on the right.
We measure compatibility between multiple 3D models using embeddings extracted from our network (Eq.~\ref{eq:entirescene}).
To find whether one collection of 3D models is more style-wise compatible than the other, we calculate the sum of the style-compatibility distances between all object pairs.
In the above examples, each row contains the same object classes, e.g., sofa, and accent chair and a coffee table for the first row.
} \label{fig:morelesscomp}
\end{figure}

\subsection{Experiment: Single Style Reference} \label{sec:single}
Manually searching for 3D models that fit a certain style or reference object is laborious.
We wanted to evaluate if we can assist professional users in creating virtual scenes by constraining available 3D models to a filtered, style-compatible set.
To that end, we asked a professional designer to complete several room building tasks.
For each task, we placed one 3D model in a generic virtual room, which provides a style reference.
The goal is to complete-the-look of a specific scene type, by adding objects from a style-compatible set, which is generated based on the reference object's embeddings (Eq.~\ref{eq:singleseed}).
Such set contains up to $150$ 3D models.
We choose this number experimentally for diversity in object types.
After placing style-compatible objects, users could add minor room object types not supported by out method, such as accessories, clutter, wall art, and other decor objects. Users were also allowed to modify architectural features, such as the floor plan, colors, doors and windows.
Figures~\ref{fig:roomresults} show completed room examples.

A natural question to ask is whether our tool assists in creating style-compatible scenes. We investigated this question by asking non-professionals, since our system is ultimately targeted at such users.
We first educate users by showing multiple images for each style.
Then, we instruct users to assess whether created scenes are style compatible with scores ranging from 1 to 5, where 1 denotes low agreement, and 5 denotes high agreement.
Table~\ref{tbl:singleseed} summarizes their responses.
The results were collected from 14 participants with no prior professional experience in interior design or scene modeling.
The majority of the responses from participants were positive regarding the style-compatibility of the resulting scenes.

\begin{table}[t]
\centering
\begin{center}
\caption{
User study on scenes created based on a single furniture style reference.
For each scene, a user initially selects a seed 3D model.
Our system recommends compatible furniture's which are then added to the scene. Further,
auxillary 3D models such as doors, windows, and clutter were later added to enhance the scene (Figure~\ref{fig:roomresults}).
Column "style" denotes the leading scene style.
Column "\#" denotes the number of objects selected with our recommendation model.
The last column refers to results from our user study,
where we asked users to rank the resulting scene is stylistically compatible, where 1 denotes low agreement, and 5 denotes high agreement. Error bars indicate one standard deviation.
} \label{tbl:singleseed}
\begin{tabular}{ l | r | l | c }
 Scene & \# & Style & Compatible \\
 \hline
 Dining Room &  $2$  & Traditional &
 \multirow{8}{*}{%
 \begin{tikzpicture}[xscale=0.5,yscale=0.38]
   \coordinate [] (nodea) at (0,7.5);
   \coordinate [] (nodeb) at (1,7.5);
   \coordinate [] (nodec) at  (2,7.5);
   \coordinate [] (noded) at  (3,7.5);
   \coordinate [] (nodee) at  (4,7.5);
   \coordinate [label=below:$1$] (nodeaup) at (0,-0.75);
   \coordinate [label=below:$2$] (nodebup) at (1,-0.75);
   \coordinate [label=below:$3$] (nodecup) at  (2,-0.75);
   \coordinate [label=below:$4$]  (nodedup) at  (3,-0.75);
   \coordinate [label=below:$5$]  (nodeeup) at  (4,-0.75);
   \draw [densely dotted] (nodea) -- (nodeaup);
   \draw [densely dotted] (nodeb) -- (nodebup);
   \draw [densely dotted] (nodec) -- (nodecup);
   \draw [densely dotted] (noded) -- (nodedup);
   \draw [densely dotted] (nodee) -- (nodeeup);
   \draw[|-|, thick, black ] (0.973, 7) to (3.3, 7);
   \node[circle,draw=black,fill=white, scale=0.5] (a) at (2.14, 7) {};
   \draw[|-|, thick, black ] (0.83, 6) to (3.17, 6);
   \node[circle,draw=black,fill=white, scale=0.5] (a) at (2 ,6) {};
   \draw[|-|, thick, black ] (1.91 ,5) to (3.79 ,5);
   \node[circle,draw=black,fill=white, scale=0.5] (a) at (2.85 ,5) {};
   \draw[|-|, thick, black ] (1.05 ,4) to (3.95 ,4);
   \node[circle,draw=black,fill=white, scale=0.5] (a) at (2.5 ,4) {};
   \draw[|-|, thick, black ] (0.85 ,3) to (3.43 ,3);
   \node[circle,draw=black,fill=white,scale=0.5] (a) at (2.14 ,3) {};
   \draw[|-|, thick, black ] (1.27 ,2) to (3.57 ,2);
   \node[circle,draw=black,fill=white,scale=0.5] (a) at (2.42 ,2) {};
   \draw[|-|, thick, black ] (1.72 ,1) to (3.7 ,1);
   \node[circle,draw=black,fill=white, scale=0.5] (a) at (2.71 ,1) {};
   \draw[|-|, thick, black ] (2.19 ,0) to (3.65 ,0);
   \node[circle,draw=black,fill=white, scale=0.5] (a) at (2.92 ,0) {};
 \end{tikzpicture} } \\
 Junior Bedroom & $4$ & Modern &  \\
 Meeting Room & $6$  & Coastal & \\
 Family Room & $6$   & Modern & \\
 Asian Resturant &  $6$ &  Modern & \\
 Hip Studio & $8$  &  Modern & \\
 Lounge &  $8$ & Traditional &  \\
 Living Room &  $9$ & Cottage & \\
 \hline
\end{tabular}
\end{center}
\end{table}

\subsection{Experiment: Multiple Style References} \label{sec:multiref}
Next, we evaluated our method in case of multiple inputs (Section~\ref{sec:curating}).
We first studied our style-compatibility in case of object
triplets of several common furniture combinations.
To generate the triplets, we selected furniture classes, such as sofas, accent chairs, and tables.
Then, for all objects that belong to such classes,
we measured the distance compatibility for permutations created by selecting objects
from each class (Eq.~\ref{eq:entirescene}).
For each of the combinations in Figure~\ref{fig:morelesscomp},
the more style-wise compatible a furniture is, the less the total embedding distance is:
\begin{itemize}[label={},leftmargin=*]
  \item \textbf{\hyperlink{fig:morelesscomp:a}{a}.}
  Sofa, coffee tables and accent chair: The scene on the left has a more modern style, with color and textures complementing each other. The scene on the right is mixed between the somewhat traditional coffee table, and modern accent chair.
  \item \textbf{\hyperlink{fig:morelesscomp:b}{b}.}
  Dining table and a pair of dining chairs:
  Note, the configuration on the left is more traditional in style. The configuration on the right is a combination of a traditional-looking table with a more modern chair.
  \item \textbf{\hyperlink{fig:morelesscomp:c}{c}.}
  Two different accent chairs and an end table:
  The two accent chairs and an end table on the left are leaning to the traditional style, and are more similar in shape. On the right, the style is not conclusive, and the color of each object does not complement each other.
  \item \textbf{\hyperlink{fig:morelesscomp:d}{d}.}
  Accent chair, nightstand and table lamp:
  For the arrangement on the left, our method picked a lamp that has similar color to the accent chair, while the nightstand does not detract from the scenes color harmony. On the right, the furniture might be considered compatible, however the accent chair and nightstand are not quite stylistically compatible.
\end{itemize}

\begin{figure}[bt]
\centering
\raisebox{20mm}{\parbox[b]{2mm}{\rotatebox[origin=b]{90}{\emph{Initial Scene}}}}
\includegraphics[width=0.93\columnwidth,keepaspectratio]{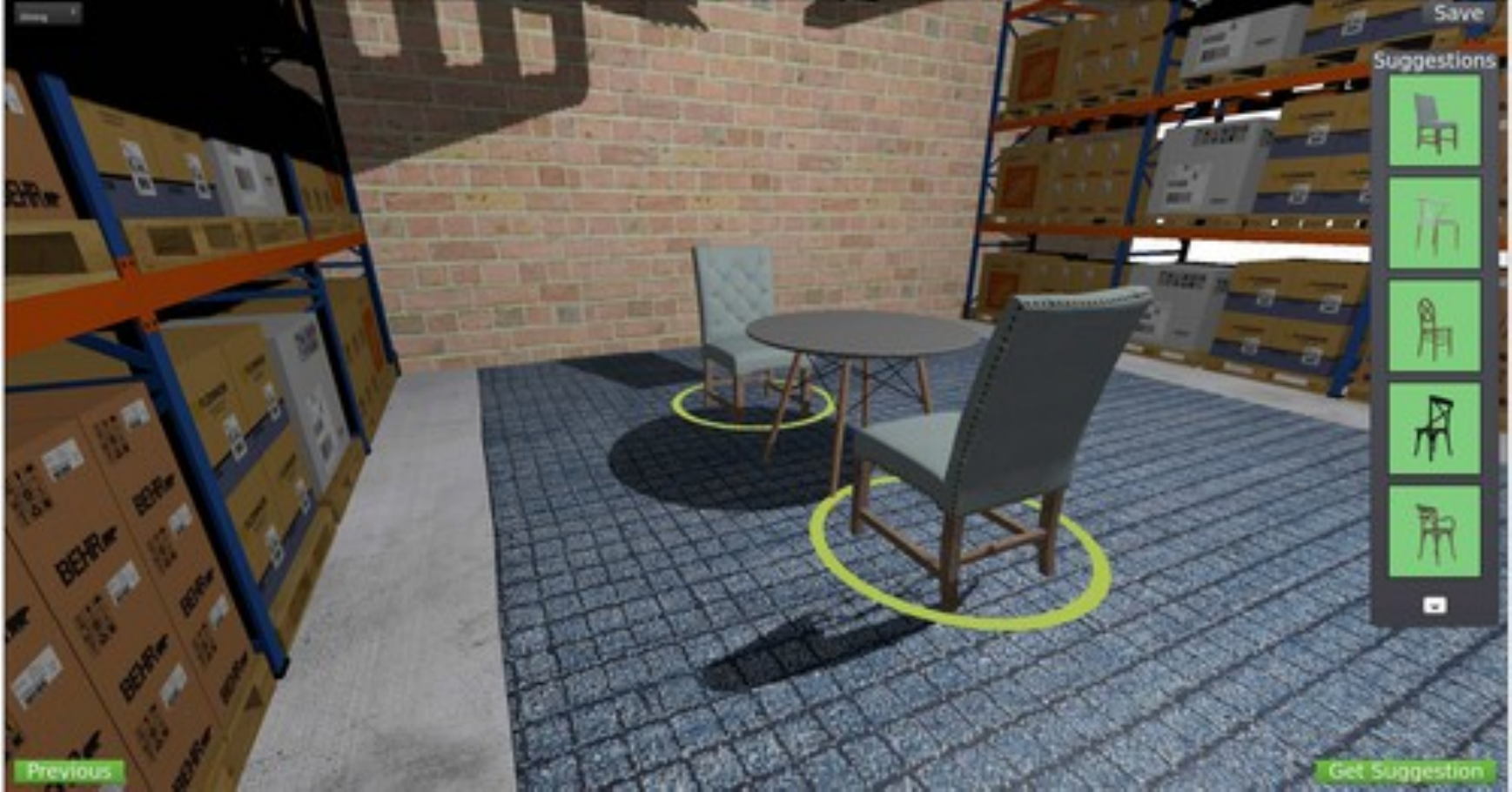}\\
\raisebox{20mm}{\parbox[b]{2mm}{\rotatebox[origin=b]{90}{\emph{After Selection}}}}
\includegraphics[width=0.93\columnwidth,keepaspectratio]{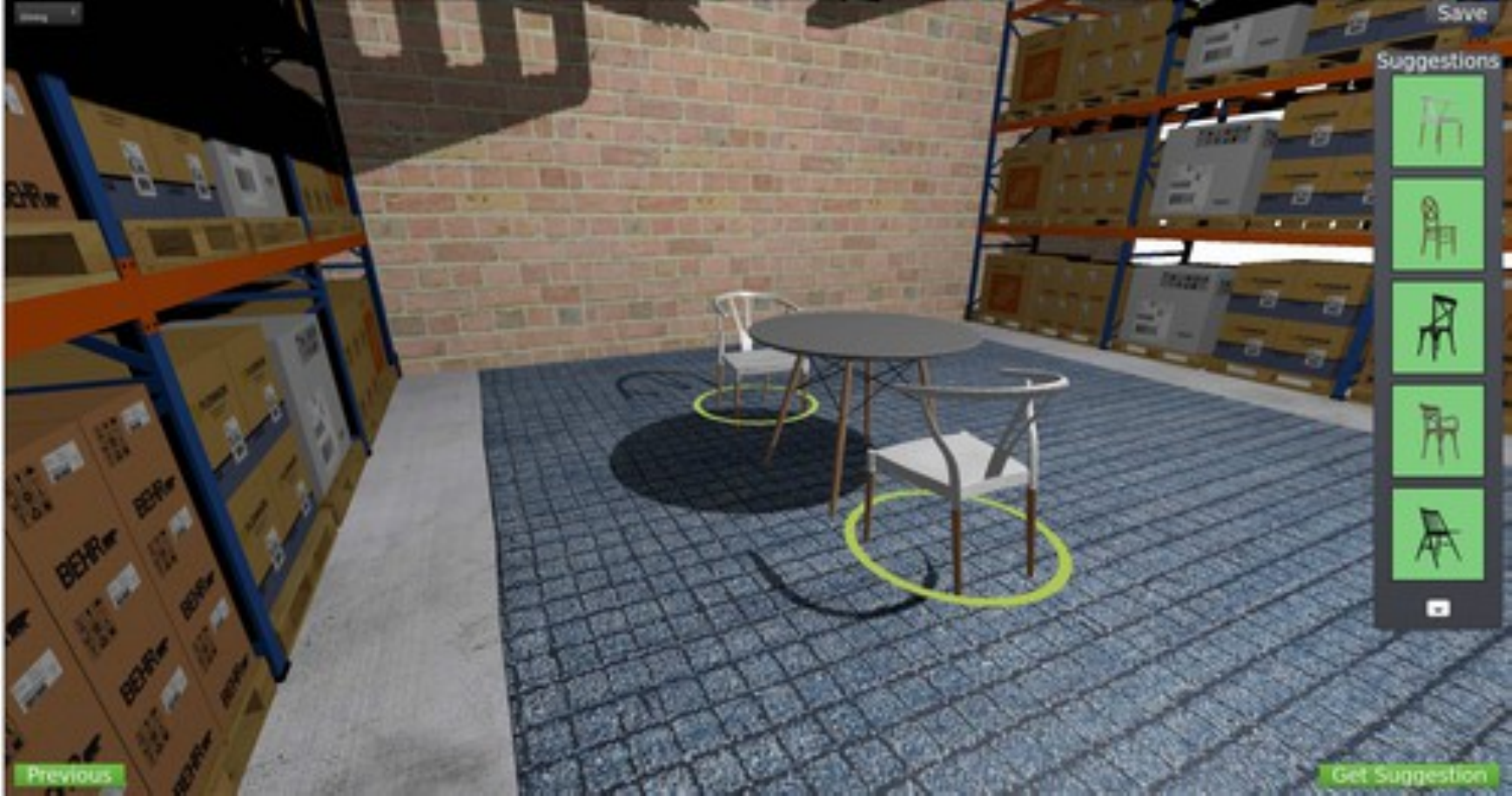}
\caption{We implemented an interactive system for suggesting style-compatible furniture.
The top image shows the initial scene configuration, where the user has selected two chairs to be swapped, annotated with circles.
By clicking, "Get Suggestion", suggestions are generated on the right.
The bottom image shows the result of swapping chairs using our style-compatible suggestions.
}
\label{fig:interactive}
\end{figure}

\begin{figure*}[t]
  \centering
  \includegraphics[width=0.54\textwidth]{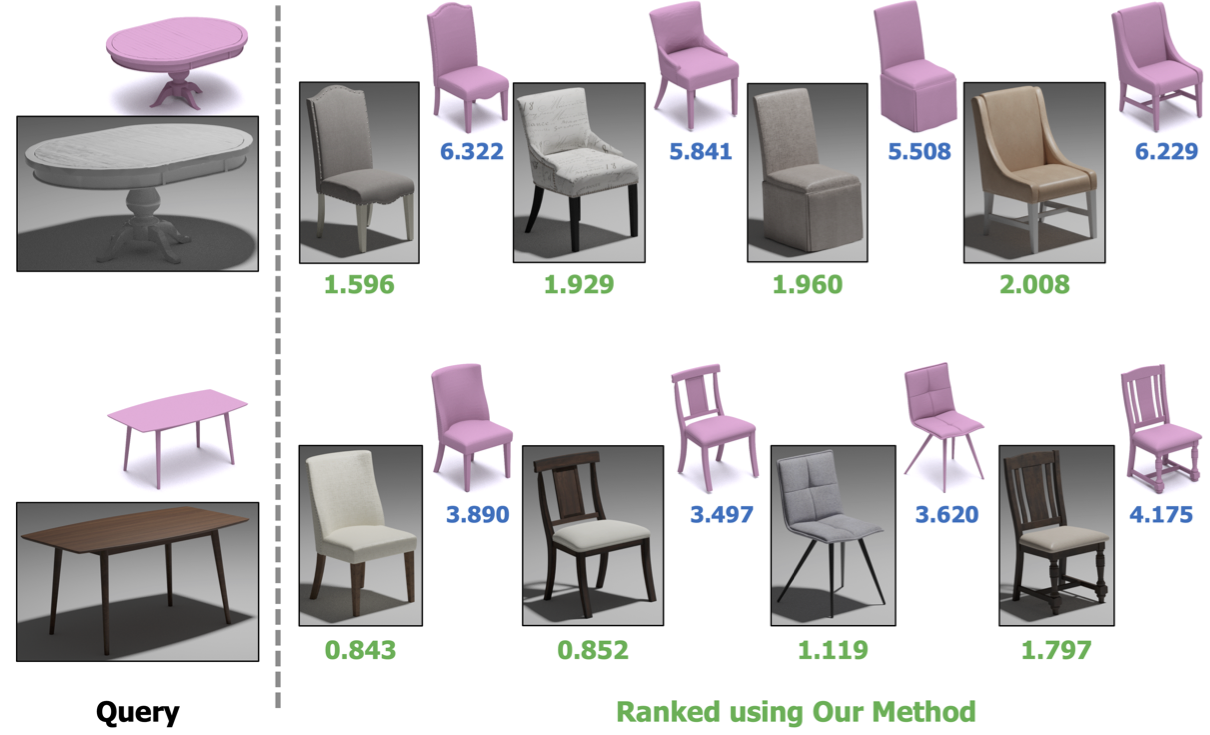}
  \includegraphics[width=0.44\textwidth]{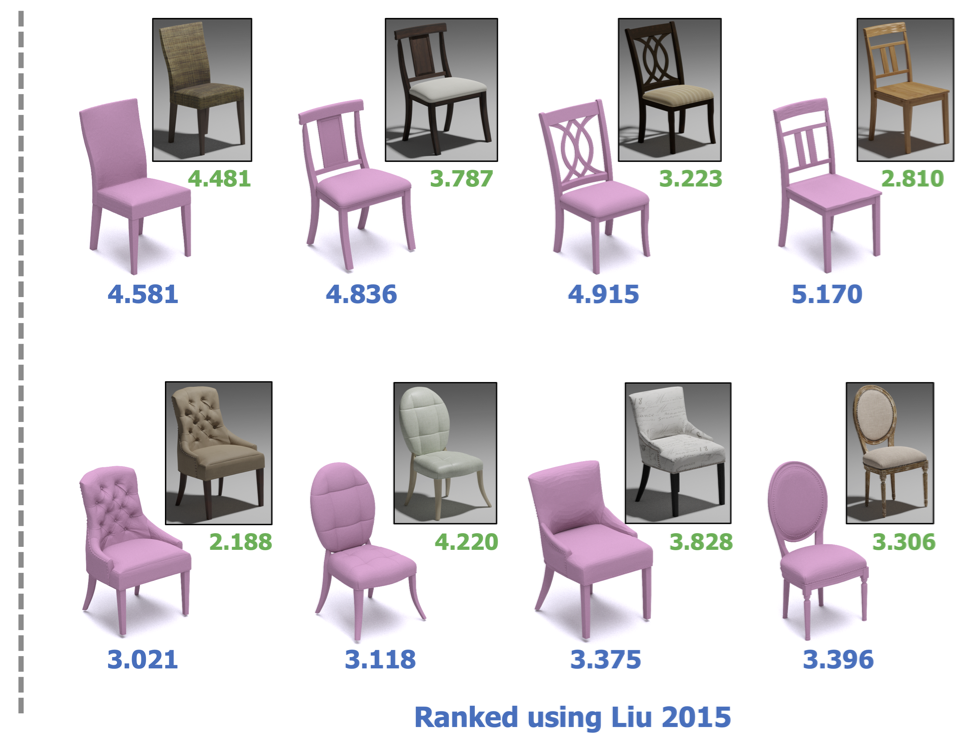}
  \caption{Style-aware shape retrieval. Given a query 3D model such as a dining table, we compare the suggestions from our method (green) which use image features such as color, texture and material with~\cite{liu2015style} (blue) which use geometric features computed directly on the 3D shape.
  For both methods we show their top 4 most stylistically compatible chairs along with their compatibility distances to the query table. We also show for all chairs the compatibility distance if the other method was used, in their top inset. Note, both methods report the compatibility distances at different scales but share similar meaning of lower values being more compatible.}
\label{fig:style_distances}
\end{figure*}

\subsection{Application: Style-Aware Scene Modeling} \label{sec:scenemodeling}
To further experiment with style compatibility for multiple reference 3D models, we implemented a web-based interactive system for style-aware augmentation of scenes.
Our system provides user with style-compatible furniture suggestions.
We based our application on the proposal of~\cite{lun2015elements}, which demonstrates style-based suggestions for scene modeling, based on a context furniture.
Similar to their suggestion,
we present style-compatible suggestions with a list,
in which furniture is ranked according style compatibility fit, from most compatible to least compatible (Eq.~\ref{eq:multiseed}).
To generate such list, users select existing scene objects.
Such selection constraints suggestions to the type of the selected object, e.g., if the user selects a coffee table,
the suggestion list will only display coffee tables.
After user selections, the suggestions list is interactively generated, allowing users to swap scene objects with 3D models in the suggestion list (Figure~\ref{fig:interactive}).

\begin{wrapfigure}{l}{0.34\columnwidth}
  \captionsetup{labelformat=empty,font=footnotesize}
  \begin{center}
    \includegraphics[width=0.35\columnwidth]{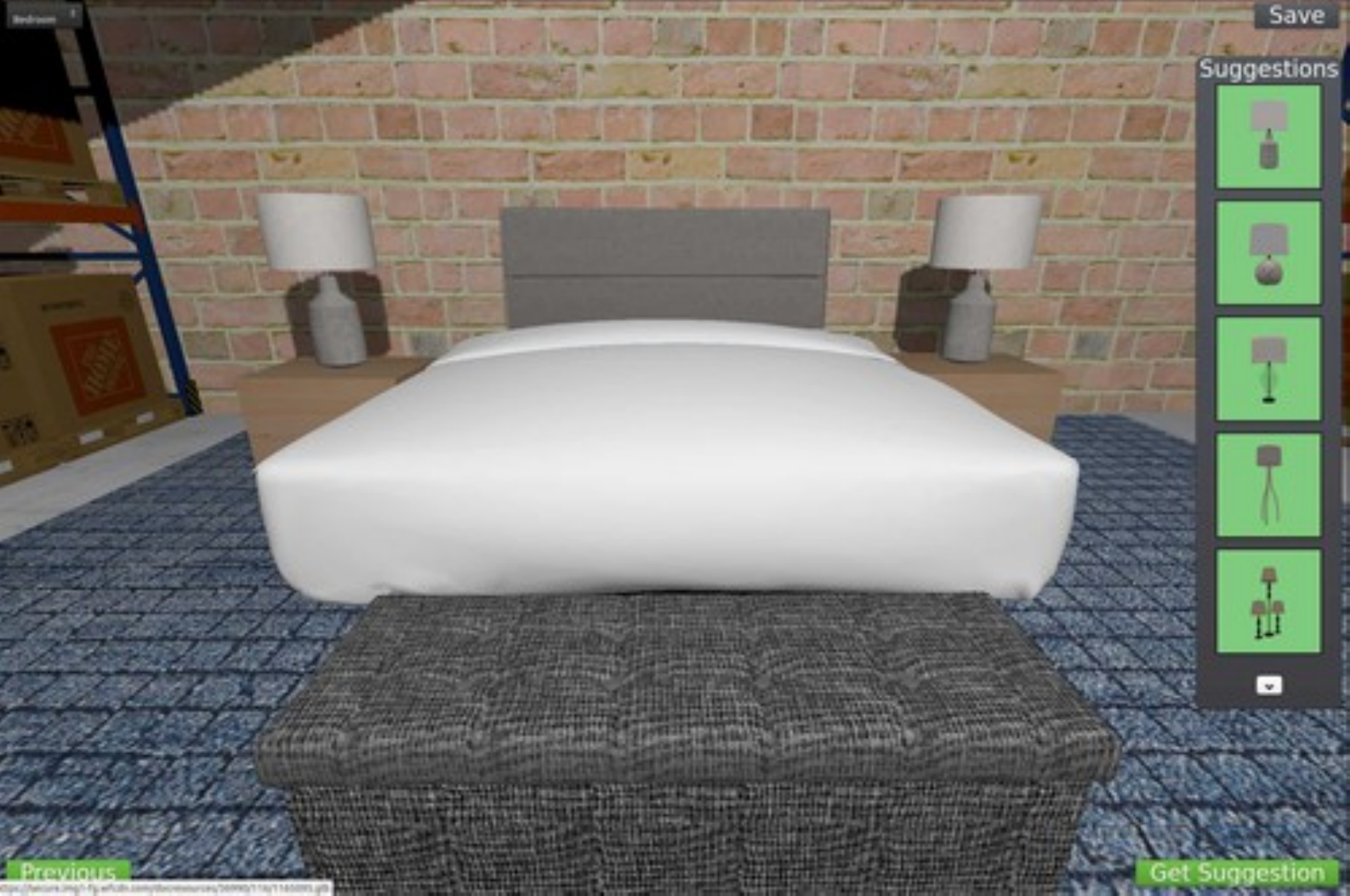}
  \end{center}
  \vspace{-3pt}
  \caption{\emph{Similar-style lamp suggestions for a bedroom.}}
\end{wrapfigure}
\vspace{-10pt}

Our system allows several user interactions.
To anchor the visual experience, we set the background to be static for all interaction modes.
We use a warehouse-like space to be room-function and style generic, so that architectural details are static.
Initially, users either select an existing scenario, or create a new one in a free flow interactive manner.
After selecting a scenario,
users can either add new furniture, or swap existing furniture.
Users control the placement of furniture, in terms of position, rotation, and can also place on top of other furniture, e.g., a lamp on top of a night stand. For style compatible suggestions, users select a target object in the scene. Users may then click on "Get Suggestion" button for style-compatible suggestions, and swap selected furniture with the most-compatible furniture.
We allow users to revert their actions, and save their work.

To evaluate the effectiveness of our system, we experimented with several scenarios,
including bedroom, dining and living room furniture arrangements (Figure~\ref{fig:interactive}).
Interacting with our system is simple and requires little to no training.
Through clicking and dragging actions, users have intuitive control,
without needing to be aware of the style labels.
Usability testing would be needed to further substantiate the intuitive nature of our user interface.
Our system generated style-compatible suggestions in real-time,
which we recorded (see supplemental video).

\subsection{Comparison with Previous Work}
We compare the results from our image-based style framework with~\cite{liu2015style} which uses geometric features of 3D models to compute style compatibility across different furniture classes.
We show this comparison in Figure~\ref{fig:style_distances} where we retrieve the top 4 stylistically compatible chairs, along with their compatibility distance, to a query dining table using both methods. We make the following observations:
\begin{itemize}
\item Our system is able to successfully retrieve stylistically compatible chairs for the query table model. For both the tables, the first 4 chairs from the left are stylistically more compatible in terms of color and texture. Whereas the last 4 chairs do not complement the tables, reflected in the high compatibility distance from our method.
\item  Style suggestions based purely on geometric approaches differ from image-based frameworks such as ours. This is natural as while our framework focuses on color and texture of furniture pieces in an image, geometric-based approaches use their 3D models to compute other properties such as curvature, surface area, and bounding boxes. This can also be seen in the last 4 chairs of the second column where the ordering of the chairs by our compatibility distance is different than the ranking by~\cite{liu2015style}.
\item Models which are close in the geometric embedding space, may not be as close in the image embedding space. For example, in second row, the 5th and 6th chair from the left have a minor difference in their geometric compatibility distance, but texturally differ, a quality which is reflected in a large image compatibility distance. Such style distance has foundation since a model can exhibit different styles by using different textures while retaining the same 3D geometry.
\end{itemize}

\section{System Scope and Limitations}
Our system provides style-based suggestions for input furniture, working with both images and 3D models.
Since style-compatibility of 3D furniture models is inferred from images,
suggested 3D models might not fit together for non-style reasons, which we summarize below:

\emph{Functionality and Synthesis.}
Considering we infer style-compatibility from images,
our method does not take furniture functionality into account, e.g., bedroom versus living room furniture. Similarly, we do not incorporate furniture arrangement and synthesis~\cite{yu2011make,weiss2018fast,zhou2019scenegraphnet} when modeling 3D scenes. While important, such methods are orthogonal to our style-focused scope.

\emph{Scene Parameters.}
3D scenes vary in terms of conditions, ranging from lights~\cite{jin2019lighting}, camera setup~\cite{liu2015composition}, to contents. Such conditions have a perceptual affect on style.
For example, in case of more multiple scene objects, their color and combined appearance might not be harmonious.
However, experimentally we found that selecting from a set of style-compatible furniture is sufficient for creating an harmonious look and ambience.

\emph{Architectural Elements.}
We do not consider scene architectural space elements, such as wall and floor color. Nevertheless, our current approach demonstrates that existing scene objects provide an anchor for selecting new furniture pieces for a room.

\emph{Interaction and Visual Quality.}
We implemented our system to work within a web-based framework.
Currently, such web-based frameworks limits lighting configurations, texture, and mesh complexity.
Such frameworks do not easily incorporate quality real-time rendering. %
Additionally, since our goal is to support multiple users, visual quality is set to allow real-time performance for most web users, including mobile phones.
However, since all our 3D models are real-time PBR models in GLTF format, our framework enables cross-platform usage with web and other interactive use cases.

\emph{Compatibility of Furniture from Different Styles.}
Furniture items from different style categories might be suggested to users.
For example, a modern furniture piece could be suggested for a traditional scene, since our network learns the abstract notion of style from pixels in order to satisfy expert labels.
Since style is subjective, the resulting scene might be satisfactory. Additionally, our system allows users to edit and select preferred furniture with a user interface.

\emph{Affects of Curated Imagery on Style Estimation.}
As mentioned in Section~\ref{sec:ret3d}, style compatibility between two furniture pieces is calculated via their minimum image embedding distance.
Images depict furniture in the context of a styled scene, with multiple surrounding features, such as floor, wall, decor and other furniture items.
Our system considers the entire image, including the surrouding features, when predicting style.
Currently, we do not experiment with the affects of individual surrounding features on percieved style, since such features may be multiple and high-dimensional.
Nevertheless, our system successfully learns style from images, using both the surroundings and the furniture, as we show qualitatively in Figures~\ref{fig:roomresults}, \ref{fig:morelesscomp} and \ref{fig:style_distances}, and quantitatively in Figure~\ref{fig:v1_vs_v2_4class}.

\begin{table}[tb]
\caption{Similar style furniture images. Given a context image on the left, our model recommends similar style furniture. Here, we demonstrate recommending a specific furniture type for context furniture.
From top to bottom: given a TV stand, entertainment center, bench, and bookcase on the left, we recommend bookcases (x2), accent chairs, and sofas on the right.} \label{tbl:room}
\vspace{-2mm}
  \begin{center}
  \normalsize
  \begin{tabular}
      {l!{\color{amethyst}\vrule}!{\color{amethyst}\vrule}l} \emph{Context}  & \emph{Ranked Results} \\
      \noalign{\global\arrayrulewidth=0.2mm}
      \arrayrulecolor{amethyst}\hline
      {\includegraphics[height=1.30cm,keepaspectratio]{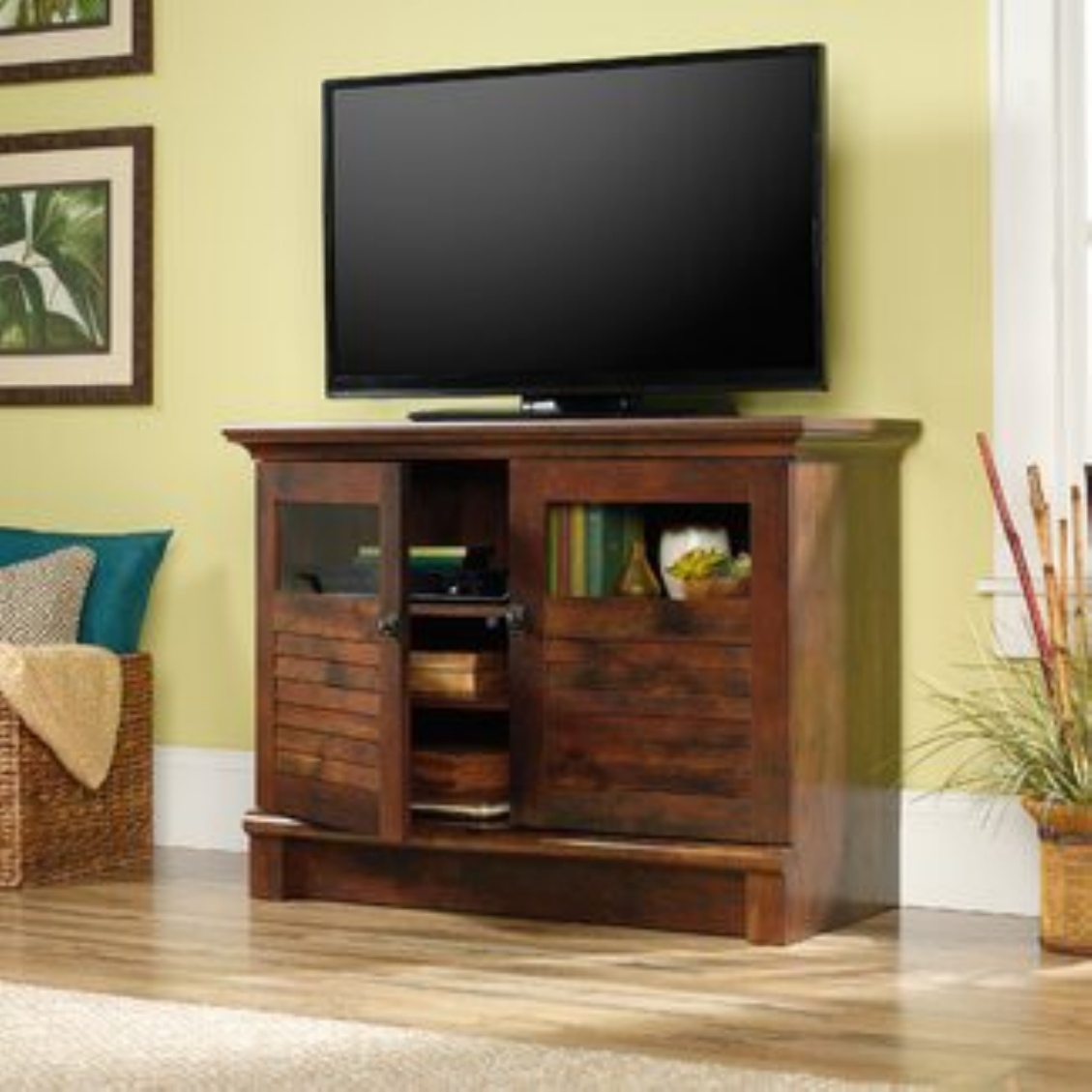}} &
      {\includegraphics[height=1.30cm,keepaspectratio]{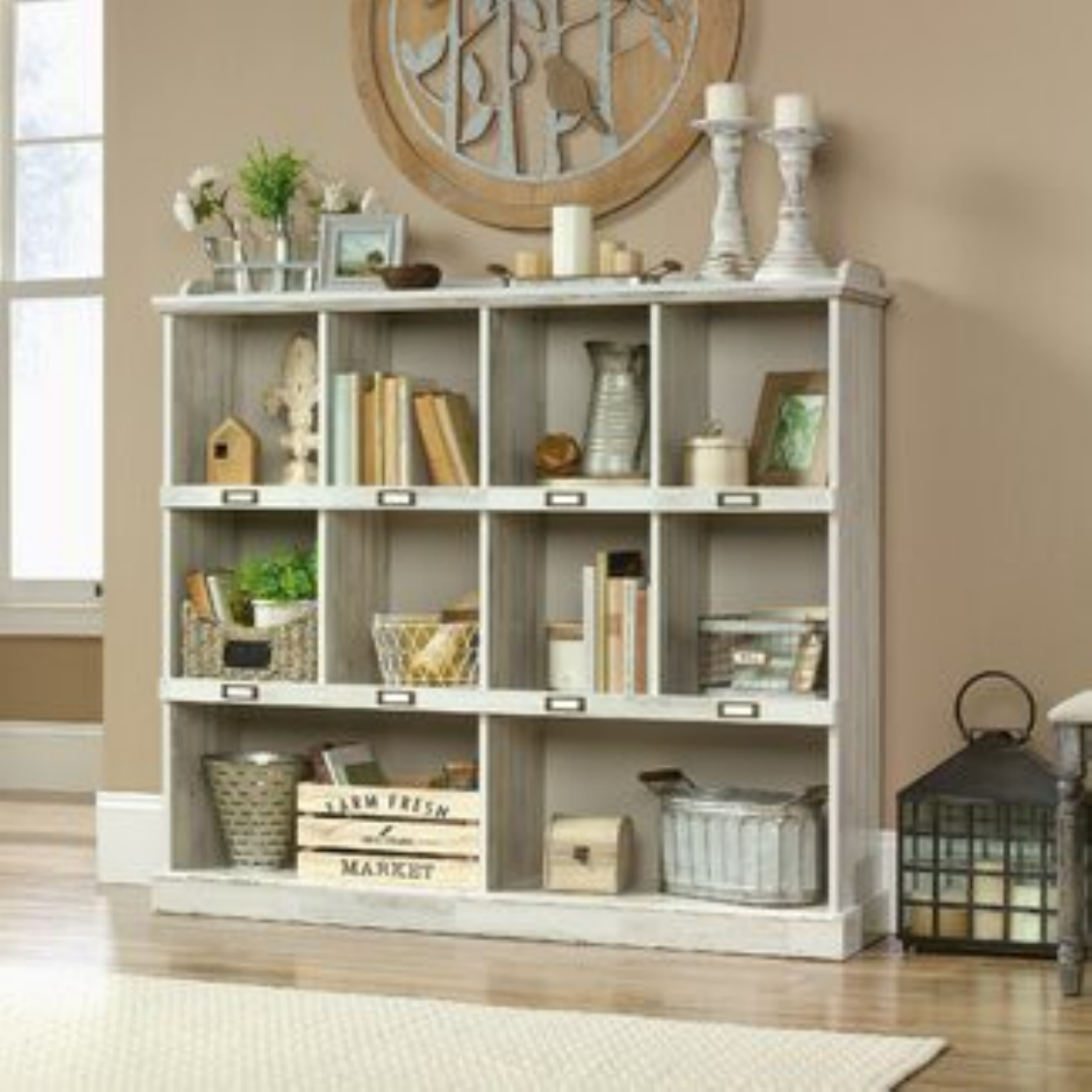}}
      {\includegraphics[height=1.30cm,keepaspectratio]{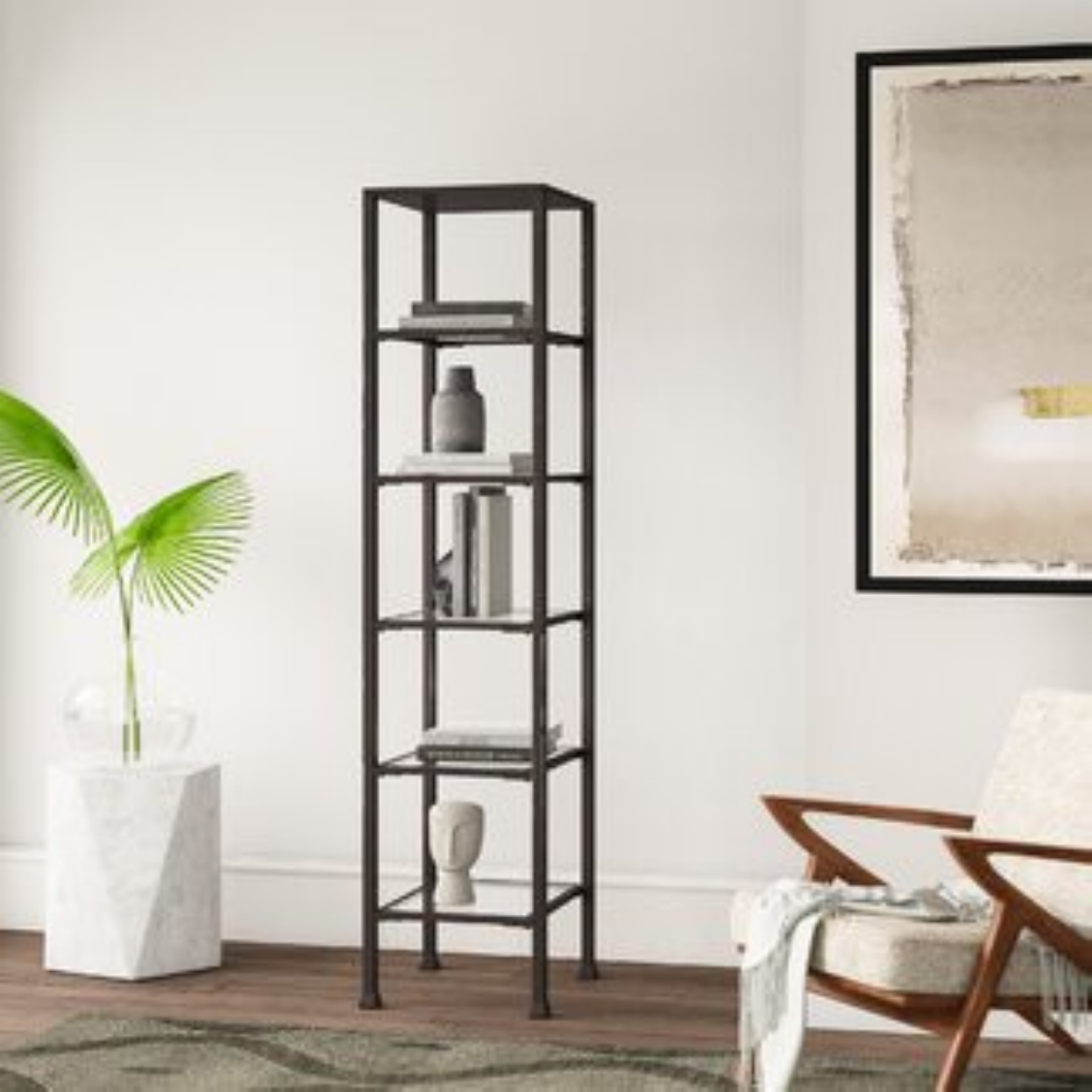}}
      {\includegraphics[height=1.30cm,keepaspectratio]{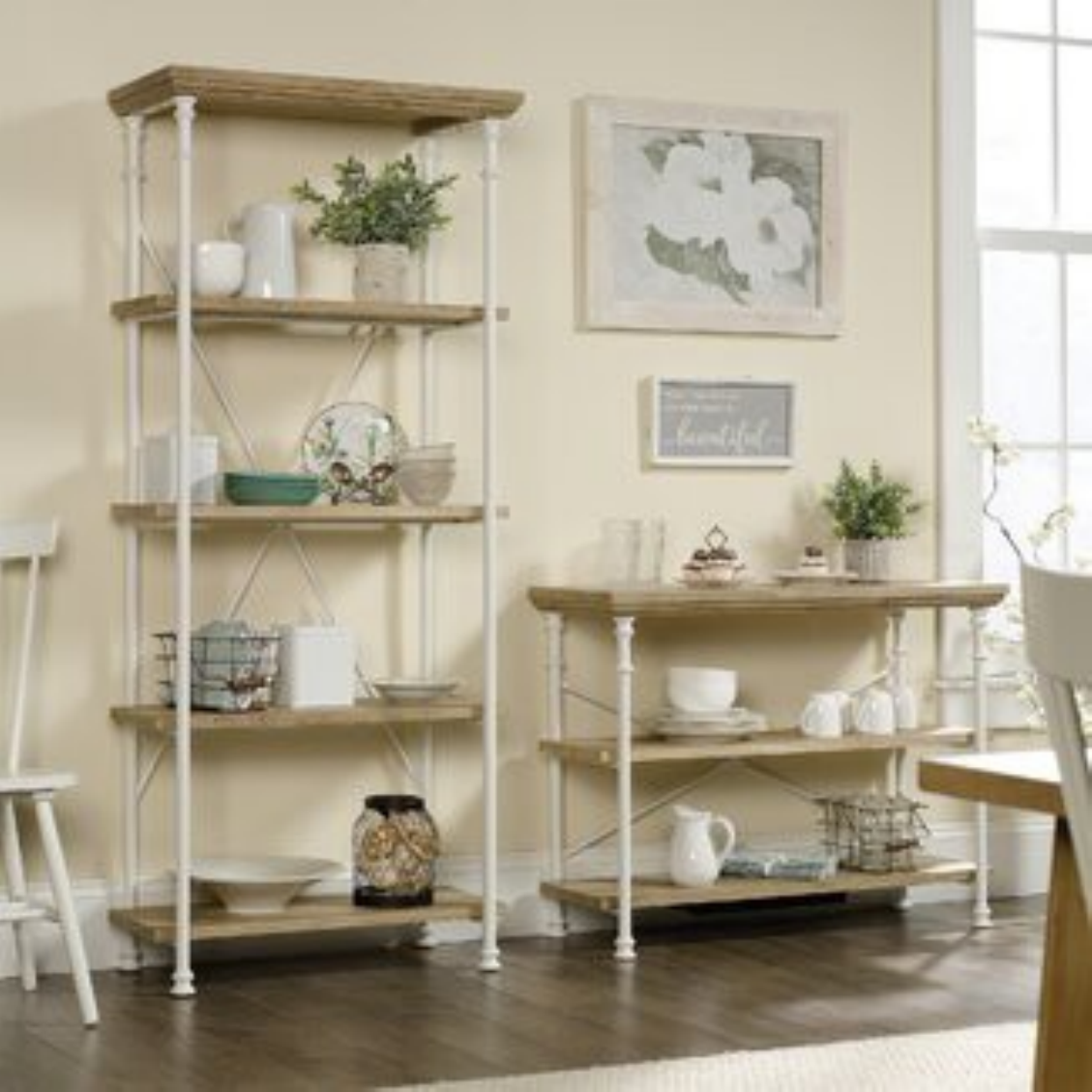}}
      {\includegraphics[height=1.30cm,keepaspectratio]{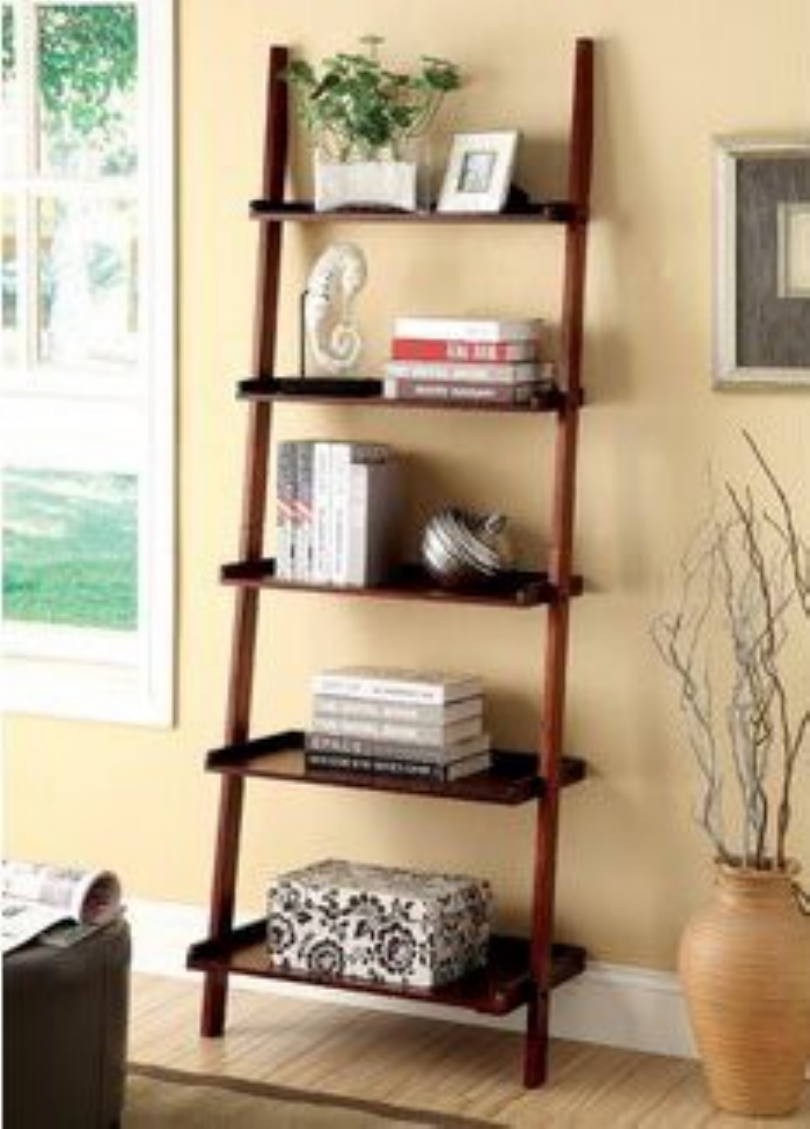}}
      {\includegraphics[trim=0 0 30 0, clip,height=1.30cm,keepaspectratio]{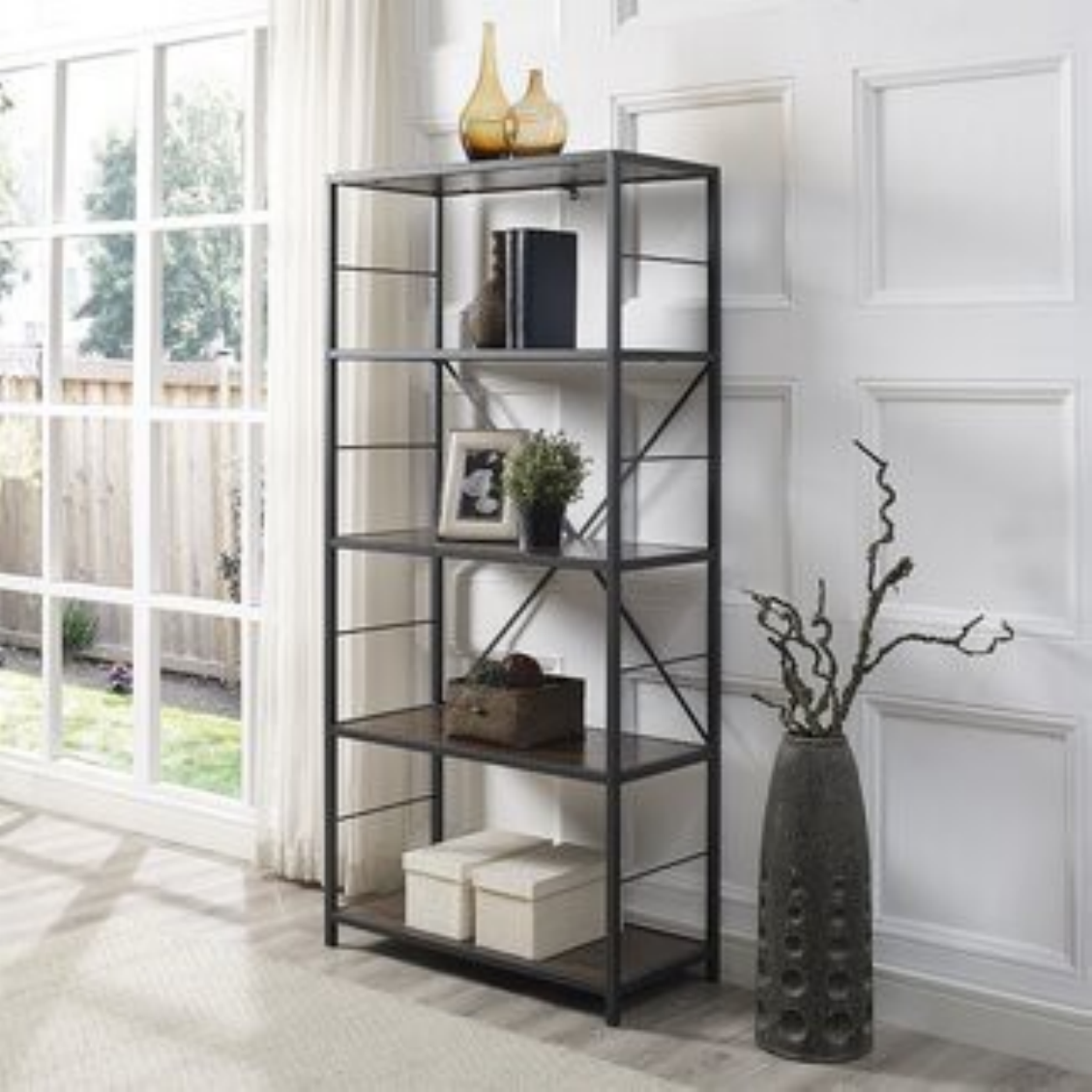}} \\[-0.03in]
      \hline
      {\includegraphics[height=1.30cm,keepaspectratio]{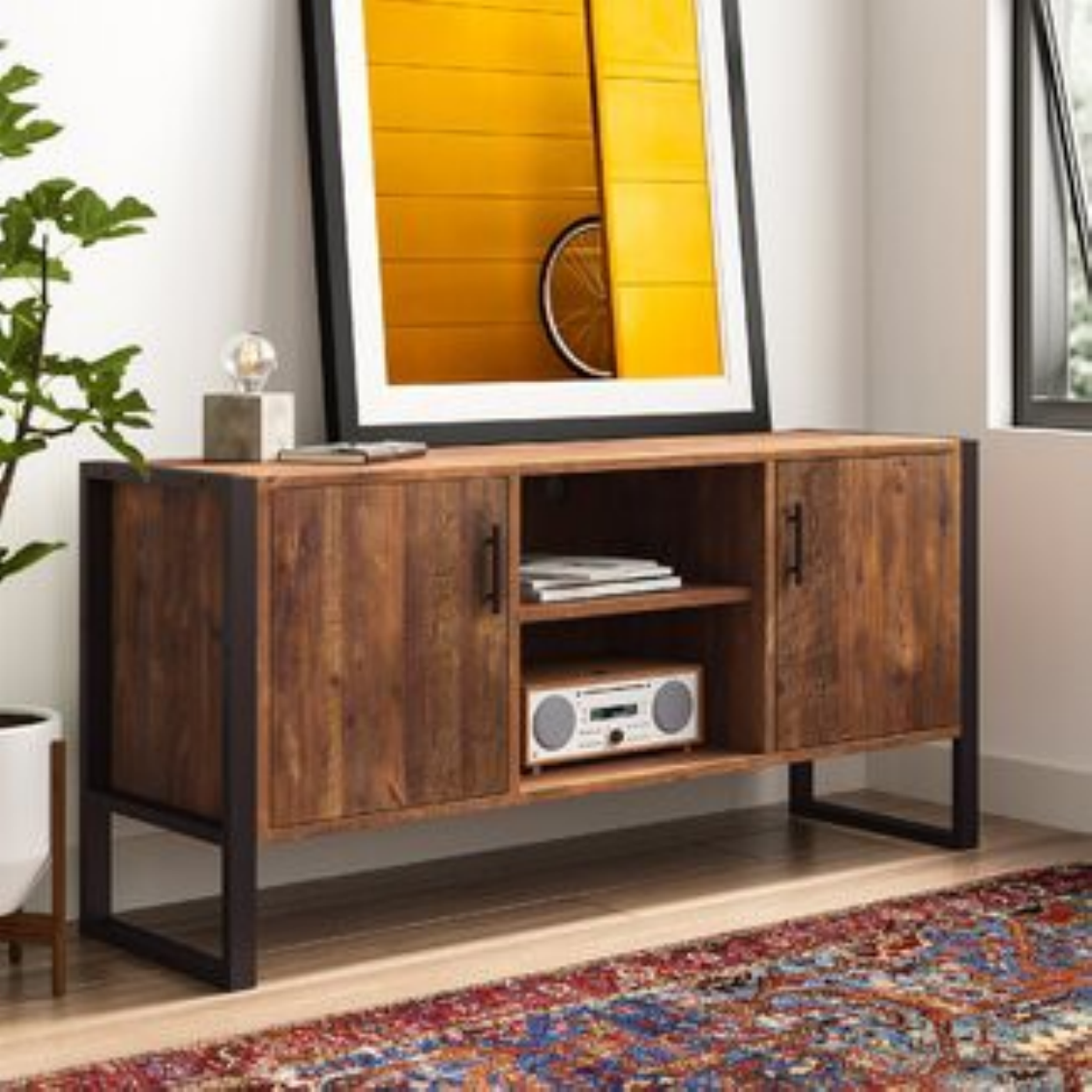}} &
      {\includegraphics[trim=20 0 20 0, clip,height=1.30cm,keepaspectratio]{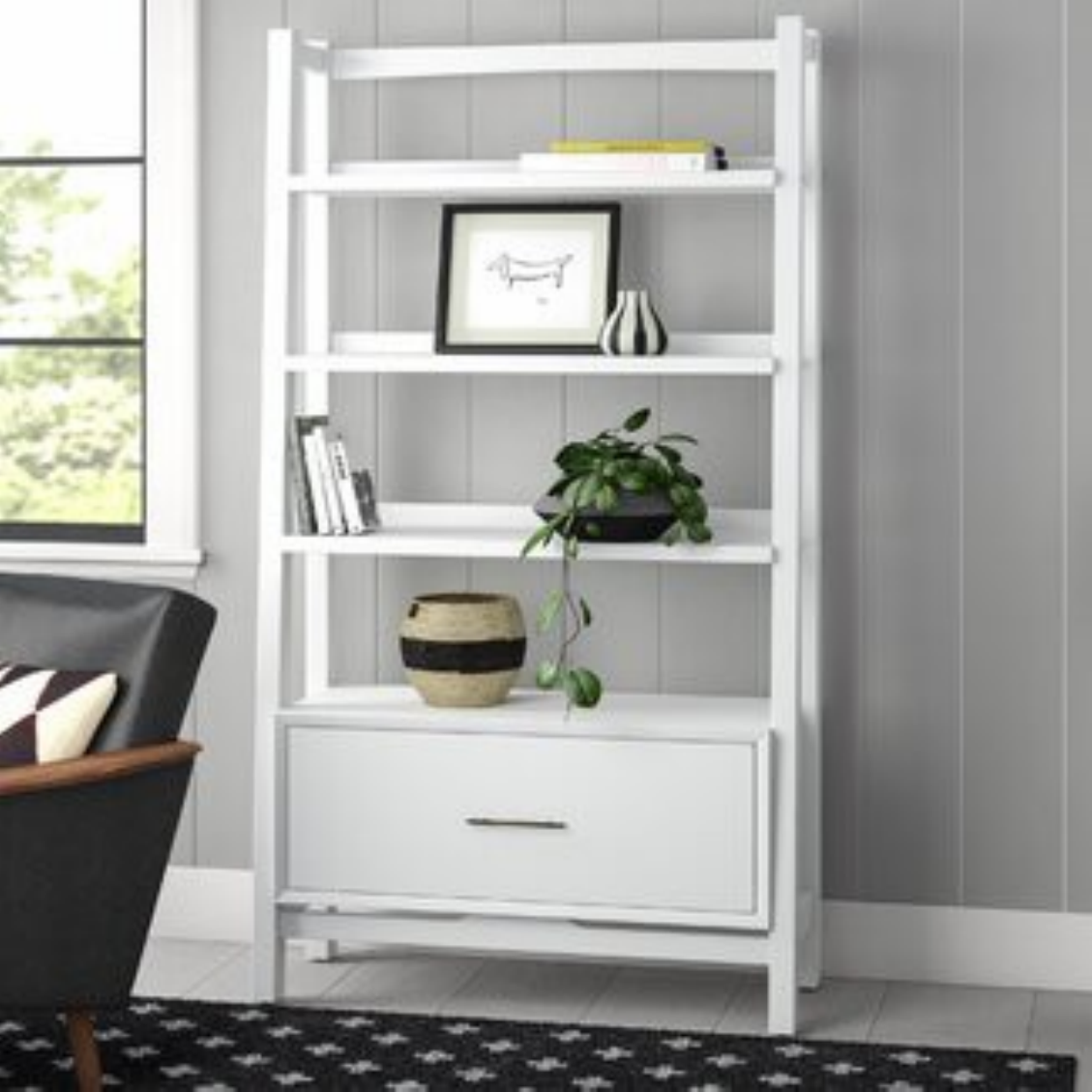}}
      {\includegraphics[height=1.30cm,keepaspectratio]{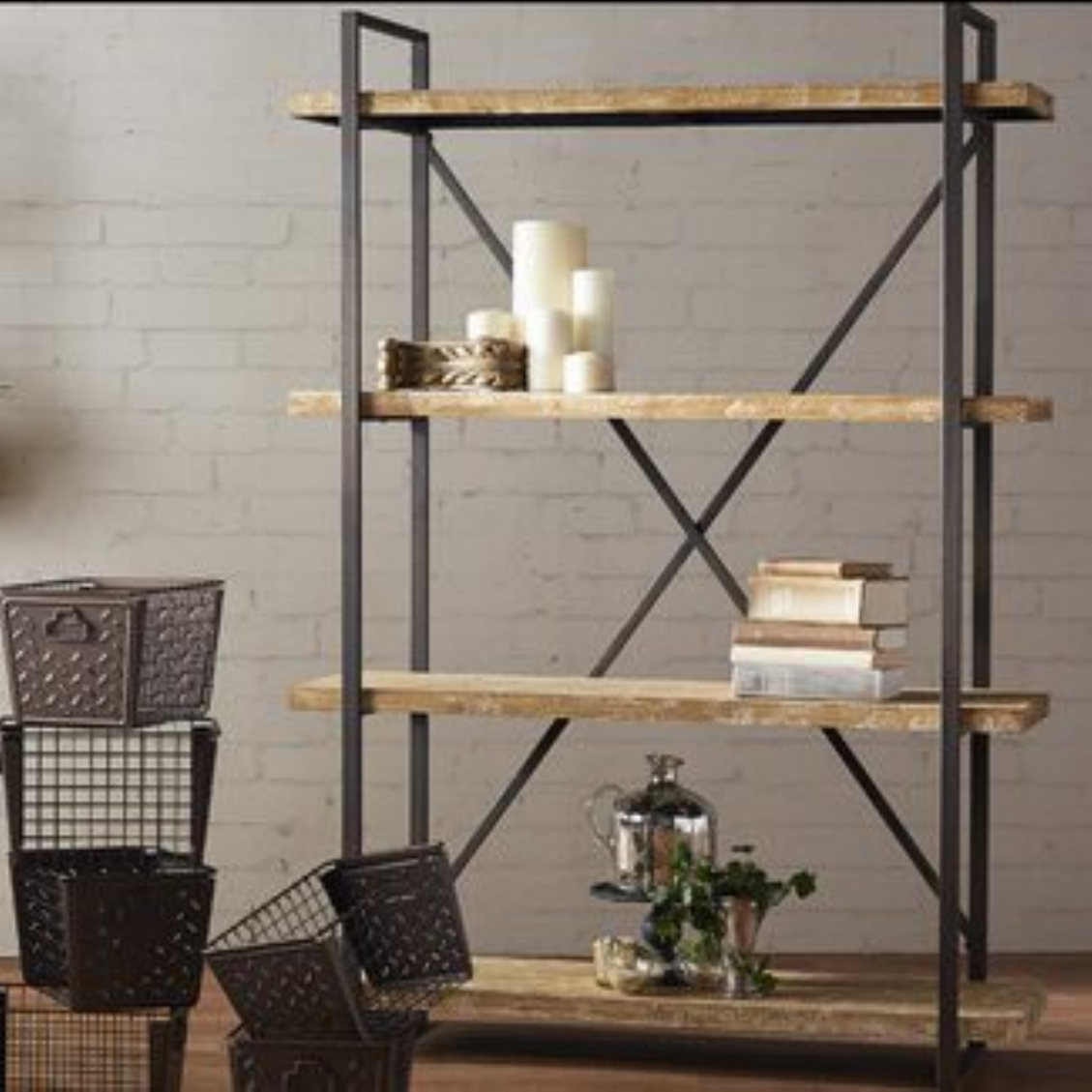}}
      {\includegraphics[height=1.30cm,keepaspectratio]{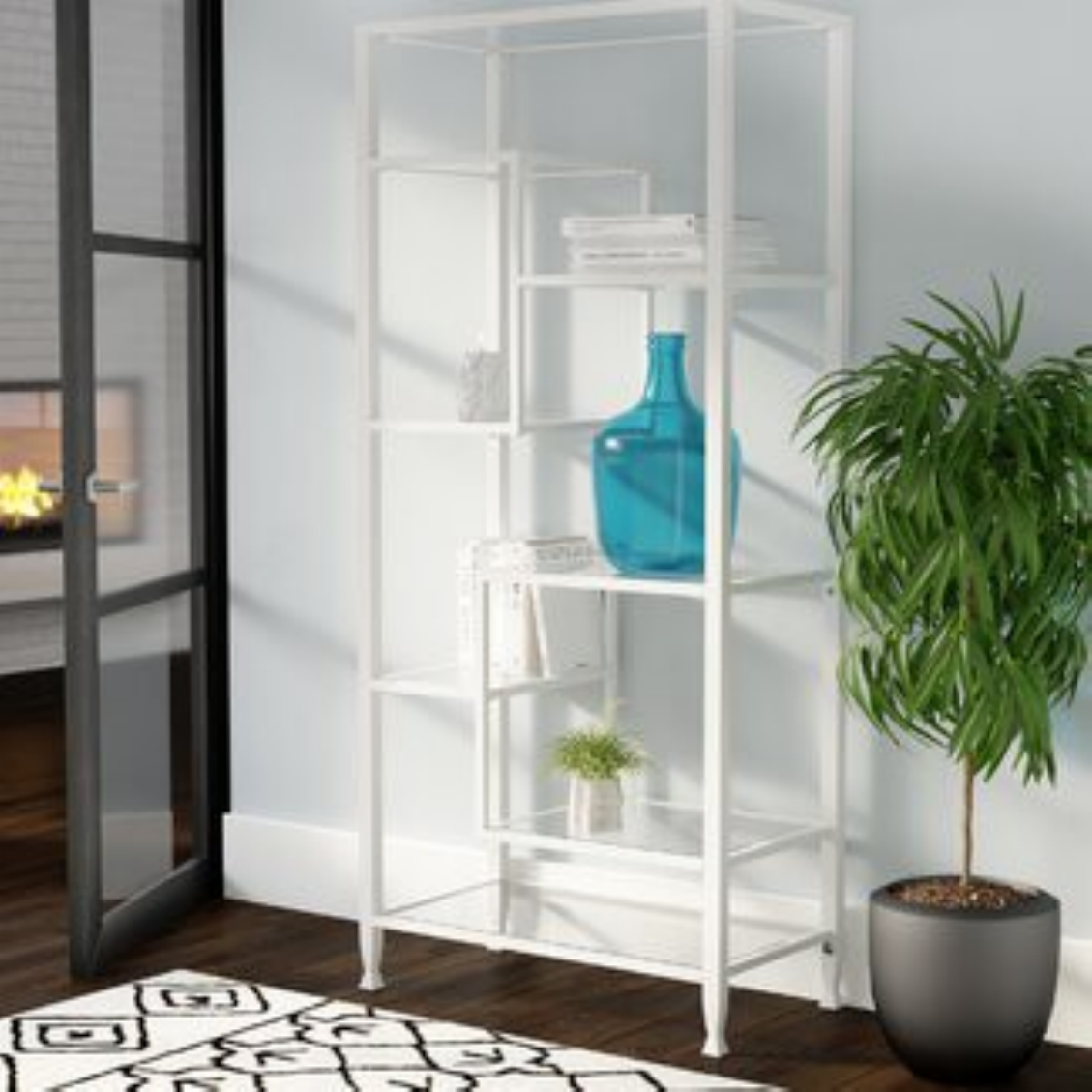}}
      {\includegraphics[height=1.30cm,keepaspectratio]{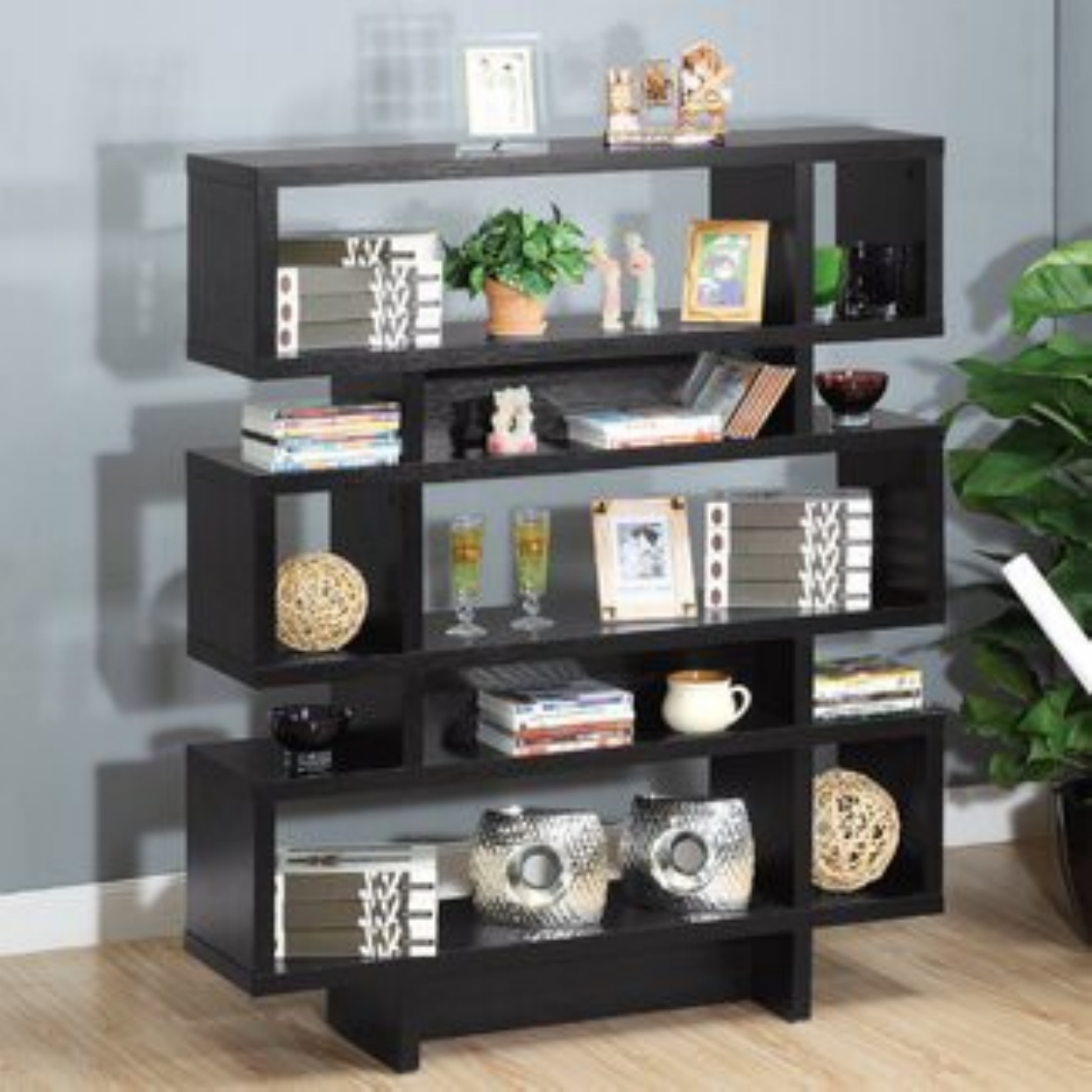}}
      {\includegraphics[trim=60 0 30 0, clip,height=1.30cm,keepaspectratio]{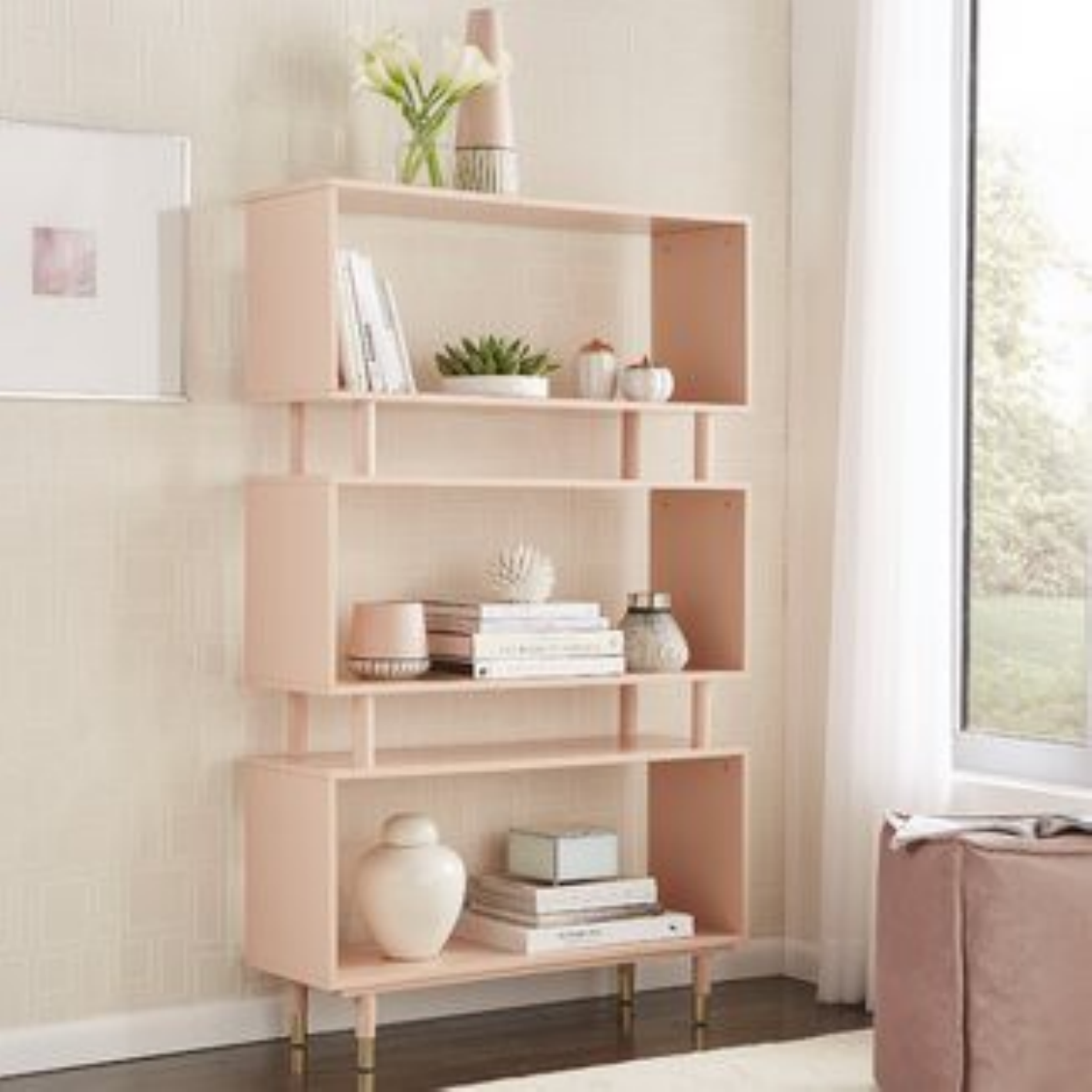}} \\[-0.03in]
      \hline
      {\includegraphics[height=1.30cm,keepaspectratio]{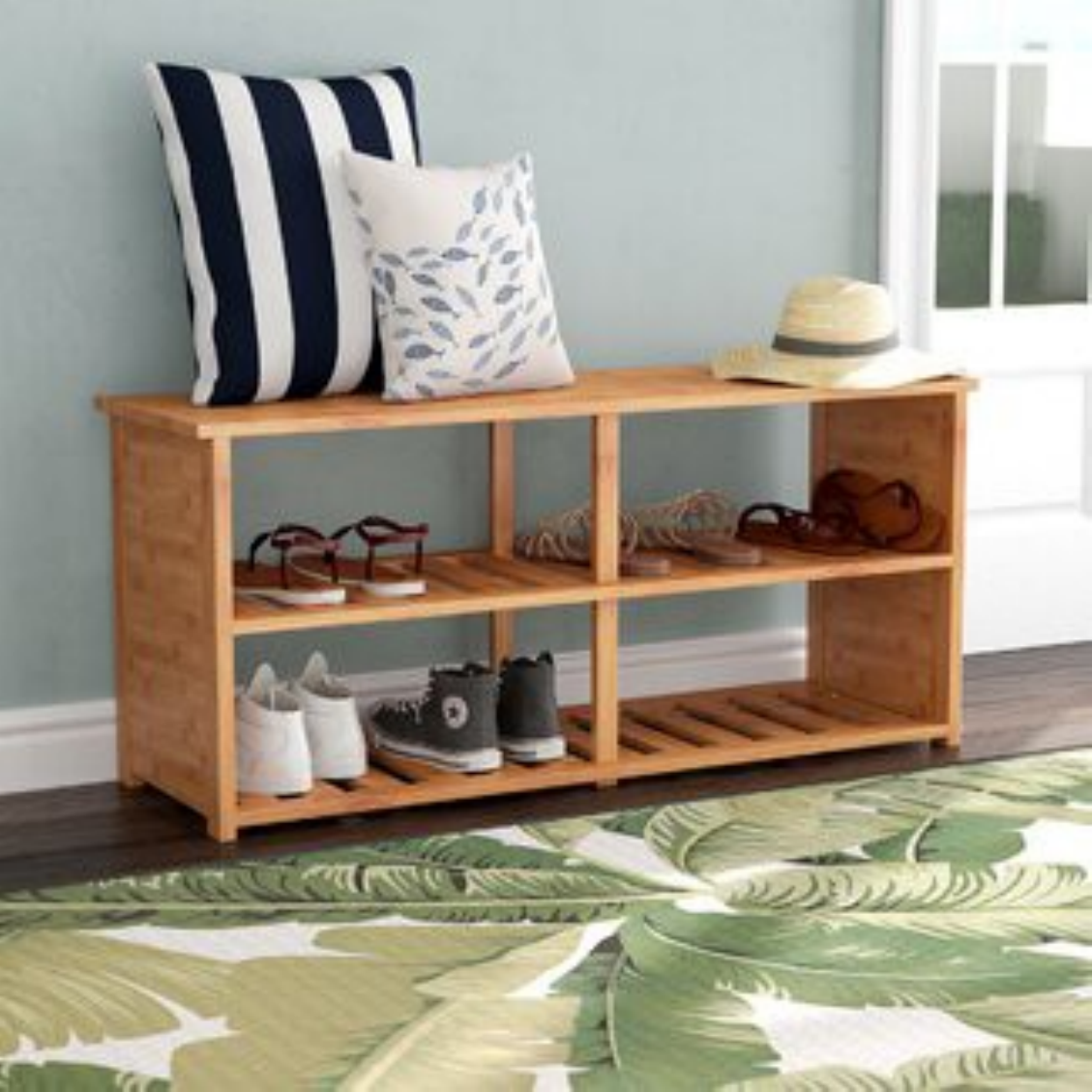}} &
      {\includegraphics[height=1.30cm,keepaspectratio]{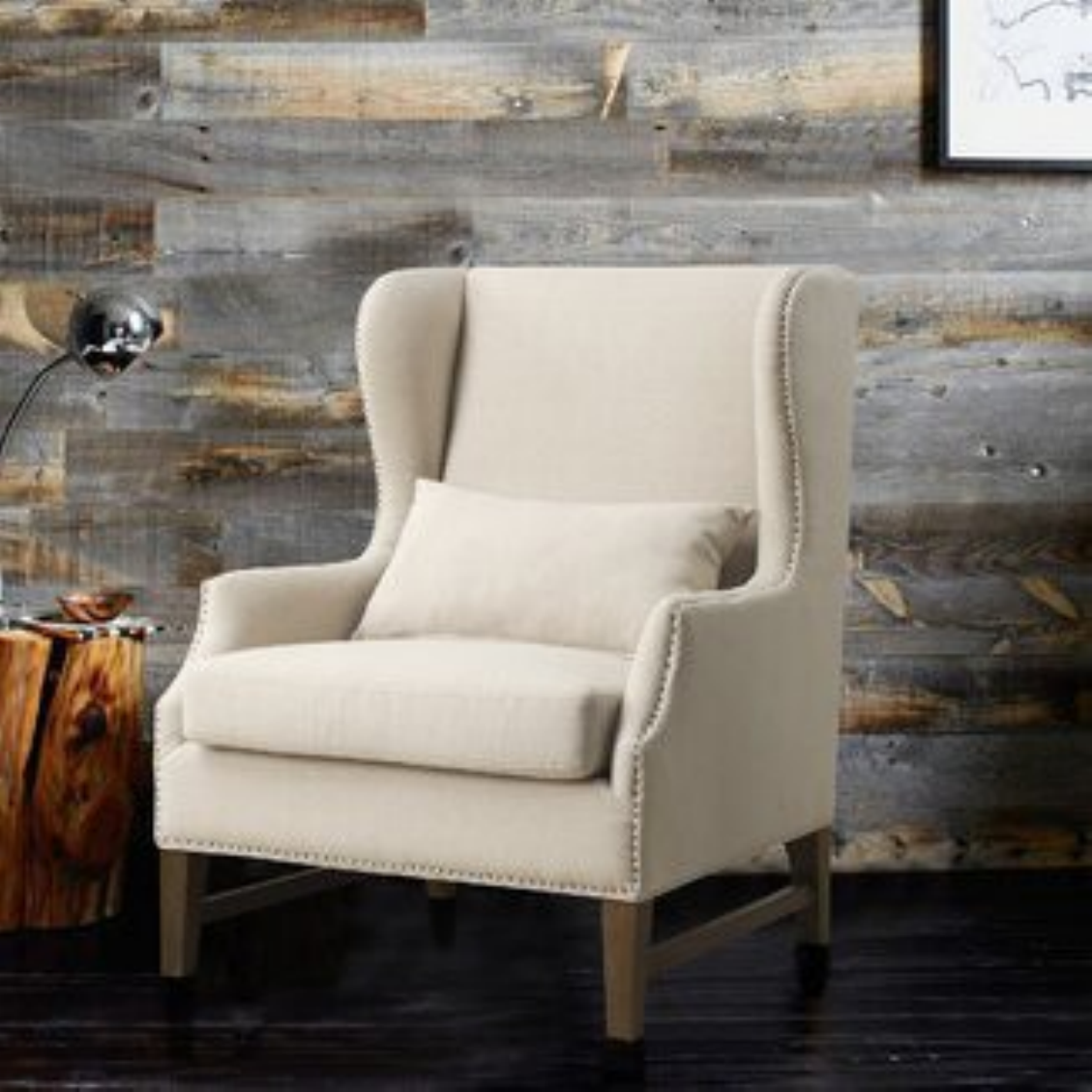}}
      {\includegraphics[height=1.30cm,keepaspectratio]{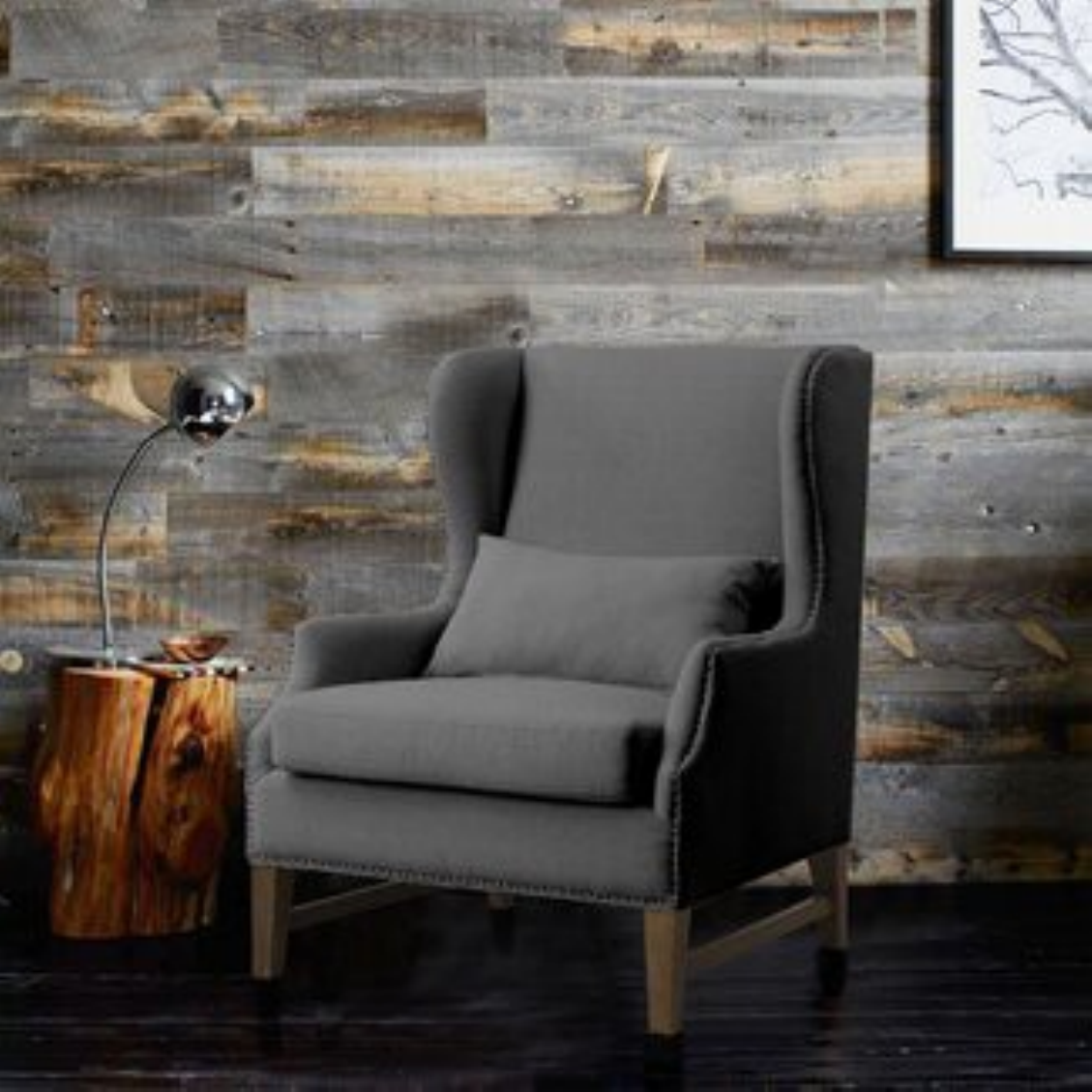}}
      {\includegraphics[height=1.30cm,keepaspectratio]{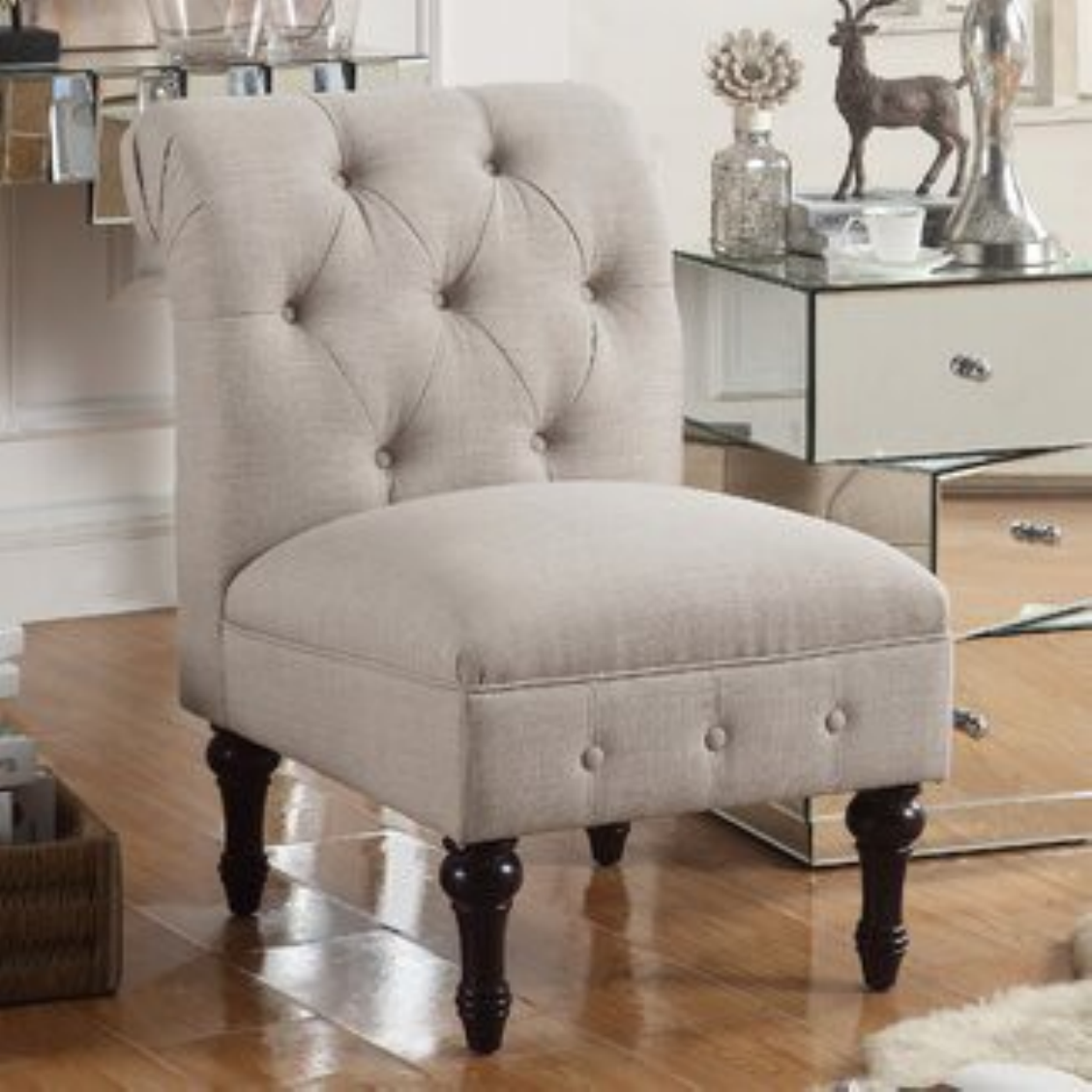}}
      {\includegraphics[height=1.30cm,keepaspectratio]{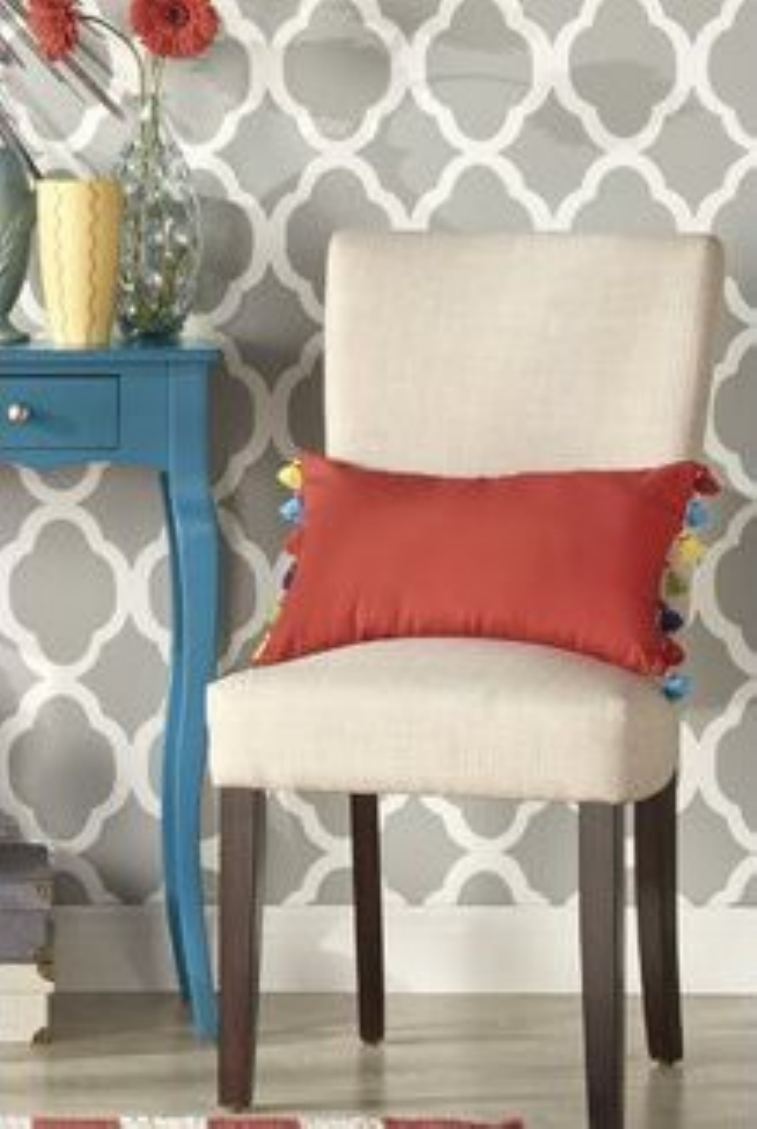}}
      {\includegraphics[height=1.30cm,keepaspectratio]{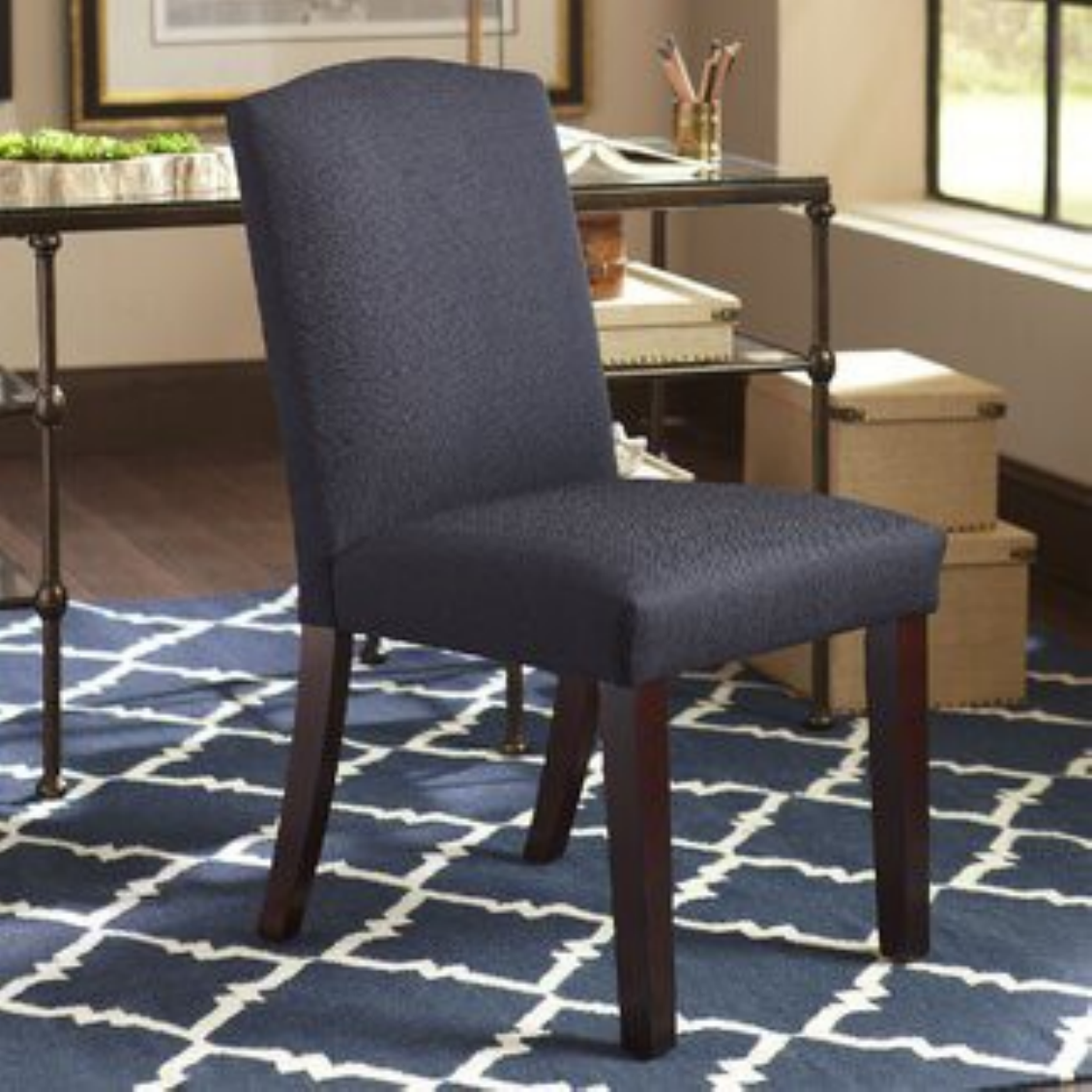}} \\[-0.03in]
      \hline
      {\includegraphics[height=1.30cm,keepaspectratio]{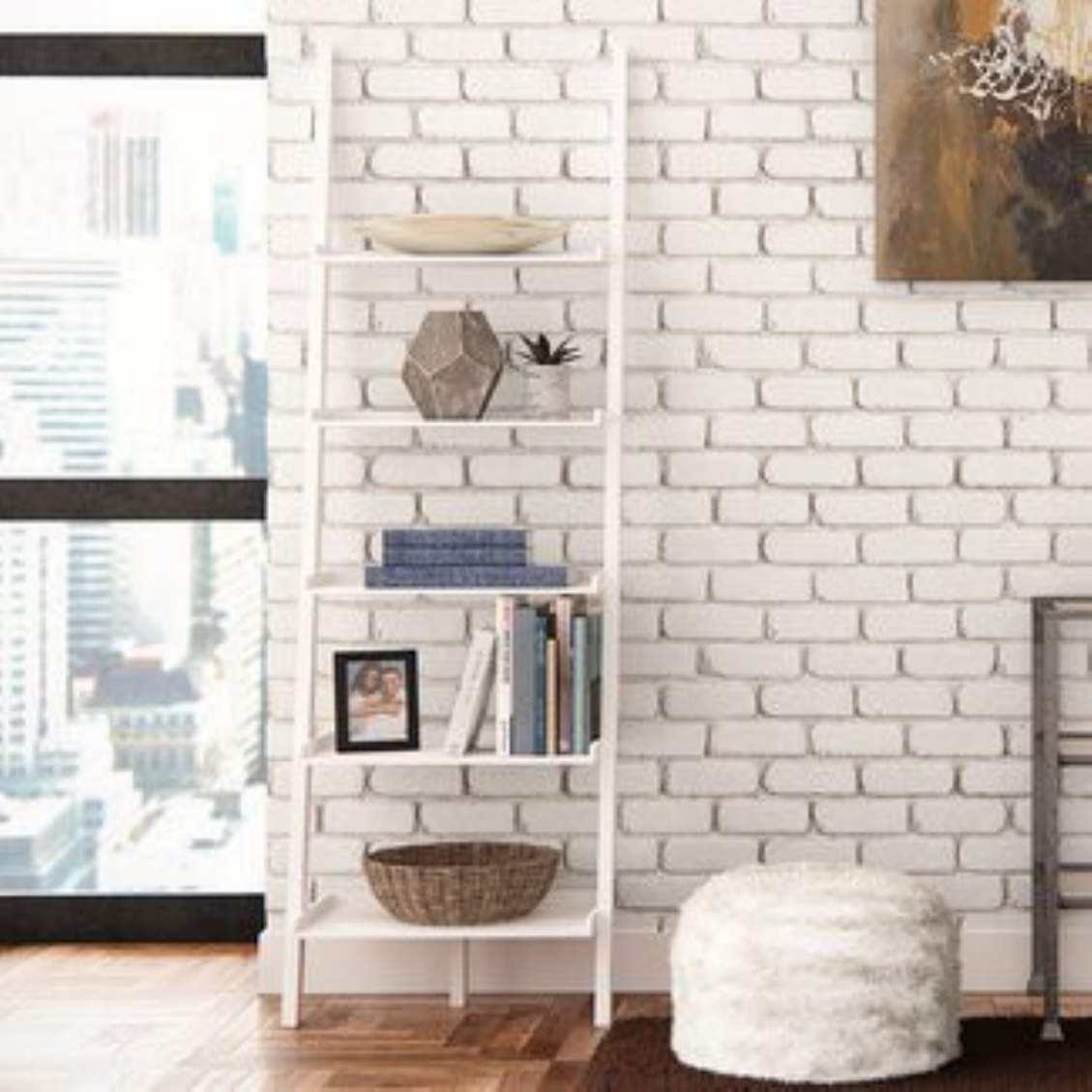}} &
      {\includegraphics[height=1.30cm,keepaspectratio]{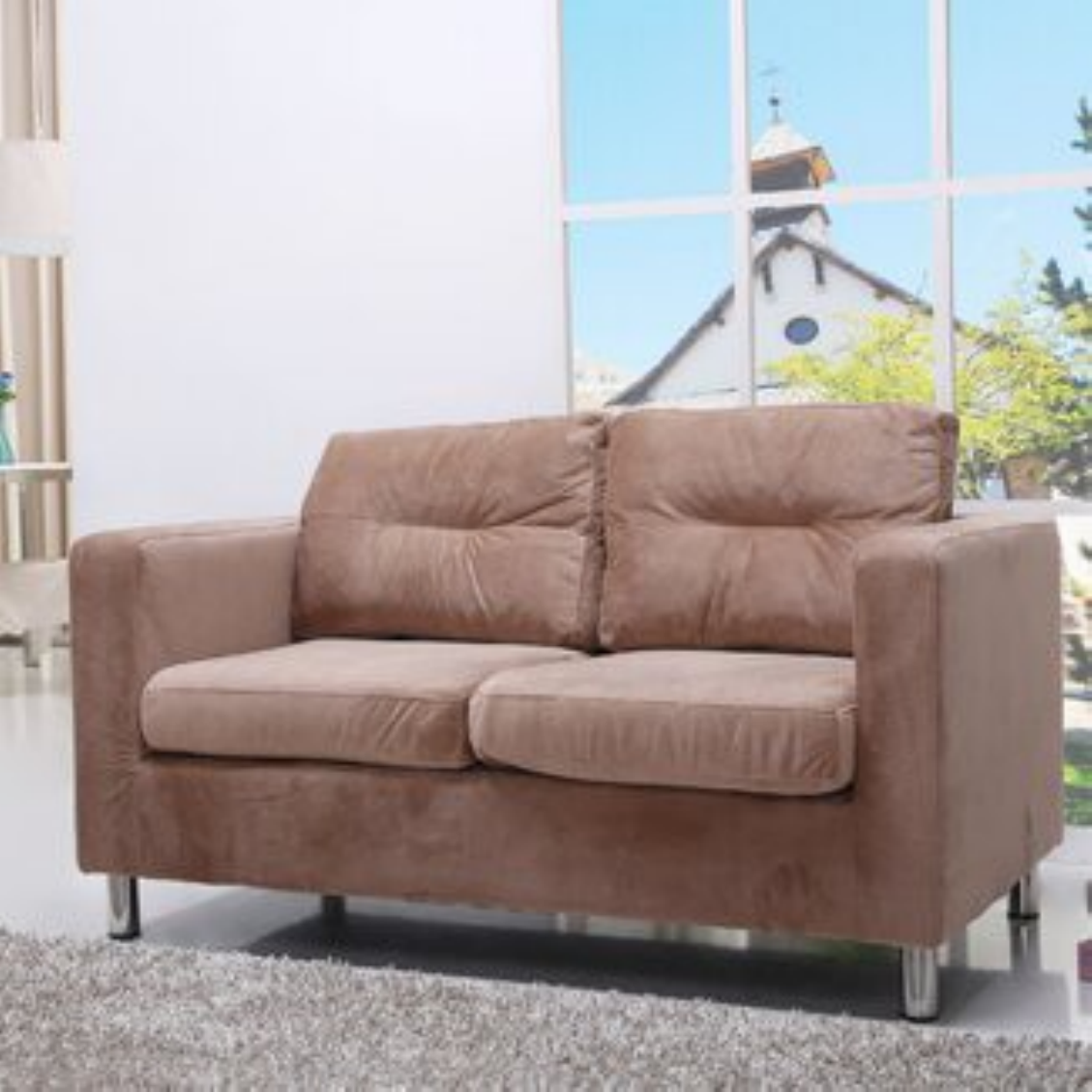}}
      {\includegraphics[height=1.30cm,keepaspectratio]{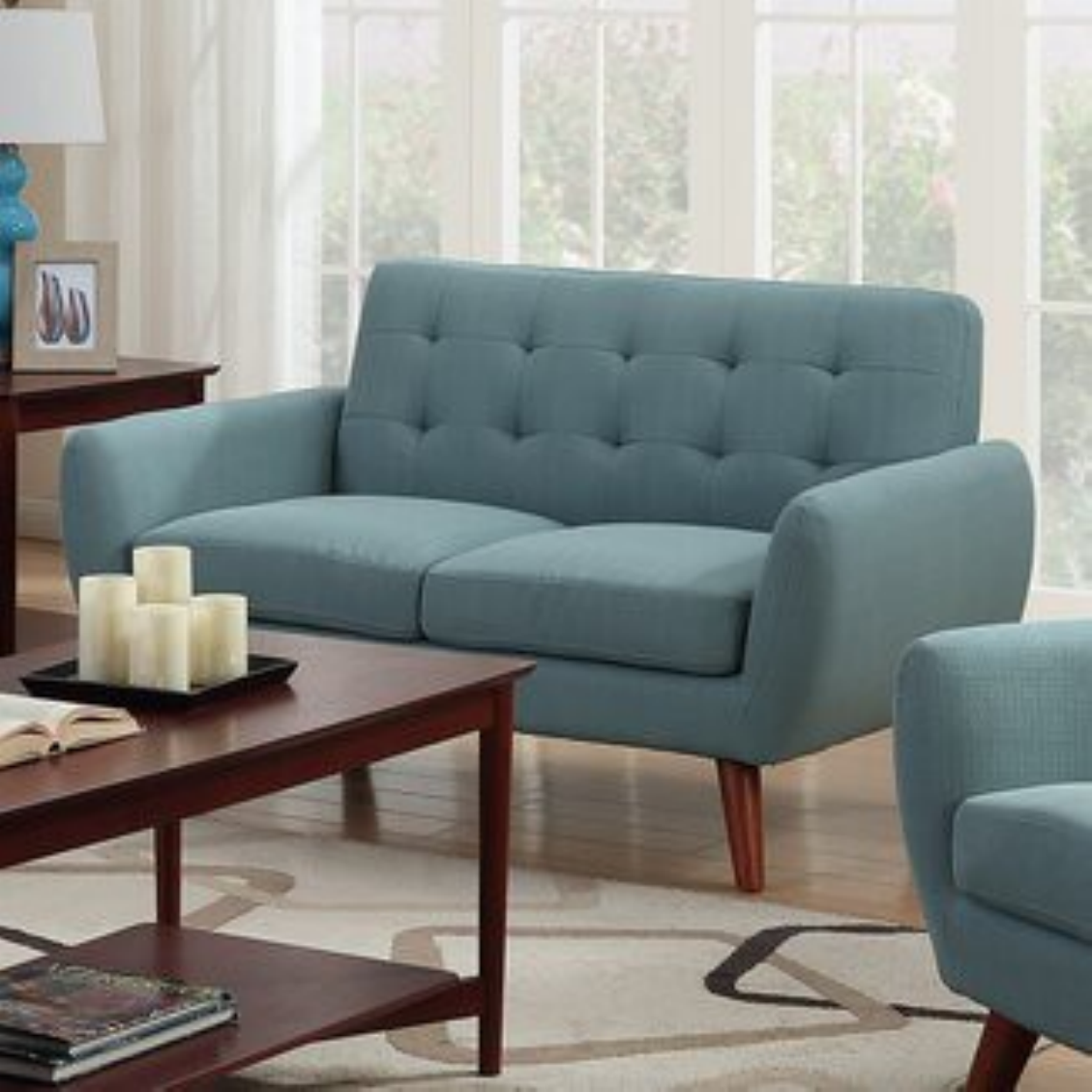}}
      {\includegraphics[height=1.30cm,keepaspectratio]{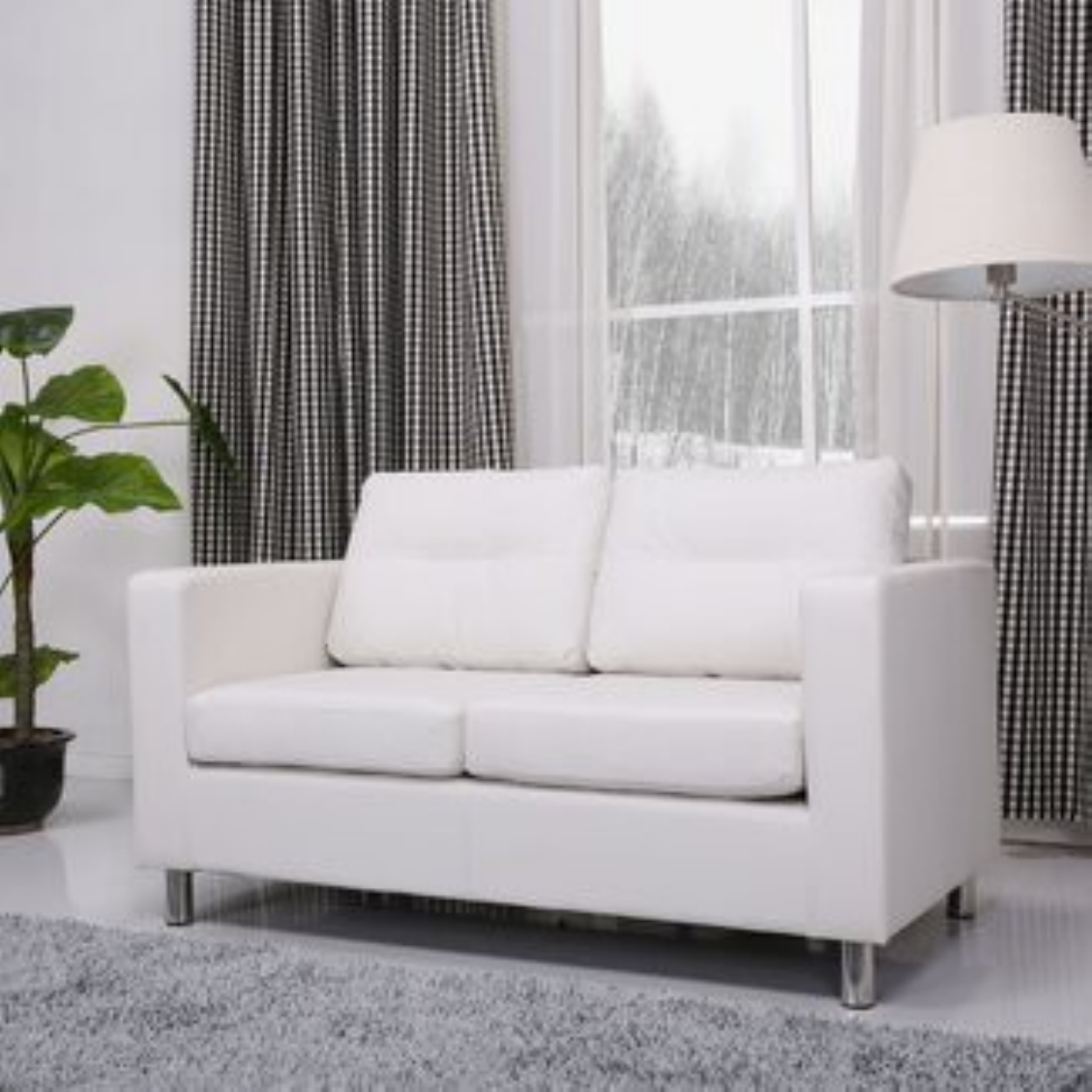}}
      {\includegraphics[trim=80 0 80 0, clip,height=1.30cm,keepaspectratio]{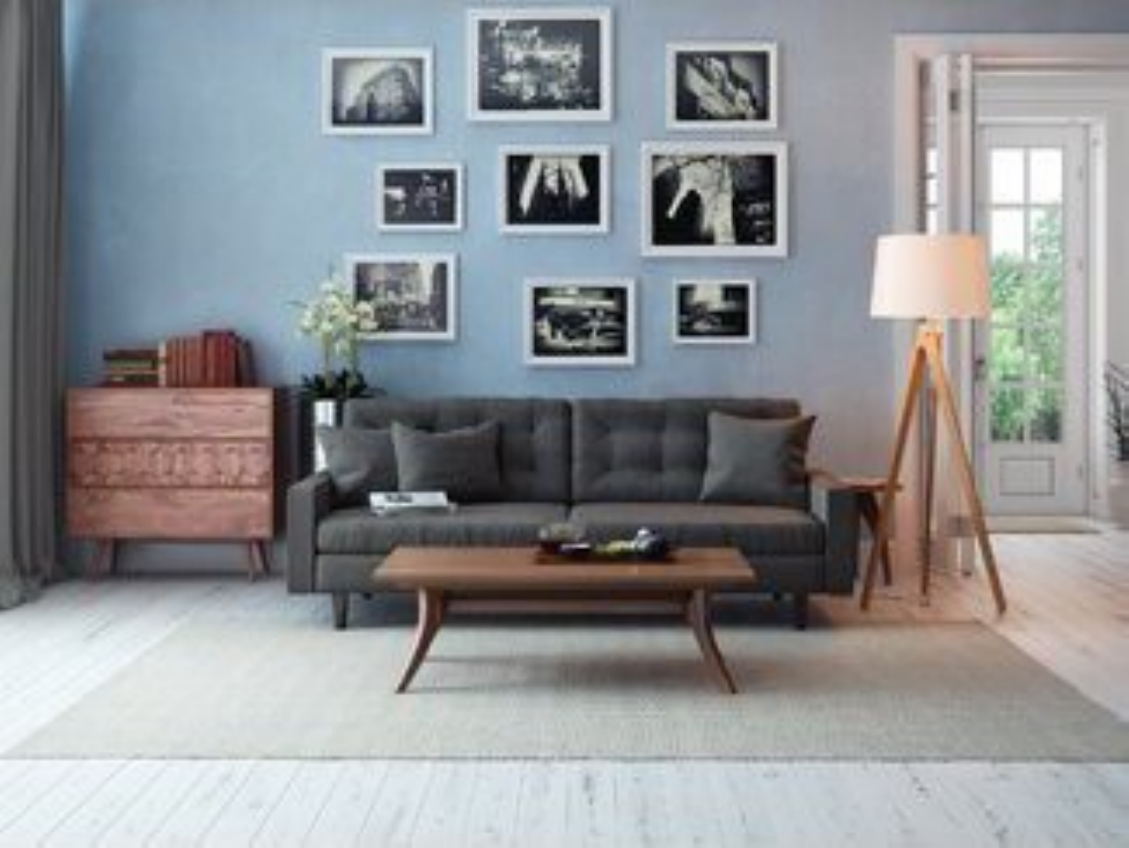}}
      {\includegraphics[trim=10 0 0 0, clip,height=1.30cm,keepaspectratio]{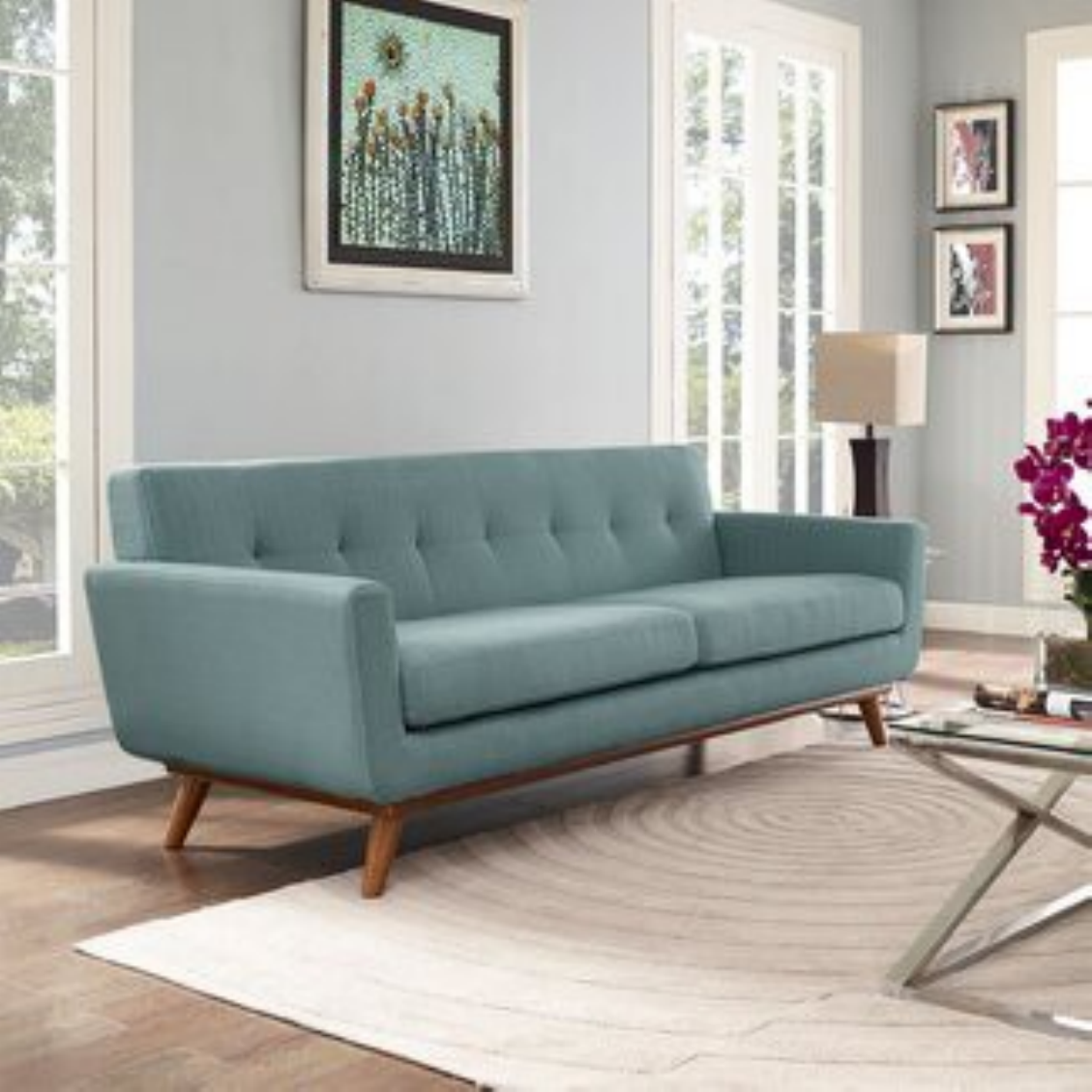}} \\
  \end{tabular}
  \vspace{-3mm}
  \end{center}
\end{table}

\section{Conclusion \& Future Work}
In this work, we introduced a deep learning based system for estimating
style-compatibility of furniture images and 3D models.
We then integrated our method with a multiple component system to exemplify 3D model style-compatibility.
The system includes a deep learning network for style estimation,
visual validation of furniture images and respective 3D models, and a web-based application for creating style-compatible scenes.

In contrast to previous work,
we learn furniture style using curated interior images.
Images are labeled with $4$ styles by multiple experts.
We introduced comparison labels, which allows us to distinguish between a wider spectrum of styles, even for ambiguous inputs.
Furthermore, comparing style of different classes is not possible.
For example, being modern cannot be greater or less than being traditional.
To evaluate our method, we measured accuracy and performance of various style estimation tasks.
We conducted multiple scene modeling experiments for demonstrating 3D model style-compatibility in a variety of scene and style settings.

Our system has several notable directions for future work.
First, we are working toward adding more styles to our
model. The main challenge would be in
accumulating sufficient annotated images, while
maintaining style estimation performance attained on less styles.
As previously reported (Table~\ref{fig:confusion}), styles are subjective even among experts, so it will be beneficial to understand what new styles to introduce that would be sufficiently distinct from current styles.
It will also be interesting to utilize an unsupervised approach to learn style.
Second, reducing the number of manual user steps in the system would be beneficial to future users.
One such opportunity is when validating if a furniture image's is visually consistent with its 3D model.
Here we can draw inspiration from deep networks that aim to learn a common embedding between 3D models and images~\cite{li2015joint}.
Third, our model infers style compatibility based on categories such as modern, traditional, cottage and coastal.
Style-compatibility is not sufficient to guarantee that scenes will be harmonious, for non stylistic reasons. For example, furniture combinations might not match in respect to function or color. It will be worthwhile to extend our framework for such considerations.
Lastly, our network measures style compatibility of individual objects by using images of curated scenes as a context.
Inferring interior style-compatibility spectrum of individual furniture objects directly would be beneficial. We believe this would be an exciting direction for future work.

\section*{Acknowledgements}
The authors would like to thank the computer vision team at Wayfair, as well as Michael Rinaldi and Michael Schenck.
Ataer-Cansizoglu contributed to this work while employed by Wayfair.
The work is partially supported by Weiss’ StartUp Grant from the New Jersey Institute of Technology.

\printbibliography

\end{document}